\documentclass[onefignum,onetabnum]{siamart251216}

\usepackage{amssymb}
\hypersetup{colorlinks=true,linkcolor=red,citecolor=blue,urlcolor=cyan}
\usepackage{bm}
\usepackage{booktabs}
\usepackage{graphicx}
\usepackage{placeins}
\usepackage{array}
\usepackage{tikz}
\usetikzlibrary{arrows.meta,calc,decorations.pathreplacing}
\newcommand{\RR}{\mathbb{R}}

\newcommand{\CC}{\mathbb{C}}
\newcommand{\calA}{\mathcal{A}}
\newcommand{\calB}{\mathcal{B}}
\newcommand{\calC}{\mathcal{C}}
\newcommand{\calD}{\mathcal{D}}
\newcommand{\calE}{\mathcal{E}}
\newcommand{\calF}{\mathcal{F}}
\newcommand{\calG}{\mathcal{G}}
\newcommand{\calH}{\mathcal{H}}
\newcommand{\calI}{\mathcal{I}}
\newcommand{\calK}{\mathcal{K}}
\newcommand{\calL}{\mathcal{L}}

\newcommand{\calN}{\mathcal{N}}
\newcommand{\calP}{\mathcal{P}}

\newcommand{\calS}{\mathcal{S}}

\newcommand{\calV}{\mathcal{V}}
\newcommand{\calW}{\mathcal{W}}
\newcommand{\calZ}{\mathcal{Z}}

\newsiamremark{remark}{Remark}

\AtBeginDocument{%
  \crefformat{section}{\S#2#1#3}%
  \crefformat{subsection}{\S#2#1#3}%
  \crefformat{subsubsection}{\S#2#1#3}%
  \Crefformat{section}{\S#2#1#3}%
  \Crefformat{subsection}{\S#2#1#3}%
  \Crefformat{subsubsection}{\S#2#1#3}%
}

\headers{Universality of Equivariant Message Passing}{P.~Ming and H.~Wang}

\title{Universal Approximation by Equivariant Message-Passing Graph Neural Networks
  with a Bilinear Readout% \thanks{Submitted to the editors DATE.}
}

\author{Pingbing Ming\thanks{SKLMS, Institute of Computational Mathematics and
  Scientific/Engineering Computing, AMSS, Chinese Academy of Sciences,
  No.~55, East Road Zhong-Guan-Cun, Beijing 100190, China
  (\email{mpb@lsec.cc.ac.cn}).}
  \and Han Wang\thanks{National Key Laboratory of Computational Physics,
  Institute of Applied Physics and Computational Mathematics,
  Fenghao East Road 2, Beijing 100094, P.R.~China, and
  HEDPS, CAPT, College of Engineering, Peking University,
  Beijing 100871, P.R.~China
  (\email{wang\_han@iapcm.ac.cn}).}}
\usepackage{docmute}
\newcommand{\arxivbuild}{}

\begin{document}
% Restore section and reference handling after the main-text appendices.
\let\arxivmainsection\section
\makeatletter
\let\arxivmainrefstepcounter\H@refstepcounter
\let\arxivmainchapapp\Hy@chapapp
\makeatother

\maketitle

\begin{abstract}
% OUTLINE
% BREMP reference architecture: linear equivariant message passing + bilinear readout.
% One convolution layer is already equivariantly complete; the filter degree is the
% shell multiplicity kappa of kappa-generic environments (kappa=1 generic, kappa=M all).
% Bilinear readout turns equivariant completeness into UAT.
% L layers are complete on the L-hop environment under connectivity alone: no overlap,
% no genericity; depth lowers the per-layer degree from kappa(R_c) to kappa(r_c).
% DPA4 inherits universal approximation.
We prove universal approximation for the bilinear-readout equivariant message-passing network (BREMP), a reference architecture for equivariant graph neural network interatomic potentials.
Its layers combine equivariant convolutions with affine maps, and its readout forms invariants from products of the equivariant features.
We construct a one-layer feature that determines the local environment up to permutations of atoms of the same type.
The construction uses spherical harmonic filters of degree no larger than the largest number of neighbors at a common distance from the center.
Iterating the construction recovers the environment reachable in as many hops as there are layers, using the same bound on filter degrees at the layer cutoff.
The readout approximates every continuous invariant function of the features uniformly on compact sets.
Together these give universal approximation of atomic energies at a prescribed physical range using a shorter layer cutoff.
We assume that, throughout the admissible class, message passing reaches all neighbors in each physical environment within a common number of layers.
% Removed 2026-09-20: "This number is one for almost every configuration, and it is larger only
% when several neighbors lie on one shell, as in a crystal."
Finally, DPA4, an interatomic potential architecture in production use, inherits this result by approximating the BREMP layers constructed in the proof and realizing the bilinear readout.
% Superseded abstract (2026-09-18):
% We prove universal approximation for a class of equivariant graph neural network interatomic potentials.
% The reference architecture consists of linear equivariant message-passing layers with spherical harmonic filters and a readout that forms the invariants by Clebsch--Gordan bilinear blocks.
% One layer is shown to be complete, that is, to separate the local environments that are not related by a rotation, with filters of degree equal to the largest number of neighbors at a common distance from the center, which is one for almost every configuration and reaches the number of neighbors on a shell for a perfect crystal.
% The bilinear readout approximates every continuous invariant function of the features and realizes polynomial equivariants exactly.
% Combined, these give universal approximation for the atomic energy of a prescribed physical range by a network whose layers use a shorter cutoff, under the sole condition that every pair of atoms within that range is connected through the layer cutoff, without any genericity assumption.
% Depth extends the range and does not complete the features.
% The Wigner bilinear blocks of DPA4 realize the bilinear blocks with a single nonzero frame index, so that DPA4 inherits the result.
% \WH{abstract to be written}
\end{abstract}



%%%%%%%%%%%%%%%%%%%%%%%%%%%%%%%%%%%%%%%%%%%%%%%%%%%%%%%%%%%%%%%%%%
\section{Introduction}
\label{sec:intro}
%%%%%%%%%%%%%%%%%%%%%%%%%%%%%%%%%%%%%%%%%%%%%%%%%%%%%%%%%%%%%%%%%%
% OUTLINE (about 2 pages)
% - MLIPs and the two routes: invariant features (HGNN family, companion) vs equivariant
%   features (TFN, NequIP, MACE, eSCN/Equiformer, DPA4).
% - Three questions: can a multi-layer equivariant GNN on a cutoff graph approximate any
%   PES; which of the nonlinearities in use suffice; how high a degree l is needed.
% - Known results and gaps: Dym--Maron (fully connected, raw tensor product); Joshi,
%   Sverdlov--Dym (separation on sparse graphs, generic/idealized); Nigam ("only in the
%   limit of infinitely deep"); Cen 2024 (high degrees necessary); MACE borrows ACE.
% - Contributions (five): (i) one-layer equivariant completeness with degree kappa on
%   kappa-generic environments; (ii) UAT of the bilinear network and exact realization of
%   polynomial equivariants; (iii) single- and multi-layer UAT under connectivity alone;
%   (iv) depth lowers the per-layer degree from kappa(R_c) to kappa(r_c).
%   (Dropped 2026-09-17: (v) necessity, gated readouts are two-body; Section 4.2 removed.)
% - Relation to the companion paper: shared PES setting and completeness<=>UAT; the
%   equivariant route removes the overlap and genericity hypotheses at the price of a
%   degree growing with kappa.
% - Organization.
% (Written 2026-09-17 as five paragraphs: routes, known results and gap, three questions,
%  architecture and main result in words, organization. Contributions are not itemized.)

Machine-learned interatomic potentials write the energy of an atomic configuration as a sum of atomic energies~\cite{behler2007generalized,bartok2010gaussian}, a decomposition justified by the locality of electronic structure models~\cite{chen2016locality,ortner2020locality,thomas2022body}.
Each atomic energy is a continuous function of the environment of one atom within a cutoff radius, invariant under rotations and under permutations of atoms of the same type.
Whether a network can approximate every function of this kind is a question of universal approximation.
Two families of architectures approach it differently.
The first builds invariant features from distances and angles and passes messages between them, as in CHGNet~\cite{deng2023chgnet}, DPA3~\cite{zhang2025dpa3}, and the invariant networks analyzed in~\cite{hgnn}.
The second carries features that transform under the irreducible representations of the rotation group, passes messages between them by equivariant convolutions, and forms invariants from these features.
This family includes tensor field networks~\cite{thomas2018tfn}, NequIP~\cite{nequip2022}, MACE~\cite{batatia2022mace}, eSCN and Equiformer~\cite{passaro2023reducing,liao2023equiformerv2}, and DPA4~\cite{dpa4}.
This paper concerns the second family.

Universal approximation of equivariant maps is understood on fully connected inputs, that is, for a fixed number of atoms with every pair interacting.
For a compact group, a perceptron on a finite generating set of polynomial invariants, multiplied into a finite generating set of polynomial equivariants, approximates every continuous equivariant map~\cite{yarotsky2022universal}.
For point clouds with a fixed number of atoms such approximations are realized by tensor field networks~\cite{dym2021universality}, by layers of tensor powers~\cite{finkelshtein2022simple}, and, without irreducible representations, by perceptrons on invariant scalars whose outputs multiply the input vectors~\cite{villar2021scalars}.
In the construction of~\cite{dym2021universality} the degree of the polynomial, the degree of the irreducible representations, and the number of layers grow together.
Cen et al.~\cite{cen2025universally} obtain complete equivariant graph networks through a canonical form of the input, again on fully connected graphs.
For a single atomic environment, the atomic cluster expansion provides a polynomial basis of invariants that is dense as body order and degree increase~\cite{dusson2022ace}.
Nigam et al.~\cite{nigam2024completeness} interpret equivariant network features as density correlations and discuss existing completeness guarantees in the limit of unbounded depth.
They also give a complete local construction that applies nonlinear functions to correlations with two reference neighbors before summing over the reference pairs.

On cutoff graphs the available results are separation results, which become approximation results by factoring a continuous invariant function through a separating map~\cite{dym2024low}.
Joshi et al.~\cite{joshi2023expressive} compare geometric networks with a Weisfeiler--Leman test whose aggregation is an idealized injective map.
Sverdlov and Dym~\cite{sverdlov2025expressive} prove that message passing with vector features identifies generic connected graphs, with the exceptional set left implicit and the parameters obtained for almost every choice.
% Superseded opening (2026-09-20):
% For fully connected point clouds with a fixed number of atoms, Dym and Maron~\cite{dym2021universality} proved that networks of this kind approximate every continuous equivariant function.
% Their construction represents equivariant polynomials exactly, and the degree of the polynomial, the degree of the irreducible representations, and the number of layers grow together.
% Nigam et al.~\cite{nigam2024completeness} read the features of such networks as density correlations of increasing body order and conclude that their completeness is guaranteed only in the limit of infinitely deep networks.
% On cutoff graphs the available results are separation results.
On the negative side, message passing on distances alone is incomplete at every depth~\cite{pozdnyakov2022incompleteness}.
Point symmetries can also force equivariant outputs in certain nonzero degrees to vanish~\cite{cen2024hegnn}.
What is missing is an approximation theorem for atomic energies on a cutoff graph at a prescribed physical range.
Such a theorem should apply to every configuration satisfying explicit geometric conditions, construct the network from the stated operations, and specify sufficient representation degrees.

Three questions are addressed.
First, can a network with a layer cutoff shorter than the physical range approximate every atomic energy of that range?
We assume that all neighbors in each physical environment are reached within a common number of message-passing layers.
Second, which degrees of the irreducible representations suffice in the message-passing layers, and how do they depend on the environment?
Third, does the production architecture DPA4 inherit this approximation property?

We answer these questions for the BREMP, a reference architecture whose layers combine equivariant convolutions with affine maps and whose readout forms invariants from products of the equivariant features.
The following is an informal statement of \cref{thm:uat-multi}.

\smallskip
\noindent\textbf{Main result.}
\emph{Consider a class of configurations and fix a common depth $T$.
Assume that all neighbors within the physical range of every atom are reachable in at most $T$ steps on the neighbor graph defined by the layer cutoff.
Assume also that, within the layer cutoff, no atom has more than $\kappa$ neighbors at a common distance.
Then $T$-layer BREMPs with spherical harmonic filter degrees at most $\kappa$ are universal approximators for the atomic energy at the physical range, with their feature spaces and bilinear readouts chosen freely.}
\smallskip

Any architecture that can simulate the BREMPs constructed in the proof inherits their approximation property.
\Cref{cor:dpa4} establishes this transfer for DPA4.
% (2026-09-17: sentences on the shell multiplicity extremes, the overlap and genericity
%  contrast, the one-layer completeness, and the DPA4 frame restriction removed here;
%  they are all developed in the body.)

The paper is organized as follows.
\Cref{sec:setting} fixes the configurations and target functions and introduces the network architectures.
\Cref{sec:complete} proves completeness for one layer and for several layers.
\Cref{sec:readout} proves the universality of the bilinear network and its exact realization of polynomial equivariants.
\Cref{sec:uat} combines these results to prove universal approximation by the BREMP.
\Cref{sec:uat-dpa4} transfers this property to DPA4.
\Cref{sec:discussion} discusses the results and lists open problems.
Appendix~\ref{app:dpa4} specifies the DPA4 operators used in the proofs.
Appendix~\ref{app:wbb-quadrature} constructs the exact quadrature for Wigner analysis.
% \WH{introduction to be written}

%%%%%%%%%%%%%%%%%%%%%%%%%%%%%%%%%%%%%%%%%%%%%%%%%%%%%%%%%%%%%%%%%%
\section{Setting and network architectures}
\label{sec:setting}
%%%%%%%%%%%%%%%%%%%%%%%%%%%%%%%%%%%%%%%%%%%%%%%%%%%%%%%%%%%%%%%%%%
% OUTLINE (about 5 pages)

\subsection{Atomic environments and target functions}
\label{sec:setting-pes}

We study atomic energies with a finite interaction range.
The environment and target spaces below adapt the definitions in~\cite{hgnn}.
Consider a configuration of $N$ atoms with positions $\bm r_j\in\RR^3$ and types $z_j\in\calZ=\{1,\ldots,Z\}$.
No two atoms occupy the same position.
We assume a uniform density bound: for some integer $M\ge1$ independent of the configuration, every ball with one of the cutoff radii used below contains at most $M$ atoms.
Let $B_r=\{\bm x\in\RR^3:|\bm x|<r\}$ denote the open ball of radius $r$ centered at the origin.
The \emph{local environment} of atom $j$ at cutoff $r>0$ consists of its type and the unordered set of typed displacements to its neighbors,
\begin{equation}\label{eq:local-env}
  \calD^{(r)}_j=\bigl(z_j,\ \{(\Delta\bm r_{jk},z_k):k\ne j,\ \Delta\bm r_{jk}\in B_r\}\bigr),
  \qquad \Delta\bm r_{jk}=\bm r_k-\bm r_j .
\end{equation}
Using displacements removes the dependence on the absolute position of the center.

For a physical interaction range $R_c>0$, the total potential energy has the form
\begin{equation}\label{eq:extensive}
  E(\{\bm r_i,z_i\}_{i=1}^{N})=\sum_{i=1}^{N}\varepsilon_i,
\end{equation}
where a single atomic energy function $\varepsilon$ determines each contribution through
\begin{equation}\label{eq:locality}
  \varepsilon_i=\varepsilon\bigl(\calD^{(R_c)}_i\bigr).
\end{equation}
The approximation problem is to represent $\varepsilon$ as a function of the local geometry and atom types.
We first specify its domain and symmetry.

\begin{definition}[Environment space]
\label{def:env-space}
Fix $r>0$, a center type $z_0\in\calZ$, and a nondecreasing tuple of neighbor types $\bm z=(z_1,\ldots,z_n)$ with $0\le n\le M$.
The \emph{stratum} $\calE^{(r)}_{z_0,\bm z}$ consists of the ordered displacement tuples
\begin{equation}\label{eq:stratum}
  \calD=(\Delta\bm r_1,\ldots,\Delta\bm r_n)\in(B_r\setminus\{0\})^n,
  \qquad \Delta\bm r_a\ne\Delta\bm r_b\quad(a\ne b),
\end{equation}
with center type $z_0$ and neighbor types $\bm z$.
It carries the Euclidean topology inherited from $\RR^{3n}$.
For $n=0$, the stratum is a single point recording the center type.
The \emph{environment space} is the disjoint union
\begin{equation}\label{eq:env-space}
  \calE^{(r)}
  =\bigsqcup_{z_0\in\calZ}\ \bigsqcup_{n=0}^{M}\
    \bigsqcup_{\substack{\bm z\in\calZ^n\\ z_1\le\cdots\le z_n}}
    \calE^{(r)}_{z_0,\bm z},
\end{equation}
equipped with the disjoint union topology.
\end{definition}

The ordered tuples provide coordinates for the unordered environments in~\cref{eq:local-env}.
Reordering neighbors of the same type changes these coordinates without changing the physical environment.
For the completeness analysis, an equivalent description is given by the center type and the \emph{neighbor measures}.
For $\calD\in\calE^{(r)}_{z_0,\bm z}$ and $z\in\calZ$, set
\begin{equation}\label{eq:neighbor-measure}
  n_z(\calD)=\sum_{\substack{1\le a\le n\\ z_a=z}}\delta_{\Delta\bm r_a},
\end{equation}
where $\delta_{\bm x}$ is the unit point mass at $\bm x$.
These finite measures on $B_r\setminus\{0\}$ retain all typed displacements and do not depend on the ordering of neighbors.

If $n_t$ neighbors have type $t$, the group acting on the stratum is
\begin{equation}\label{eq:sym-group}
  G=SO(3)\times\Gamma,\qquad \Gamma=S_{n_1}\times\cdots\times S_{n_Z}.
\end{equation}
Each factor $S_{n_t}$ consists of the permutations among neighbors with type $t$.
The action is
\begin{equation}\label{eq:group-action}
  (R,\pi)\cdot(\Delta\bm r_1,\ldots,\Delta\bm r_n)
  =(R\,\Delta\bm r_{\pi^{-1}(1)},\ldots,R\,\Delta\bm r_{\pi^{-1}(n)}).
\end{equation}
We write $\calD\sim\calD'$ when two points in the same stratum are related by this action.
Thus equivalence allows a common rotation and a permutation within each type.
On environment spaces, the notation $G$ is understood stratum by stratum.

\begin{definition}[Target function space]
\label{def:target-space}
The space $\calF^0_{R_c}$ consists of functions $\varepsilon:\calE^{(R_c)}\to\RR$ satisfying the following conditions.
\begin{enumerate}
\item \emph{Invariance.} For each stratum,
$\varepsilon((Q,\pi)\cdot\calD)=\varepsilon(\calD)$ for every $Q\in O(3)$ and $\pi\in\Gamma$, with the action in~\cref{eq:group-action} extended to orthogonal matrices.
\item \emph{Continuity within strata.} The restriction of $\varepsilon$ to every $\calE^{(R_c)}_{z_0,\bm z}$ is continuous.
\item \emph{Continuity at the cutoff.} Suppose that a sequence $\calD_q$ in one stratum converges in $\RR^{3n}$ to a tuple $\calD^\ast$ whose neighbors remain distinct from one another and from the center.
Let $\calD^{\ast-}$ be the environment obtained by deleting all neighbors of $\calD^\ast$ at distance $R_c$.
Then $\varepsilon(\calD_q)\to\varepsilon(\calD^{\ast-})$.
\end{enumerate}
\end{definition}

The third condition supplements continuity in the disjoint union topology.
It prevents an energy jump when one or several neighbors leave the interaction range, even if the retained neighbors move at the same time.
For example, a pair energy $\varepsilon(\calD)=\sum_{a=1}^{n}u(z_0,z_a,|\Delta\bm r_a|)$ satisfies this condition when $u(z_0,z,\cdot)$ is continuous on $(0,R_c]$ and vanishes at $R_c$.

The geometric quantity that controls the required angular resolution is the number of neighbors at a common distance from the center.
For $\rho>0$, write $S_\rho(\calD)=\{a:|\Delta\bm r_a|=\rho\}$ for the corresponding \emph{shell}, counting neighbors of all types.

\begin{definition}[$\kappa$-generic environments]
\label{def:kappa-generic}
The \emph{shell multiplicity} of $\calD$ is
$\kappa(\calD)=\max_{\rho>0}|S_\rho(\calD)|$, with $\kappa(\calD)=0$ when there are no neighbors.
For an integer $\kappa\ge1$, an environment is \emph{$\kappa$-generic} if its shell multiplicity is at most $\kappa$.
We denote this class by
\begin{equation}\label{eq:env-kappa}
  \calE^{(r)}_\kappa
  =\{\calD\in\calE^{(r)}:\kappa(\calD)\le\kappa\}.
\end{equation}
A configuration is $\kappa$-generic at cutoff $r$ if $\calD^{(r)}_j\in\calE^{(r)}_\kappa$ for every atom $j$.
\end{definition}

For $\kappa=1$, all neighbor distances from the center are distinct, including distances to neighbors of different types.
The excluded tuples satisfy at least one equality $|\Delta\bm r_a|^2=|\Delta\bm r_b|^2$ with $a\ne b$.
They form a set of Lebesgue measure zero in each stratum.
Consequently, configurations sampled from a distribution with a density in the position coordinates are $1$-generic almost surely.
For $\kappa=M$, the restriction disappears and $\calE^{(r)}_M=\calE^{(r)}$.
This case includes crystalline environments with several neighbors on the same shell.
More generally, $\calE^{(r)}_\kappa$ is open in each stratum because its complement is a finite union of sets where $\kappa+1$ neighbor distances coincide.

A message-passing network assigns a feature vector to each atom.
Each layer updates this vector using information from atoms within a \emph{layer cutoff} $0<r_c\le R_c$.
Successive layers pass information along chains of neighboring atoms.
The \emph{neighbor graph} has one vertex per atom and an edge between distinct atoms $j$ and $k$ when $|\bm r_k-\bm r_j|<r_c$.
Write $\calN(j)$ for the neighbors of $j$ in this graph, and abbreviate $\calE=\calE^{(r_c)}$.

\begin{definition}[$L$-hop environment and connectivity]
\label{def:l-hop}
For an integer $L\ge0$, define the reachable sets recursively by $\calN^0(j)=\emptyset$ and
\begin{equation}\label{eq:l-hop-set}
  \calN^{L+1}(j)=\calN(j)\cup\bigcup_{k\in\calN(j)}\calN^{L}(k).
\end{equation}
The \emph{$L$-hop environment} of $j$ is
\begin{equation}\label{eq:l-hop-env}
  \calD^{[r_c,L]}_j
  =\bigl(z_j,\ \{(\Delta\bm r_{jk},z_k):k\in\calN^L(j)\setminus\{j\}\}\bigr).
\end{equation}
The space of these environments is denoted by $\calE^{[r_c,L]}$.
It is stratified by the center type and neighbor types.
Each stratum carries the topology induced by its displacement coordinates and the group action~\cref{eq:group-action}.
A configuration satisfies the \emph{connectivity condition} for $(r_c,R_c)$ if every pair of distinct atoms at distance less than $R_c$ is joined by a path in the neighbor graph.
\end{definition}

We write $\calE^{[r_c,L]}_\kappa$ for the $L$-hop environments arising from configurations that are $\kappa$-generic at cutoff $r_c$.
This condition bounds shell multiplicities in every one-hop neighborhood, not in the entire $L$-hop environment.
The class is invariant under rotations and type-preserving permutations.

Connectivity ensures that each finite configuration admits a depth $T$ for which
\begin{equation}\label{eq:physical-range-coverage}
  \calD^{(R_c)}_j\subseteq\calD^{[r_c,T]}_j
  \qquad\text{for every atom }j.
\end{equation}
The approximation results require this inclusion with a common $T$ for the entire class of configurations.
When $r_c=R_c$, $T=1$ suffices and $\calD^{[R_c,1]}_j=\calD^{(R_c)}_j$.

\subsection{Equivariant features and convolutions}
\label{sec:blocks}

A graph neural network assigns each atom a vector of internal variables, called its \emph{atomic feature}, to represent information about its type and surroundings.
Each layer updates these features through message passing on the neighbor graph, using neighboring features and relative positions.
In an \emph{equivariant graph neural network}, rotating the configuration and input features before an update gives the same result as rotating the updated features.
For example, scalar components remain unchanged under rotations, while vector components rotate with the configuration.

We describe these rotation rules using real representations of $SO(3)$.
For each integer $l\ge0$, let $V_l=\RR^{2l+1}$ carry the irreducible representation of degree $l$.
In orthonormal coordinates, a rotation $R\in SO(3)$ acts through a real orthogonal Wigner matrix $D^{(l)}(R)$.
Degree $0$ gives rotation-invariant scalars, with $D^{(0)}(R)=1$, and degree $1$ is equivalent to the usual vector representation.
Every finite-dimensional continuous representation of $SO(3)$ decomposes into copies of these irreducible representations~\cite{varshalovich1988}.

For representation spaces $X$ and $Y$ with actions $D_X$ and $D_Y$, a map $f:X\to Y$ is \emph{equivariant} if $f(D_X(R)x)=D_Y(R)f(x)$ for every $R\in SO(3)$ and $x\in X$.
If $D_Y(R)=\mathrm{Id}_Y$ for every $R$, rotations leave the output unchanged, and $f$ is \emph{invariant}: $f(D_X(R)x)=f(x)$.
Composing equivariant maps preserves equivariance.

Atomic features take values in the \emph{feature space} of maximal degree $l_{\max}\ge0$ and channel width $C\ge1$,
\begin{equation}\label{eq:feature-space}
  \calV=\bigoplus_{l=0}^{l_{\max}}V_l\otimes\RR^{C},
  \qquad
  D_{\calV}(R)=\bigoplus_{l=0}^{l_{\max}}D^{(l)}(R)\otimes\mathrm{Id}_C .
\end{equation}
The $C$ copies of each $V_l$ are called \emph{channels}.
For a configuration of $N$ atoms, the features form a family $h=(h_j)_{j=1}^{N}\in\calV^N$.
A rotation acts on every $h_j$ by $D_{\calV}(R)$, and a permutation of the atoms relabels the features.
We write $h^{(l)}_{j,m,c}$ for the component with degree $l$, representation index $m\in\{-l,\ldots,l\}$, and channel $c\in\{1,\ldots,C\}$.
The notation $h^{(l)}_{j,\cdot,c}\in V_l$ collects all representation indices of one channel.
For a single feature $h\in\calV$, we omit the atom index.

By Schur's lemma, an equivariant affine map mixes channels within each degree and has the form
\begin{equation}\label{eq:linear}
  (Lh)^{(l)}_{m,d}=\sum_c W^{(l)}_{dc}\,h^{(l)}_{m,c}+\delta_{l0}\,b_d ,
\end{equation}
where $W^{(l)}$ is a matrix between the input and output channel spaces, shared by all indices $m$ at degree $l$.
The bias is confined to the degree-$0$ channels.
The map is linear precisely when $b=0$.
Applying the same map at every atom gives the self-interaction operation of tensor field networks~\cite{thomas2018tfn}.

The direction of a bond is represented by real spherical harmonics.
For $\hat{\bm r}$ on the unit sphere $S^2$, write $Y^{(l)}(\hat{\bm r})=(Y^{(l)}_m(\hat{\bm r}))_{|m|\le l}\in V_l$.
Let $\tilde Y^{(l)}_m$ be the complex spherical harmonics in the Condon--Shortley convention, orthonormal for the surface measure of total mass $4\pi$.
We fix the real coordinates by~\cite{ivanic1996rotation,blanco1997rotation}
\begin{equation}\label{eq:real-harmonics}
  Y^{(l)}_0=\tilde Y^{(l)}_0,\quad
  Y^{(l)}_m=\sqrt2(-1)^m\operatorname{Re}\tilde Y^{(l)}_m,\quad
  Y^{(l)}_{-m}=\sqrt2(-1)^m\operatorname{Im}\tilde Y^{(l)}_m,\quad 1\le m\le l .
\end{equation}
The harmonics satisfy $Y^{(l)}(R\hat{\bm r})=D^{(l)}(R)Y^{(l)}(\hat{\bm r})$ in the chosen coordinates~\cite{varshalovich1988}.

To combine two features, choose an orthogonal Clebsch--Gordan decomposition
$V_{l_1}\otimes V_{l_2}\cong\bigoplus_{l=|l_1-l_2|}^{l_1+l_2}V_l$~\cite{varshalovich1988}.
For each $l$ in this range, the projection onto its summand defines an equivariant bilinear map
$(u,v)\mapsto(u\otimes v)^{(l)}$, with coordinates
$(u\otimes v)^{(l)}_m=\sum_{m_1,m_2}C^{(l,m)}_{(l_1,m_1)(l_2,m_2)}u_{m_1}v_{m_2}$.
The real Clebsch--Gordan coefficients $C^{(l,m)}_{(l_1,m_1)(l_2,m_2)}$ are normalized by this orthogonal decomposition.
This fixes each map up to a sign, which we choose once for each triple $(l_1,l_2,l)$.
Every equivariant bilinear map $V_{l_1}\times V_{l_2}\to V_l$ is a scalar multiple of this coupling.
Outside the triangle range $|l_1-l_2|\le l\le l_1+l_2$, no nonzero equivariant bilinear map exists.
The couplings $1\otimes1\to0$ and $1\otimes1\to1$ give, up to normalization, the dot and cross products.

An \emph{equivariant convolution} couples each neighbor feature with a spherical harmonic of the bond direction, weights the result by a radial function, and sums over neighbors~\cite{thomas2018tfn,nequip2022}.
Write $r_{jk}=|\Delta\bm r_{jk}|$ and $\hat{\bm r}_{jk}=\Delta\bm r_{jk}/r_{jk}$.
We denote the input, filter, and output degrees by $l_i$, $l_f$, and $l_o$, respectively.
A triple of nonnegative integer degrees $(l_i,l_f,l_o)$ is an \emph{admissible path} if $l_i,l_o\le l_{\max}$ and $|l_i-l_f|\le l_o\le l_i+l_f$.
Let $\calP(l_o)$ be the set of pairs $(l_i,l_f)$ admissible with output degree $l_o$.
The convolution is
\begin{equation}\label{eq:conv}
  \mathrm{Conv}(h)^{(l_o)}_{j,\cdot,c_o}
  =\sum_{(l_i,l_f)\in\calP(l_o)}\ \sum_{c_i}\ \sum_{k\in\calN(j)}
    R^{(l_o,l_f,l_i)}_{c_oc_i}(r_{jk})\,
    \bigl(Y^{(l_f)}(\hat{\bm r}_{jk})\otimes h^{(l_i)}_{k,\cdot,c_i}\bigr)^{(l_o)}.
\end{equation}
The filter degree $l_f$ may be bounded separately from the maximal feature degree $l_{\max}$.
For fixed positions, the convolution is linear in $h$.

The \emph{radial functions} vanish for $r\ge r_c$ and have the form
\begin{equation}\label{eq:radial-form}
  R^{(l_o,l_f,l_i)}_{c_oc_i}(r)
  =\chi(r)\,p^{(l_o,l_f,l_i)}_{c_oc_i}(r),
  \qquad 0\le r\le r_c .
\end{equation}
The \emph{cutoff envelope} $\chi:[0,\infty)\to[0,\infty)$ is fixed and continuous, with $\chi(r)>0$ for $0\le r<r_c$ and $\chi(r)=0$ for $r\ge r_c$.
Each path and channel pair has its own continuous \emph{profile} $p^{(l_o,l_f,l_i)}_{c_oc_i}:[0,r_c]\to\RR$, while all paths share the envelope.
Thus the contribution of a neighbor vanishes continuously as it reaches the cutoff.

Restricting~\cref{eq:conv} to the path $(0,0,0)$ gives a sum of scalar features weighted by functions of distance.
Filters with $l_f>0$ also convey bond directions.

\subsection{Bilinear networks and the BREMP}
\label{sec:bremp}

We approximate an atomic energy $\varepsilon\in\calF^0_{R_c}$ by a member of a parametric family $\{\varepsilon_\theta\}_\theta$.
The BREMP operates in two steps.
A stack of $T$ message-passing layers of cutoff $r_c$ assigns to every atom a feature that rotates with the environment.
Successive layers extend its dependence on the configuration over the neighbor graph, as described in \cref{def:l-hop}.
We call the map from an atomic environment to the final BREMP feature the \emph{equivariant representation}.
A bilinear readout then nonlinearly combines the components of the final feature to produce a rotation-invariant atomic energy.
We first define the blocks and network used in the readout, then combine them with the convolutions of \cref{sec:blocks}.

\begin{definition}[Bilinear block]
\label{def:bb}
Let $\calH=\bigoplus_{l=0}^{\lambda}V_l\otimes\RR^{C_{\mathrm H}}$ be a feature space with maximal degree $\lambda\ge0$ and width $C_{\mathrm H}\ge1$, called the hidden space.
A \emph{bilinear block} (BB) is a map $\calB:\calH\to\calH$ that forms factors $u=L_1x$ and $v=L_2x$ using equivariant affine maps $L_1,L_2:\calH\to\calH$ of the form \cref{eq:linear}.
Their Clebsch--Gordan products are mixed within each output degree,
\begin{equation}\label{eq:bb-product}
  t^{(l)}_{m,c}
  =\sum_{\substack{0\le l_1,l_2\le\lambda\\ |l_1-l_2|\le l\le l_1+l_2}}\ \sum_{c_1,c_2=1}^{C_{\mathrm H}}
    U^{(l_1l_2l)}_{c,c_1c_2}\,
    \bigl(u^{(l_1)}_{\cdot,c_1}\otimes v^{(l_2)}_{\cdot,c_2}\bigr)^{(l)}_m ,
  \qquad 0\le l\le\lambda ,
\end{equation}
where $1\le c\le C_{\mathrm H}$ and there is one coefficient tensor $U^{(l_1l_2l)}$ per coupling path.
The output is
\begin{equation}\label{eq:bb-out}
  \calB(x)=L_3x+L_4t ,
\end{equation}
with equivariant affine maps $L_3,L_4$ on $\calH$.
\end{definition}

A BB is a polynomial map of degree at most two.
Composing such blocks allows higher-order products of the input features.

\begin{definition}[Bilinear network]
\label{def:bn}
A \emph{bilinear network} (BN) from a feature space $\calV$ to a feature space $\calW$ is a composition
\begin{equation}\label{eq:bn}
  \mathrm{BN}=\pi\circ\calB_{K_{\mathrm{BN}}}\circ\cdots\circ\calB_1\circ\iota:\ \calV\to\calW ,
\end{equation}
of $K_{\mathrm{BN}}\ge1$ bilinear blocks on a common hidden space $\calH$, with equivariant affine maps $\iota:\calV\to\calH$ and $\pi:\calH\to\calW$.
The hidden maximal degree $\lambda$, width $C_{\mathrm H}$, and depth $K_{\mathrm{BN}}$ are its size parameters.
\end{definition}

Consequently, a BN with $K_{\mathrm{BN}}$ blocks has polynomial degree at most $2^{K_{\mathrm{BN}}}$.
Combining this readout with the equivariant convolutions of \cref{sec:blocks} gives the full BREMP architecture.

\begin{definition}[BREMP]
\label{def:bremp}
A BREMP with cutoff $r_c$, $T\ge1$ layers, and feature space $\calV$ computes the atomic energies of a configuration by
\begin{align}
  h^{[0],(0)}_{j,\cdot,c}&=\mathrm{Embed}_c(z_j),\qquad h^{[0],(l)}_{j}=0\ \text{for }l\ge1, \label{eq:embed}\\
  h^{[t+1]}_j&=h^{[t]}_j+L^{[t]}_{\mathrm{msg}}\,\mathrm{Conv}^{[t]}(h^{[t]})_j ,\qquad t=0,\ldots,T-1, \label{eq:layer}\\
  \varepsilon_\theta(\calD^{[r_c,T]}_j)&=\bigl[\mathrm{BN}(h^{[T]}_j)\bigr]^{(0)}_{0,1}. \label{eq:readout}
\end{align}
Here $\mathrm{Embed}:\calZ\to\RR^{C}$ is a learnable table, $\mathrm{Conv}^{[t]}$ is an equivariant convolution of the form \cref{eq:conv,eq:radial-form}, $L^{[t]}_{\mathrm{msg}}$ is an equivariant affine map of the form \cref{eq:linear} into $\calV$, and $\mathrm{BN}$ is a bilinear network of \cref{def:bn} whose output space contains a degree-$0$ channel.
All parameters, collected in $\theta$, are shared across atoms.
\end{definition}

\Cref{eq:embed,eq:layer,eq:readout} specify the type embedding, message-passing updates, and energy readout, respectively.
In these equations, superscripts $[t]$ index layers and superscripts $(l)$ index representation degrees.
For fixed positions, the message-passing layers are affine in the features.
The bilinear readout forms products of these features.

The final feature depends only on the $T$-hop environment $\calD^{[r_c,T]}_j$, giving the representation map
\begin{equation}\label{eq:rep-map}
  \Phi^{[T]}_\theta:\ \calE^{[r_c,T]}\to\calV,\qquad \calD^{[r_c,T]}_j\mapsto h^{[T]}_j .
\end{equation}
\Cref{eq:readout} factors the atomic energy as $\varepsilon_\theta=\rho\circ\mathrm{BN}\circ\Phi^{[T]}_\theta$ with $\rho$ the coordinate projection onto the designated channel.
The symmetries of the representation and atomic energy follow from those of the convolutions and bilinear readout.

\begin{proposition}
\label{prop:bremp-symmetry}
The convolution \cref{eq:conv} is equivariant under simultaneous rotations of positions and features and under relabeling of atoms.
It is invariant under translations of the positions.
Every bilinear block and bilinear network is $SO(3)$-equivariant.
For every parameter $\theta$, the BREMP representation is invariant under translations and satisfies
\begin{equation}\label{eq:rep-equivariance}
  \Phi^{[T]}_\theta\bigl((R,\sigma)\cdot\calD\bigr)=D_{\calV}(R)\,\Phi^{[T]}_\theta(\calD),
  \qquad (R,\sigma)\in G,\quad \calD\in\calE^{[r_c,T]}.
\end{equation}
The atomic energy $\varepsilon_\theta$ is invariant under translations, rotations, and type-preserving permutations of the neighbors.
\end{proposition}

\begin{proof}
Spherical harmonics and Clebsch--Gordan couplings are equivariant, while radial functions are rotation invariant.
Channel mixing, sums, and composition therefore give rotation equivariance of the convolution, bilinear blocks, and bilinear networks.
Starting from the scalar type embedding, induction through \cref{eq:layer} gives rotation equivariance of $\Phi^{[T]}_\theta$.
Shared atomwise maps and neighbor sums commute with relabeling, and all position dependence is through displacements.
This proves \cref{eq:rep-equivariance} and translation invariance.
The equivariant BN readout preserves these symmetries, and selecting its degree-$0$ output gives a rotation-invariant atomic energy.
\end{proof}

Within a stratum of $\calE^{[r_c,T]}$, the $T$-hop set has fixed atom types and size, but its internal neighbor graph can change as atoms move.
The cutoff envelope in \cref{eq:radial-form} ensures continuity when edges appear or disappear.

\begin{lemma}
\label{lem:rep-continuous}
For every parameter $\theta$, the map $\Phi^{[T]}_\theta$ is continuous on every stratum of $\calE^{[r_c,T]}$.
\end{lemma}

\begin{proof}
Fix a stratum and treat the atoms of the $T$-hop environment, including the center, as a configuration by themselves.
This leaves the final center feature unchanged.
Write \cref{eq:conv} as a sum over all other atoms of this configuration, with each term set to zero for $r_{jk}\ge r_c$.
Each product of a radial function and a spherical harmonic is continuous away from collisions, including at $r_{jk}=r_c$, where $\chi$ vanishes and the profile and harmonic remain bounded.
Starting from the position-independent embedding, each layer combines continuous features and filters through bilinear couplings, finite sums, and affine maps.
Induction over the layers proves continuity of the final center feature.
\end{proof}

\begin{remark}[Reflections]
\label{rem:reflections}
The BREMP guarantees rotation invariance of the atomic energy, but does not enforce reflection invariance.
Exact reflection invariance can be enforced by assigning each channel a \emph{parity} $p\in\{+1,-1\}$ under inversion $\bm r\mapsto-\bm r$.
Even channels ($p=+1$) remain unchanged, while odd channels ($p=-1$) change sign.
The harmonics $Y^{(l)}$ have parity $(-1)^l$, and a Clebsch--Gordan product of parities $p_1,p_2$ has parity $p_1p_2$.
Linear maps mix only channels with the same $(l,p)$, the type embedding and biases occupy only even degree-$0$ channels, and the energy is selected from an even degree-$0$ output.
These rules give $O(3)$-equivariant layers and reflection-invariant energies.
\end{remark}

\subsection{Completeness and universal approximation}
\label{sec:completeness-uat}

We now state the two properties used in the approximation argument.
Completeness concerns the information retained by the equivariant representation, and universal approximation concerns the attainable accuracy of the atomic energy.

\begin{definition}[Equivariant completeness]
\label{def:complete}
Let $A\subseteq\calE^{[r_c,T]}$ be invariant under $G$.
A $G$-equivariant map $\Phi:\calE^{[r_c,T]}\to\calV$ is \emph{complete on $A$} if, for all $\calD,\calD'\in A$ and $R\in SO(3)$,
\begin{equation}\label{eq:complete}
  \Phi(\calD)=D_{\calV}(R)\,\Phi(\calD')
  \quad\Longrightarrow\quad
  \calD=(R,\pi)\cdot\calD'\ \text{for some }\pi\in\Gamma .
\end{equation}
\end{definition}

Equivariance gives the reverse implication in \cref{eq:complete}.
Completeness thus separates $G$-orbits and additionally requires the same rotation to relate the features and the environments.
Taking $R=I$ shows that equal features determine the environment up to a type-preserving permutation.

To compare the network energy with a physical target, the $T$-hop input must contain the entire environment at range $R_c$, as required by \cref{eq:physical-range-coverage}.
We use the notion of universal approximation from~\cite{hgnn}, stated here for such inputs.

\begin{definition}[Universal approximation]
\label{def:uat}
Let $A\subseteq\calE^{[r_c,T]}$ be invariant under $G$ and consist of environments satisfying $\calD^{(R_c)}_j\subseteq\calD^{[r_c,T]}_j$.
A family $\{\varepsilon_\theta\}_\theta$ of continuous functions on $\calE^{[r_c,T]}$ is \emph{universal} for $\calF^0_{R_c}$ on $A$ if, for every target $\varepsilon\in\calF^0_{R_c}$, every compact $K\subseteq A$, and every tolerance $\delta>0$, there is a parameter $\theta$ such that
\begin{equation}\label{eq:uat-def}
  \sup_{\calD^{[r_c,T]}_j\in K}
  \bigl|\varepsilon_\theta(\calD^{[r_c,T]}_j)-\varepsilon(\calD^{(R_c)}_j)\bigr|<\delta .
\end{equation}
\end{definition}

Compactness is understood in the disjoint union topology of the environment strata described in \cref{def:env-space,def:l-hop}.
For $T=1$ and $r_c=R_c$, the input is the physical environment itself, and \cref{eq:uat-def} reduces to $\sup_K|\varepsilon_\theta-\varepsilon|<\delta$.

Completeness is a property of one fixed map $\Phi$ on the whole class $A$.
In the approximation family, the coefficients and any unrestricted network sizes may depend on the target $\varepsilon$, the compact set $K$, and the tolerance $\delta$.

\subsection{Wigner bilinear blocks}
\label{sec:wbb}

To connect the BREMP readout with DPA4, we now describe the Wigner construction of bilinear interactions.
It maps features to functions on $SO(3)$, multiplies these functions pointwise, and extracts coefficients from the products.
The product identity below relates this construction to the Clebsch--Gordan couplings in \cref{def:bb}.

For $g\in SO(3)$, let $D^{(l)}_{mk}(g)$, $|m|,|k|\le l$, denote the entries of the real Wigner matrix introduced in \cref{sec:blocks}.
Viewed as functions of $g$, these entries are the \emph{Wigner functions}.
The index $m$ is the representation index of the feature, and we call the column index $k$ the \emph{frame index}.
With respect to the normalized Haar measure $dg$ on $SO(3)$, they satisfy the orthogonality relation
\begin{equation}\label{eq:wigner-orth}
  (2l+1)\int_{SO(3)}D^{(l)}_{mk}(g)\,D^{(l')}_{m'k'}(g)\,dg=\delta_{ll'}\delta_{mm'}\delta_{kk'}.
\end{equation}
Their products satisfy
\begin{equation}\label{eq:wigner-product}
  D^{(l_1)}_{m_1k_1}(g)\,D^{(l_2)}_{m_2k_2}(g)
  =\sum_{L=|l_1-l_2|}^{l_1+l_2}\ \sum_{|M|,|K|\le L}
    C^{(L,M)}_{(l_1,m_1)(l_2,m_2)}\,C^{(L,K)}_{(l_1,k_1)(l_2,k_2)}\,D^{(L)}_{MK}(g).
\end{equation}
Both identities hold in the real coordinates and with the Clebsch--Gordan normalization fixed in \cref{sec:blocks}~\cite{varshalovich1988}.
The first Clebsch--Gordan coefficient in \cref{eq:wigner-product} couples the representation indices, as in \cref{eq:bb-product}, and the second couples the frame indices.
This identity connects pointwise multiplication of functions to bilinear coupling of features.

\begin{definition}[Wigner bilinear block]
\label{def:wbb}
Let $\calH$ be a hidden space of maximal degree $\lambda$ and width $C_{\mathrm H}$ as in \cref{def:bb}.
For each $0\le l\le\lambda$, choose a \emph{frame set} $\calS_l\subseteq\{-l,\ldots,l\}$, with $\calS_0=\{0\}$.
For an integer $F\ge1$, a \emph{Wigner bilinear block} (WBB) maps $x\in\calH$ to $y\in\calH$ using $F$ pairs of real-valued functions on $SO(3)$, called fields and indexed by $a=1,\ldots,F$.
The \emph{lift} mixes input channels separately for each degree, frame, and field,
\begin{equation}\label{eq:wbb-lift}
  u^{(l)}_{m,k,a}=\sum_{c}A^{(l)}_{ka,c}\,x^{(l)}_{m,c},
  \qquad
  v^{(l)}_{m,k,a}=\sum_{c}B^{(l)}_{ka,c}\,x^{(l)}_{m,c},
  \qquad k\in\calS_l .
\end{equation}
The \emph{synthesis} forms the fields $\hat u_a,\hat v_a:SO(3)\to\RR$,
\begin{equation}\label{eq:wbb-synth}
  \hat u_a(g)=\sum_{l,m,k}u^{(l)}_{m,k,a}\,D^{(l)}_{mk}(g),
  \qquad
  \hat v_a(g)=\sum_{l,m,k}v^{(l)}_{m,k,a}\,D^{(l)}_{mk}(g),
\end{equation}
where both sums run over $0\le l\le\lambda$, $|m|\le l$, and $k\in\calS_l$.
The \emph{analysis} extracts the retained coefficients of the pointwise product $\hat z_a(g)=\hat u_a(g)\hat v_a(g)$,
\begin{equation}\label{eq:wbb-analysis}
  z^{(l)}_{m,k,a}=(2l+1)\int_{SO(3)}D^{(l)}_{mk}(g)\,\hat z_a(g)\,dg,
  \qquad 0\le l\le\lambda,\ k\in\calS_l .
\end{equation}
The output combines these coefficients with an affine map of the input,
\begin{equation}\label{eq:wbb-out}
  y^{(l)}_{m,d}=\sum_{c}P^{(l)}_{dc}\,x^{(l)}_{m,c}+\sum_{k\in\calS_l}\sum_{a=1}^{F}Q^{(l)}_{d,ka}\,z^{(l)}_{m,k,a}+\delta_{l0}\,b_d .
\end{equation}
The arrays $A$, $B$, $P$, and $Q$ and the degree-$0$ bias $b$ are real parameters.
The channel indices $c$ and $d$ range from $1$ to $C_{\mathrm H}$.
\end{definition}

When the input is rotated by $R$, each synthesized field $\hat u_a(g)$ becomes $\hat u_a(R^{-1}g)$, and the same holds for $\hat v_a$.
Pointwise multiplication preserves this transformation rule, and the analyzed coefficients rotate in their respective spaces $V_l$.
The WBB is therefore equivariant for every choice of frame sets.

For an integer $k_{\max}\ge0$, \emph{frame truncation} means taking $\calS_l=\{k\in\mathbb Z:|k|\le\min(k_{\max},l)\}$.
The block has \emph{full frames} when $\calS_l=\{-l,\ldots,l\}$ for every $l\le\lambda$.
The choice $k_{\max}=0$ gives functions on $S^2$, since $D^{(l)}_{m0}(g)$ is proportional to $Y^{(l)}_m(g\bm e_z)$, where $\bm e_z=(0,0,1)$.
Thus the fields in this case depend only on the direction of the rotated pole.

The synthesized fields have \emph{band limit} at most $\lambda$, meaning that no Wigner degree above $\lambda$ occurs.
Their products $\hat z_a$ can contain degrees up to $2\lambda$.
By \cref{eq:wigner-orth}, the analysis retains precisely their coefficients in degrees $l\le\lambda$ and frames $k\in\calS_l$.
The integrand of \cref{eq:wbb-analysis} is a product of $\hat z_a$ with a Wigner function of degree at most $\lambda$, so its band limit is at most $3\lambda$.
For frame truncation $k_{\max}$, these integrals are evaluated exactly by a sphere rule exact for polynomials of degree at most $3\lambda$, combined with $3k_{\max}+1$ equally spaced rotations about the pole.
Appendix~\ref{app:wbb-quadrature} gives the construction and proves exactness.

\subsection{The DPA4 architecture}
\label{sec:dpa4}

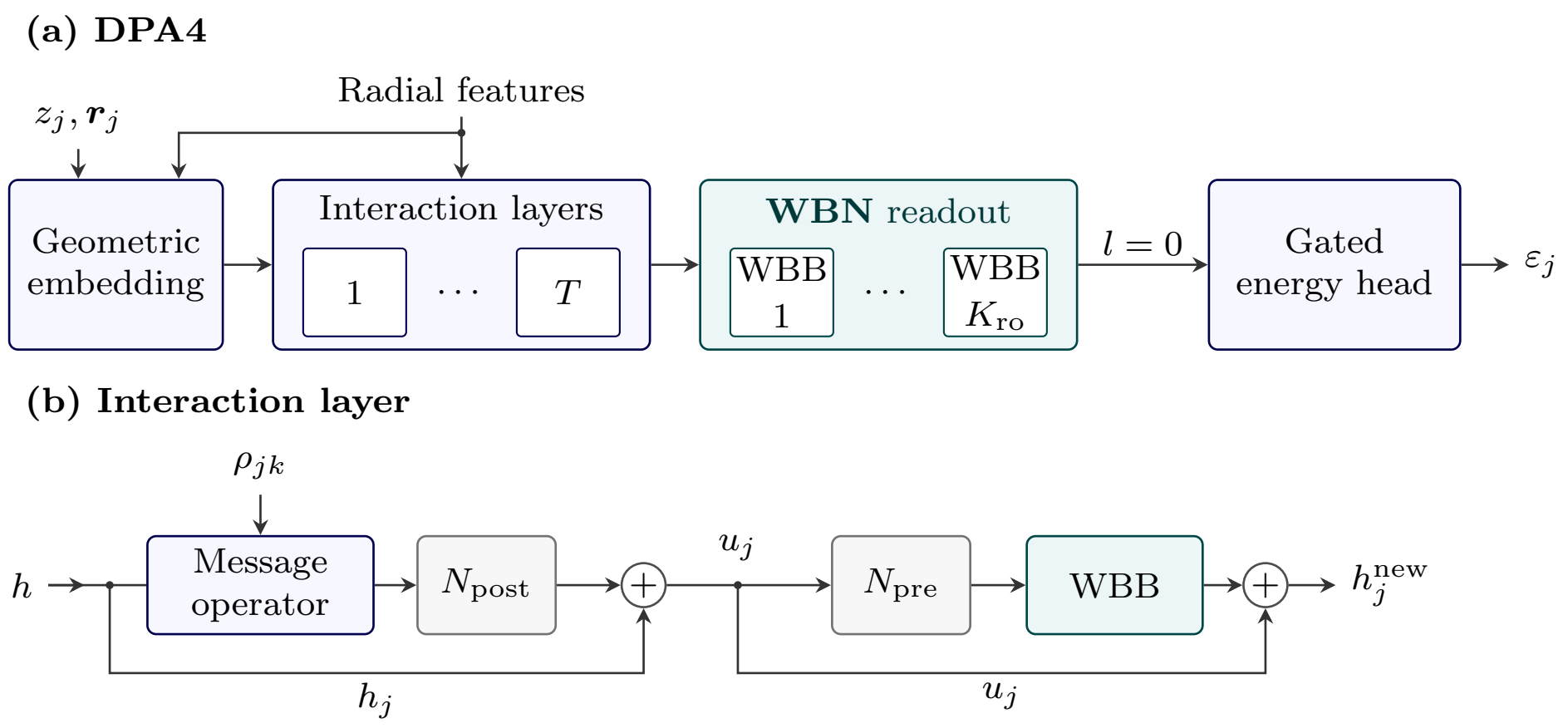
\begin{figure}[!t]
  \centering
  \begin{tikzpicture}[
  font=\footnotesize,
  stage/.style={draw=blue!30!black,fill=blue!3,rounded corners=2pt,
    line width=0.5pt,align=center,inner sep=2.5pt,minimum height=1.35cm},
  wstage/.style={stage,draw=teal!55!black,fill=teal!7},
  block/.style={draw=blue!30!black,fill=blue!3,rounded corners=2pt,
    line width=0.5pt,align=center,inner sep=2.5pt,minimum height=0.78cm},
  norm/.style={block,draw=black!55,fill=black!3},
  wbb/.style={block,draw=teal!55!black,fill=teal!7},
  cell/.style={draw=blue!30!black,fill=white,rounded corners=1pt,
    line width=0.4pt,align=center,text width=0.74cm,inner sep=1pt,
    minimum width=0.82cm,minimum height=0.7cm},
  wcell/.style={cell,draw=teal!55!black},
  flow/.style={-{Stealth[length=3.5pt,width=3.5pt]},line width=0.5pt,
    draw=black!80},
  wire/.style={line width=0.5pt,draw=black!80},
  sum/.style={circle,draw=black!70,fill=white,line width=0.5pt,
    inner sep=0pt,minimum size=0.35cm},
  title/.style={anchor=west,font=\footnotesize\bfseries}
]
  \node[title] at (0,1.85) {(a) DPA4};
  \node[stage,text width=1.5cm,minimum width=1.7cm] (embed) at (0.85,0)
    {Geometric\\embedding};
  \node[stage,minimum width=3cm] (layers) at (3.6,0) {};
  \node[wstage,minimum width=3cm] (readout) at (7,0) {};
  \node[stage,text width=1.8cm,minimum width=2cm] (head) at (10.55,0)
    {Gated\\energy head};
  \node at (3.6,0.43) {Interaction layers};
  \node[text=teal!45!black] at (7,0.43) {\textbf{WBN} readout};
  \node[cell] at (2.75,-0.22) {$1$};
  \node[cell] at (4.45,-0.22) {$T$};
  \node[wcell] at (6.15,-0.22) {WBB\\[1pt]$1$};
  \node[wcell] at (7.85,-0.22) {WBB\\[1pt]$K_{\mathrm{ro}}$};
  \node at (3.6,-0.22) {$\cdots$};
  \node at (7,-0.22) {$\cdots$};

  \node (input) at (0.55,1.15) {$z_j,\bm r_j$};
  \node (radial) at (3.6,1.4) {Radial features};
  \draw[flow] (input.south) -- ($(embed.north)+(-0.3,0)$);
  \draw[wire] (radial.south) -- (3.6,1.05);
  \draw[flow] (3.6,1.05) -- (layers.north);
  \draw[flow] (3.6,1.05) -- (1.35,1.05) -- ($(embed.north)+(0.5,0)$);
  \fill[black!80] (3.6,1.05) circle (0.9pt);
  \draw[flow] (embed) -- (layers);
  \draw[flow] (layers) -- (readout);
  \draw[flow] (readout) -- node[above,inner sep=2pt] {$l=0$} (head);
  \node (energy) at (12.2,0) {$\varepsilon_j$};
  \draw[flow] (head.east) -- (energy.west);

  \begin{scope}[yshift=0.15cm]
  \node[title] at (0,-1.25) {(b) Interaction layer};
  \node (h) at (0.1,-2.7) {$h$};
  \coordinate (split) at (0.8,-2.7);
  \node[block,text width=1.6cm,minimum width=1.8cm] (message) at (2,-2.7)
    {Message\\operator};
  \node[norm,minimum width=1.1cm] (post) at (3.8,-2.7) {$N_{\mathrm{post}}$};
  \node[sum] (sum1) at (5.05,-2.7) {$+$};
  \coordinate (u) at (5.8,-2.7);
  \node[norm,minimum width=1.1cm] (pre) at (7.1,-2.7) {$N_{\mathrm{pre}}$};
  \node[wbb,minimum width=1.4cm] (product) at (8.8,-2.7) {WBB};
  \node[sum] (sum2) at (10,-2.7) {$+$};
  \node (output) at (11,-2.7) {$h_j^{\mathrm{new}}$};
  \node (rho) at (2,-1.75) {$\rho_{jk}$};
  \draw[wire] (h.east) -- (message.west);
  \draw[flow] (h.east) -- (0.6,-2.7);
  \fill[black!80] (split) circle (0.9pt);
  \draw[flow] (rho.south) -- (message.north);
  \draw[flow] (message) -- (post);
  \draw[flow] (post) -- (sum1);
  \draw[flow] (sum1) -- (pre);
  \draw[flow] (pre) -- (product);
  \draw[flow] (product) -- (sum2);
  \draw[flow] (sum2) -- (output);
  \fill[black!80] (u) circle (0.9pt);
  \node[above=2pt] at (u) {$u_j$};
  \draw[flow] (split) -- (0.8,-3.4)
    -- node[below,inner sep=2pt] {$h_j$} (5.05,-3.4) -- (sum1.south);
  \draw[flow] (u) -- (5.8,-3.4)
    -- node[below,inner sep=2pt] {$u_j$} (10,-3.4) -- (sum2.south);
  \end{scope}
\end{tikzpicture}
  \caption{Schematic of the DPA4 architecture.
  (a) Overall architecture.
  (b) Structure of an interaction layer.}
  \label{fig:dpa4}
\end{figure}

DPA4~\cite{dpa4} is an equivariant message-passing network with a \emph{Wigner bilinear network} (WBN) readout based on the WBB construction of \cref{sec:wbb}.
Throughout the network, atomic features $h_j$ lie in the space $\calV$ of \cref{eq:feature-space}, with fixed maximal degree $l_{\max}$ and channel width $C$.
Each interaction layer combines neighbor aggregation with WBB operations at the same atom.
\Cref{fig:dpa4} shows the embedding, message-passing layers, and readout.
Appendix~\ref{app:dpa4} gives the operator formulas used in \cref{sec:uat-dpa4}.

The embedding maps atomic types and local geometry to the initial equivariant atomic features for message passing.
Radial features $\rho_{jk}$ are computed from bond distances and atomic types and shared by all interaction layers.
The formulas for both are given in \cref{eq:app-dpa4-embed,eq:app-dpa4-embed-l,eq:app-dpa4-rho}.

Each interaction layer updates the atomic features in two residual steps,
\begin{equation}\label{eq:dpa4-layer}
  \begin{aligned}
    u_j&=h_j+N_{\mathrm{post}}\bigl(\calC(h)_j\bigr),\\
    h_j^{\mathrm{new}}&=u_j+\calB_{\mathrm W}\bigl(N_{\mathrm{pre}}(u_j),N_{\mathrm{pre}}(u_j)\bigr).
  \end{aligned}
\end{equation}
Here $\calC$ is an equivariant message operator, and $\calB_{\mathrm W}$ is a scalar-gated WBB acting on two features at the same atom.
The maps $N_{\mathrm{post}}$ and $N_{\mathrm{pre}}$ normalize the features on the residual branches while preserving rotation equivariance, as shown in \cref{fig:dpa4}(b).

The operator $\calC$ performs equivariant message passing.
It constructs messages using $SO(2)$ convolutions and equivariant nonlinearities in frames aligned with the bonds, where $SO(2)$ acts by rotations about the bond axis.
The convolutions use the radial features $\rho_{jk}$ and retain orders $|m|\le m_{\max}$.
After rotation back to the global frame, invariant attention aggregates the messages, and the WBB $\calB_{\mathrm W}$ couples the aggregate with the center feature, as detailed in \cref{eq:app-dpa4-conv}.
With full order $m_{\max}=l_{\max}$, the bond-frame $SO(2)$ convolutions are equivalent to the $SO(3)$ convolution \cref{eq:conv}~\cite{passaro2023reducing}.
They therefore supply the BREMP convolutional substructure within $\calC$, which is used in the layer approximation of \cref{prop:dpa4-layer}.

The block $\calB_{\mathrm W}$ couples features within $\calC$ and in the second residual step of \cref{eq:dpa4-layer}.
It lifts its two inputs and applies the synthesis, pointwise multiplication, and analysis of \cref{def:wbb}.
The analyzed coefficients are mixed, multiplied by invariant scalar gates, and combined with a nonlinear degree-$0$ term before an equivariant linear map.

After $T$ interaction layers, the WBN readout applies $K_{\mathrm{ro}}\ge1$ residual WBBs $h_j\mapsto h_j+\calB_{\mathrm W}(h_j,h_j)$ without normalization.
Only then is the degree-$0$ slice $s_j\in\RR^{C}$ taken and passed to the energy head, which uses a multilayer perceptron (MLP) with one gated hidden layer,
\begin{equation}\label{eq:dpa4-energy}
  \varepsilon_j=\bm w^{\top}\mathrm{MLP}(s_j)+b_{z_j},
  \qquad E=\sum_j\varepsilon_j .
\end{equation}
Here $\bm w$ is a learned output vector, and $b_z$ is a fixed constant for each atomic type.
\FloatBarrier

%%%%%%%%%%%%%%%%%%%%%%%%%%%%%%%%%%%%%%%%%%%%%%%%%%%%%%%%%%%%%%%%%%
\section{Completeness of the equivariant representation}
\label{sec:complete}
%%%%%%%%%%%%%%%%%%%%%%%%%%%%%%%%%%%%%%%%%%%%%%%%%%%%%%%%%%%%%%%%%%
We construct finite-width BREMP representations that are complete in the sense of \cref{def:complete}.
For one layer, radial and angular moments recover the typed neighbor positions.
For multiple layers, each convolution also recovers the features attached to those positions, allowing the center to reconstruct one further hop of its environment.

\subsection{One layer}
\label{sec:complete-one}
Throughout this subsection, $T=1$ and $r_c=R_c$, so the network input is the physical environment and $\calE^{[R_c,1]}=\calE^{(R_c)}$.
We use the one-hot embedding $\mathrm{Embed}_z(z')=\delta_{zz'}$ for $z,z'\in\calZ$ in a block of $Z$ channels, with all other channels initially zero.
The feature after one layer is $h^{[1]}_j=h^{[0]}_j+L_{\mathrm{msg}}\mathrm{Conv}(h^{[0]})_j$.

Since $h^{[0]}$ has degree-$0$ components only, the triangle condition in \cref{eq:conv} forces $l_f=l_o=l$.
By uniqueness of the Clebsch--Gordan coupling, $(v\otimes s)^{(l)}=a_lsv$ for $v\in V_l$ and $s\in V_0$, where $a_l\ne0$ depends only on $l$.
Using the neighbor measures of \cref{eq:neighbor-measure}, the first-layer message therefore becomes
\begin{equation}\label{eq:B}
  B^{(l)}_{j,\cdot,c}=\mathrm{Conv}(h^{[0]})^{(l)}_{j,\cdot,c}
  =a_l\sum_{z'\in\calZ}\int R^{(l,l,0)}_{cz'}(|\bm x|)\,Y^{(l)}(\hat{\bm x})\,dn_{z'}(\calD_j)(\bm x).
\end{equation}
Each channel is a moment against a radial function times a spherical harmonic.
The next lemma selects finitely many such moments that determine every neighbor measure on $\kappa$-generic environments.

\begin{lemma}
\label{lem:shell-moments}
Fix an integer $\kappa\ge1$ and let $l_{\max}\ge\kappa$.
With $M$ as in \cref{sec:setting-pes}, use $2ZM$ message channels indexed by $c=(z,q)$, where $z\in\calZ$ and $q\in\{0,\ldots,2M-1\}$, and choose the profiles
\begin{equation}\label{eq:radial-choice}
  p^{(l,l,0)}_{(z,q),z'}(r)=a_l^{-1}\delta_{zz'}\,r^{\,l+2q},\qquad 0\le l\le\kappa,
\end{equation}
with all other paths and all degrees $l>\kappa$ set to zero.
If $\calD,\calD'\in\calE^{(R_c)}_\kappa$ satisfy $B(\calD)=B(\calD')$, then $n_z(\calD)=n_z(\calD')$ for every type $z$.
\end{lemma}

\begin{proof}
Suppose $B(\calD)=B(\calD')$ and fix a type $z$.
Write $\mu=n_z(\calD)=\sum_{a=1}^{n}\delta_{\bm x_a}$ and $\mu'=n_z(\calD')$, where the $\bm x_a$ are the displacements of the neighbors of type $z$, $n\le M$, and $r_a=|\bm x_a|$.
By \cref{eq:B,eq:radial-form,eq:radial-choice}, the corresponding message channels are
\begin{equation}\label{eq:moments}
  b^{(l)}_{q}=\sum_{a=1}^{n}\chi(r_a)\,r_a^{\,l+2q}\,Y^{(l)}(\hat{\bm x}_a),\qquad 0\le l\le\kappa,\ 0\le q\le 2M-1.
\end{equation}
The corresponding moments of $\mu'$ agree with these by hypothesis.
We show that the shared moments determine the shell radii and populations, then the angular moments of each shell, and finally its points.
Since $Y^{(0)}$ is a nonzero constant and $\chi$ is positive inside the cutoff, their common moment $b^{(0)}_0$ vanishes exactly when both measures are empty.
We may therefore assume that both measures are nonempty.

First, let $0<\rho_1<\cdots<\rho_S<R_c$ be the distinct shell radii and $n_s$ the population of shell $s$.
Set $t_s=\rho_s^2$ and $w_s=n_s\chi(\rho_s)>0$.
After division by $Y^{(0)}$, the degree-$0$ channels give the moments $\sum_s w_st_s^q$, $q=0,\ldots,2M-1$, of $\nu=\sum_s w_s\delta_{t_s}$.
The analogous radial measure $\nu'$ for $\mu'$ has the same moments.
Each radial measure has at most $M$ support points, so their difference is $\sum_{i=1}^{J}d_i\delta_{u_i}$ at $J\le2M$ distinct points.
The available moments imply $\sum_i d_iu_i^q=0$ for $q=0,\ldots,J-1$.
The Vandermonde matrix is invertible, so every $d_i$ vanishes and $\nu=\nu'$.
Consequently, $\mu$ and $\mu'$ have the same shell radii $\rho_s$ and populations $n_s=w_s/\chi(\rho_s)$.

Second, fix $l\le\kappa$ and $|m|\le l$, and write $A^{(l)}_{s,m}=\sum_{a:\,r_a=\rho_s}Y^{(l)}_m(\hat{\bm x}_a)$ for the angular moment of shell $s$.
\Cref{eq:moments} gives $b^{(l)}_{q,m}=\sum_{s=1}^{S}\chi(\rho_s)\rho_s^{\,l}t_s^q A^{(l)}_{s,m}$.
Since $S\le M$, the moments with $q=0,\ldots,S-1$ are available.
Their coefficient matrix is a Vandermonde matrix in the distinct $t_s$, multiplied by a diagonal matrix with positive entries $\chi(\rho_s)\rho_s^{\,l}$.
Both measures have the same radii and moment values, so solving this system gives the same $A^{(l)}_{s,m}$ for every shell and every $l\le\kappa$.

Third, fix a shell of radius $\rho_s$ with $p=n_s$ points $\bm y_1,\ldots,\bm y_p$ of $\mu$.
For any $\bm v\in\RR^3$ and integer $1\le d\le p$, the restriction of $(\bm v\cdot\bm x)^d$ to this sphere is a linear combination of spherical harmonics of degrees at most $d$~\cite{varshalovich1988}.
The $\kappa$-genericity assumption gives $p\le\kappa$, so the shared angular moments determine the power sums $\sum_a(\bm v\cdot\bm y_a)^d$ for $d=1,\ldots,p$.
Newton's identities then determine the monic polynomial $\prod_{a=1}^{p}(t-\bm v\cdot\bm y_a)$.
Thus the points of $\mu$ and $\mu'$ on this shell have the same multiset of scalar projections for every $\bm v$.

To prove uniqueness, choose $\bm v$ such that $\bm v\cdot(\bm u-\bm u')\ne0$ for every pair of distinct points $\bm u,\bm u'$ in the union of the two shell point sets.
Such a vector exists because only finitely many planes through the origin are excluded.
The map $\bm x\mapsto\bm v\cdot\bm x$ is injective on that union, so equality of the projected multisets forces equality of the point sets.

The measures agree on every shell, hence $\mu=\mu'$.
Repeating the argument for each type $z$ proves the lemma.
\end{proof}

The message recovers the neighbors, while the retained embedding records the center type.
Together they give a complete representation.

\begin{proposition}[One-layer completeness]
\label{prop:one-layer-complete}
Fix an integer $\kappa\ge1$ and let $l_{\max}\ge\kappa$.
There exists a one-layer BREMP with cutoff $R_c$, maximal feature degree $l_{\max}$, and width $Z(2M+1)$ whose representation $\Phi^{[1]}_\theta$ is complete on $\calE^{(R_c)}_\kappa$.
\end{proposition}

\begin{proof}
Use the one-hot type embedding in $Z$ channels and initialize the remaining $2ZM$ channels to zero.
Choose the message $B$ as in \cref{lem:shell-moments}.
At every degree, take $L_{\mathrm{msg}}$ to be the coordinate inclusion $m\mapsto(0,m)$ into the initially zero block, with zero bias in \cref{eq:linear}.
The updated feature therefore consists of the center-type code and $B_j$ in separate blocks.
Let $\calD,\calD'\in\calE^{(R_c)}_\kappa$ and $R\in SO(3)$ satisfy $\Phi^{[1]}_\theta(\calD)=D_{\calV}(R)\,\Phi^{[1]}_\theta(\calD')$.
The degree-$0$ embedding blocks give equal center types.
By \cref{prop:bremp-symmetry}, equality of the message blocks gives $B(\calD)=B(R\cdot\calD')$.
The rotated environment remains $\kappa$-generic, so \cref{lem:shell-moments} gives $n_z(\calD)=n_z(R\cdot\calD')$ for every type $z$.
Equal center types and equal neighbor measures determine the same unordered environment.
Hence $\calD=(R,\pi)\cdot\calD'$ for a type-preserving permutation $\pi$, as required by \cref{def:complete}.
\end{proof}

\begin{remark}[Angular resolution]
\label{rem:band-necessity}
Consider two environments with the same center type, each having four neighbors of one common type at the vertices of a square or a regular tetrahedron with the same center and circumradius.
Their degree-$0$ moments agree and their degree-$1$ moments vanish, so no one-layer BREMP with $l_{\max}\le1$ can distinguish them.
Their degree-$2$ angular moments differ.
\end{remark}

\subsection{Multiple layers}
\label{sec:complete-multi}
Fix a layer cutoff $0<r_c\le R_c$.
We construct a $T$-layer representation that is complete on $T$-hop environments by induction on the depth.
If the representation after $t$ layers is complete, each neighbor's feature determines its $t$-hop environment in the global frame.
The next layer will recover both the displacements and the features of the one-hop neighbors.
The following lemma shows that these displacements and neighbor environments, together with the center type, determine the center's $(t+1)$-hop environment.

\begin{lemma}
\label{lem:hop-compose}
Fix an integer $t\ge0$ and an atom $j$.
The center type $z_j$ together with the set $\{(\Delta\bm r_{jk},\calD^{[r_c,t]}_k):k\in\calN(j)\}$ determines $\calD^{[r_c,t+1]}_j$.
\end{lemma}

\begin{proof}
The given data identify each one-hop neighbor $k$ by its displacement $\Delta\bm r_{jk}$ and the center type recorded in $\calD^{[r_c,t]}_k$.
For each such $k$ and each pair $(\Delta\bm r_{km},z_m)$ in its environment, compute
\begin{equation}\label{eq:hop-add}
  \Delta\bm r_{jm}=\Delta\bm r_{jk}+\Delta\bm r_{km}.
\end{equation}
Discard zero displacements, which correspond to $m=j$, and list each remaining position only once.
This removes repeated paths to the same atom, since distinct atoms occupy distinct positions.
Together with the one-hop neighbors and the center type $z_j$, these pairs give exactly $\calD^{[r_c,t+1]}_j$ by \cref{eq:l-hop-set,eq:l-hop-env}.
\end{proof}

We next construct a convolution that recovers the data required by \cref{lem:hop-compose}.
The one-hot channels first locate the neighbors type by type using \cref{lem:shell-moments}.
Additional channels then recover their features by polynomial interpolation on those positions.
This interpolation separates all neighbors on a shell together, which is why \cref{def:kappa-generic} counts neighbors of all types.

\begin{lemma}
\label{lem:layer-extend}
Fix integers $\kappa\ge1$ and $t,l_t\ge0$.
Suppose the $t$-layer representation $\Phi^{[t]}_\theta$ is complete on $\calE^{[r_c,t]}_\kappa$, has nonzero components only in degrees at most $l_t$, and retains the one-hot center-type code in degree-$0$ channels.
After adjoining zero components to enlarge its feature space, it can be extended by one layer of \cref{eq:layer} to a representation $\Phi^{[t+1]}_\theta$ that is complete on $\calE^{[r_c,t+1]}_\kappa$ and has nonzero components only in degrees at most $l_{t+1}=\max(\kappa,\,l_t+\kappa-1)$.
The extension preserves all input channels and uses polynomial profiles with filter degrees at most $\kappa$.
\end{lemma}

\begin{proof}
Fix a center $j$ and write $h_k=\Phi^{[t]}_\theta(\calD^{[r_c,t]}_k)$ for the input feature of each neighbor $k\in\calN(j)$.
Let $C_t$ be the number of input channels.
We choose moment channels that first recover the neighbor positions and then the features attached to them.

For every input degree $0\le l_i\le l_t$ and channel $1\le c_i\le C_t$, use filter degrees $0\le l_f\le\kappa-1$.
For the one-hot scalar channels, also include $l_i=0$ and $l_f=\kappa$.
For every such triple $(l_i,c_i,l_f)$ and radial index $q\in\{0,\ldots,2M-1\}$, reserve an output channel $c_o$.
Set $p^{(l_o,l_f,l_i)}_{c_oc_i}(r)=r^{\,l_f+2q}$ for every admissible $l_o$, and set all other profiles into that channel to zero.
By \cref{eq:conv}, the degree-$l_o$ output is the Clebsch--Gordan component of the tensor
\begin{equation}\label{eq:decorated-moment}
  \mu^{(l_i,c_i;\,l_f,q)}
  =\sum_{k\in\calN(j)}\chi(r_{jk})\,r_{jk}^{\,l_f+2q}\,Y^{(l_f)}(\hat{\bm r}_{jk})\otimes h^{(l_i)}_{k,\cdot,c_i}
  \in V_{l_f}\otimes V_{l_i}.
\end{equation}
The degree bound gives $l_i+l_f\le l_{t+1}$, so all components in the Clebsch--Gordan decomposition of $V_{l_f}\otimes V_{l_i}$ are retained.
The coupled outputs therefore determine the full tensor \cref{eq:decorated-moment}.

The tensors for the one-hot scalar channels give the moments \cref{eq:moments}.
Applying the recovery argument of \cref{lem:shell-moments} with cutoff $r_c$ in place of $R_c$ determines the typed neighbor positions $\{(\Delta\bm r_{jk},z_k):k\in\calN(j)\}$.

Let $\bm x_1,\ldots,\bm x_n$, with $n\le M$, be the recovered displacements.
For each $s\in\{1,\ldots,n\}$, we construct a scalar function that isolates $\bm x_s$.
Set $\rho_s=|\bm x_s|$ and choose a Lagrange polynomial $\ell_s(\tau)$ of degree at most $M-1$ that is $1$ at $\tau=\rho_s^2$ and $0$ at the squared radii of all other shells.
For every other point on the shell of radius $\rho_s$, choose an affine function that vanishes there and equals $1$ at $\bm x_s$.
Their product has degree at most $\kappa-1$, since there are at most $\kappa-1$ such points.
For a shell consisting only of $\bm x_s$, take this product to be $1$.
As in the harmonic decomposition used in \cref{lem:shell-moments}, its restriction to the sphere is a linear combination $\sum_{l\le\kappa-1}\sum_m a_{s,lm}Y^{(l)}_m(\hat{\bm x})$.
Define
\begin{equation}\label{eq:selector}
  p_s(\bm x)=\ell_s(|\bm x|^2)\sum_{l\le\kappa-1}\sum_{m}a_{s,lm}\,\Bigl(\frac{|\bm x|}{\rho_s}\Bigr)^{l}Y^{(l)}_m(\hat{\bm x}).
\end{equation}
This function equals $1$ at $\bm x_s$ and $0$ at every other recovered displacement.
It is a linear combination of $r^{\,l+2q}Y^{(l)}_m$ with $l\le\kappa-1$ and $q\le M-1$, all of which occur in the message.
Let $k_s\in\calN(j)$ be the neighbor with $\Delta\bm r_{jk_s}=\bm x_s$.
For every input degree $l_i$ and channel $c_i$, the corresponding linear combination of components of \cref{eq:decorated-moment} equals $\chi(\rho_s)h^{(l_i)}_{k_s,\cdot,c_i}$.
Since $0<\rho_s<r_c$, division by $\chi(\rho_s)>0$ recovers the full feature $h_{k_s}$.
The message therefore determines $\{(\Delta\bm r_{jk},h_k):k\in\calN(j)\}$.

At every degree, choose $L_{\mathrm{msg}}$ as the coordinate inclusion $m\mapsto(0,m)$ into the newly adjoined channels, with zero bias in \cref{eq:linear}.
The residual update concatenates the original feature $h_j$ and the convolution output without mixing their channels.
It thus preserves $h_j$ and its one-hot center-type code.
Let $\calD,\calD'\in\calE^{[r_c,t+1]}_\kappa$ and $R\in SO(3)$ satisfy $\Phi^{[t+1]}_\theta(\calD)=D_{\calV}(R)\,\Phi^{[t+1]}_\theta(\calD')$.
As in the proof of \cref{prop:one-layer-complete}, equivariance turns this relation into equality of the features of $\calD$ and $R\cdot\calD'$.
The retained one-hot code gives equal center types, and the message recovery gives the same neighbor positions and input features.
Completeness of $\Phi^{[t]}_\theta$, with $R=I$ in \cref{def:complete}, then gives the same $t$-hop environment at each matched neighbor in the global frame.
\Cref{lem:hop-compose} therefore gives $\calD=(R,\pi)\cdot\calD'$ for some $\pi\in\Gamma$, proving completeness.
\end{proof}

\begin{proposition}[Multi-layer completeness]
\label{thm:multi-complete}
Fix integers $\kappa,T\ge1$ and a cutoff $0<r_c\le R_c$.
There exists a finite-width $T$-layer BREMP with cutoff $r_c$, maximal feature degree $T(\kappa-1)+1$, and filter degrees at most $\kappa$ in every layer whose representation $\Phi^{[T]}_\theta$ is complete on $\calE^{[r_c,T]}_\kappa$.
\end{proposition}

\begin{proof}
Start with the one-hot embedding \cref{eq:embed}, which is complete on the $0$-hop environments because they consist only of the center type.
Its nonzero components have maximal degree $l_0=0$.
Successive applications of \cref{lem:layer-extend} preserve completeness and give $l_{t+1}=\max(\kappa,l_t+\kappa-1)$.
Thus $l_1=\kappa$ and $l_t=t(\kappa-1)+1$ for $1\le t\le T$.
Each extension adds finitely many channels.
All $T$ steps therefore fit in one finite-width feature space of maximal degree $T(\kappa-1)+1$.
Embed the intermediate feature spaces in this common space by adjoining zeros, and set the convolution profiles on unused input channels to zero.
The coordinate inclusions chosen in \cref{lem:layer-extend} then write each message into its own initially zero block, leaving all other blocks unchanged.
\end{proof}

\begin{remark}[Depth and degree]
\label{rem:depth-degree}
In \cref{thm:multi-complete}, the filter degree is bounded by $\kappa$ independently of $T$, while the feature degree bound $T(\kappa-1)+1$ increases with depth when $\kappa\ge2$.
For $\kappa=1$, degree $1$ suffices at every depth.

Reducing the layer cutoff can lower $\kappa$ by excluding shells of larger multiplicity, thereby reducing the filter degree used by the construction.
For targets of range $R_c$, the depth must also satisfy the coverage condition \cref{eq:physical-range-coverage}.
\end{remark}

%%%%%%%%%%%%%%%%%%%%%%%%%%%%%%%%%%%%%%%%%%%%%%%%%%%%%%%%%%%%%%%%%%
\section{The bilinear readout}
\label{sec:readout}
%%%%%%%%%%%%%%%%%%%%%%%%%%%%%%%%%%%%%%%%%%%%%%%%%%%%%%%%%%%%%%%%%%

The representations constructed in \cref{sec:complete} retain the atomic environment.
We now show that the BN of \cref{def:bn} can approximate any continuous equivariant map of the features, including the invariant functions needed for the energy readout.
We first realize polynomial equivariant maps exactly by iterated Clebsch--Gordan couplings, then approximate continuous equivariant maps by such polynomials.
Throughout this section, $\calV$ and $\calW$ are feature spaces of the form \cref{eq:feature-space}, and $G=SO(3)$ acts on them through $D_{\calV}$ and $D_{\calW}$.

To organize the polynomial computations, we use \emph{coupling trees}.
A leaf is an equivariant linear map $A:\calV\to V_l$, evaluated as $T(x)=Ax$.
Given trees $T_1,T_2$ with outputs in $V_{l_1},V_{l_2}$ and a degree $|l_1-l_2|\le l\le l_1+l_2$, their coupling forms a new tree $T=(T_1,T_2,l)$ with evaluation
\begin{equation}\label{eq:tree-eval}
  T(x)=\bigl(T_1(x)\otimes T_2(x)\bigr)^{(l)} .
\end{equation}
The evaluation is equivariant and homogeneous of polynomial degree $d$, where $d$ is the number of leaves.
This degree is distinct from the representation degree $l$ of the output.
We use the same symbol for a tree and its evaluation.
Its nodes are the tree itself and all its subtrees.
A leaf has depth $0$, and the depth of $(T_1,T_2,l)$ is one plus the larger depth of its two children.

\begin{lemma}
\label{lem:cg-trees}
Every polynomial $G$-equivariant map $P:\calV\to\calW$ is a sum of a constant in the degree-$0$ block of $\calW$ and finitely many coupling trees followed by equivariant linear maps $V_l\to\calW$.
\end{lemma}

\begin{proof}
Decompose $P=\sum_{d=0}^{D}P_d$ into homogeneous terms.
Comparing terms of equal degree in the equivariance identity for $P$ shows that each $P_d$ is equivariant.
Thus $P_0$ is fixed by every rotation and belongs to the degree-$0$ block of $\calW$.

Fix $d\ge1$.
Polarization gives $P_d(x)=B_d(x,\ldots,x)$ for a unique symmetric $d$-linear map $B_d:\calV^d\to\calW$.
Applying a rotation to all arguments preserves this identity, so uniqueness makes $B_d$ equivariant.
It therefore induces an equivariant linear map from $\calV^{\otimes d}$ to $\calW$.
Decomposing each input and output into irreducible channels reduces the claim to maps
$L:V_{l_1}\otimes\cdots\otimes V_{l_d}\to V_l$ that commute with rotations.

For $d=1$, Schur's lemma makes $L$ a scalar multiple of the identity when $l_1=l$, and zero otherwise.
For $d\ge2$, apply the Clebsch--Gordan decomposition successively, first to the first two factors and then to each resulting summand and the next factor.
This identifies the tensor product with a direct sum of irreducible spaces, whose projections are iterated Clebsch--Gordan couplings.
Schur's lemma then expresses $L$ as a linear combination of the projections onto copies of $V_l$~\cite{varshalovich1988}.

For each input factor, let $A_i:\calV\to V_{l_i}$ be the projection onto its selected channel, with $i=1,\ldots,d$.
Evaluating the iterated couplings at $A_1x,\ldots,A_dx$ gives coupling trees on $\calV$.
The output channel inclusions and scalar coefficients give the linear maps into $\calW$, proving the claim.
\end{proof}

The spanning lemma reduces exact realization to evaluating finitely many coupling trees.
A BB computes one level of these trees while retaining the values already computed.

\begin{proposition}[Exact realization of polynomial equivariants]
\label{prop:bn-exact}
Every polynomial $G$-equivariant map $P:\calV\to\calW$ is realized exactly by a BN of the form \cref{eq:bn}.
\end{proposition}

\begin{proof}
By \cref{lem:cg-trees}, write $P=P_0+\sum_\alpha\pi_\alpha\circ T_\alpha$ with finitely many trees $T_\alpha$ and equivariant linear maps $\pi_\alpha:V_{l_\alpha}\to\calW$.
If no trees occur, a one-block BN with output map $\pi\equiv P_0$ suffices.
Otherwise, let $N$ be the number of distinct nodes, $\lambda$ their largest output degree, and $K$ the largest tree depth.
Choose the hidden space of \cref{def:bb} with width $C_{\mathrm H}=N$ and maximal degree $\lambda$.
Assign each node $\nu$ a channel $c_\nu$ at its output degree $l_\nu$.
The input map $\iota$ places each leaf evaluation in its assigned component and sets all remaining components to zero.

Suppose that after $s$ blocks all nodes of depth at most $s$ are stored and all other components are zero.
In the next block, take $L_1=L_2=L_3=L_4=\mathrm{Id}$.
For every node $\nu=(\nu_1,\nu_2,l_\nu)$ of depth $s+1$, set the coefficient $U^{(l_{\nu_1}l_{\nu_2}l_\nu)}_{c_\nu,c_{\nu_1}c_{\nu_2}}$ in \cref{eq:bb-product} to $1$, and set all other coefficients to zero.
Its children have depth at most $s$, so the product gives the evaluation of $\nu$ in a component that was zero.
The residual term in \cref{eq:bb-out} retains all previously stored values.
This proves the induction step.

After $K$ blocks, the output map $\pi$ combines the root evaluations with the maps $\pi_\alpha$ and supplies $P_0$ through its scalar bias.
When $K=0$, insert one identity block to meet the requirement $K_{\mathrm{BN}}\ge1$ in \cref{def:bn}.
Thus depth $K_{\mathrm{BN}}=\max(1,K)$, width $N$, and hidden maximal degree $\lambda$ suffice.
\end{proof}

It remains to pass from polynomial maps to continuous maps.
Averaging a polynomial approximant of an equivariant target over $G$ makes it equivariant without increasing its uniform error on a $G$-invariant compact set~\cite{dym2021universality}.
Combined with exact realization, this yields the following approximation result.

\begin{proposition}[Universality of the bilinear network]
\label{thm:bn-uat}
Every continuous $G$-equivariant map $\calV\to\calW$ can be approximated uniformly on compact sets by BNs of the form \cref{eq:bn}.
\end{proposition}

\begin{proof}
Fix a continuous $G$-equivariant map $F:\calV\to\calW$, a compact set $\Omega\subset\calV$, and $\epsilon>0$.
The set $G\Omega=\{D_{\calV}(R)x:R\in G,\ x\in\Omega\}$ is compact.
Applying the Stone--Weierstrass theorem to each component of $F$ with tolerance $\epsilon/\sqrt{\dim\calW}$ gives a polynomial map $Q:\calV\to\calW$ with $\sup_{G\Omega}\|Q-F\|<\epsilon$ in the Euclidean norm of $\calW$.

With $dR$ the normalized Haar measure, set
\begin{equation}\label{eq:reynolds}
  P(x)=\int_{G}D_{\calW}(R)^{-1}\,Q\bigl(D_{\calV}(R)x\bigr)\,dR ,
\end{equation}
which is polynomial by integration of its coefficients and equivariant by Haar invariance.
Since $F$ is equivariant, $F(x)$ equals the same average of $F$.
Orthogonality of $D_{\calW}(R)$ and the inclusion $D_{\calV}(R)x\in G\Omega$ give
\[
\begin{aligned}
  \sup_{x\in\Omega}\|P(x)-F(x)\|
  &\le\sup_{x\in\Omega}\int_G
    \bigl\|Q\bigl(D_{\calV}(R)x\bigr)-F\bigl(D_{\calV}(R)x\bigr)\bigr\|\,dR\\
  &\le\sup_{y\in G\Omega}\|Q(y)-F(y)\|
  <\epsilon .
\end{aligned}
\]
By \cref{prop:bn-exact}, a BN realizes $P$ exactly.
\end{proof}

For the energy readout, take $\calW=V_0$, so that equivariance is invariance.
The hidden maximal degree $\lambda$, width $C_{\mathrm H}$, and depth $K_{\mathrm{BN}}$ may depend on the target function, the compact set, and the requested accuracy.
In \cref{sec:uat}, we combine this scalar case with the complete representations of \cref{sec:complete} to approximate atomic energies.

%%%%%%%%%%%%%%%%%%%%%%%%%%%%%%%%%%%%%%%%%%%%%%%%%%%%%%%%%%%%%%%%%%
\section{Universal approximation by the BREMP}
\label{sec:uat}
\label{sec:uat-bremp}
%%%%%%%%%%%%%%%%%%%%%%%%%%%%%%%%%%%%%%%%%%%%%%%%%%%%%%%%%%%%%%%%%%

We combine the complete representations of \cref{sec:complete} with the bilinear readout of \cref{sec:readout} to prove universality in the sense of \cref{def:uat}.
To apply the readout approximation, we must express the target energy as a continuous rotation-invariant function on the feature space.
Since the permutation factor $\Gamma$ acts as the identity on the center feature space $\calV$, $SO(3)$-invariance and $G$-invariance coincide for functions on $\calV$.
The following lemma provides this factorization on any prescribed compact set of environments.

\begin{lemma}
\label{lem:factorization}
Let $\calE$ be an environment space of \cref{def:env-space,def:l-hop}, and let $A\subseteq\calE$ be $G$-invariant.
Suppose $\Phi:\calE\to\calV$ is continuous, $G$-equivariant, and complete on $A$.
For every continuous $G$-invariant function $g:A\to\RR$ and compact set $K\subseteq A$, there is a continuous $SO(3)$-invariant function $f:\calV\to\RR$ such that $g=f\circ\Phi$ on $K$.
\end{lemma}

\begin{proof}
Let $K^\ast=\{(R,I)\cdot\calD:R\in SO(3),\ \calD\in K\}$.
The continuous rotation action maps $SO(3)\times K$ onto $K^\ast$, proving compactness.
It is rotation invariant and contained in $A$.
On $\Omega=\Phi(K^\ast)$, set $f_0(\Phi(\calD))=g(\calD)$.
Completeness with $R=I$ in \cref{def:complete} shows that equal features correspond to environments differing only by a type-preserving permutation.
Invariance of $g$ therefore makes $f_0$ well defined.
The continuous surjection $\Phi:K^\ast\to\Omega$ is a quotient map because its domain is compact and its image is Hausdorff.
Since $f_0\circ\Phi=g$, the function $f_0$ is continuous.
Equivariance of $\Phi$ and invariance of $g$ also make $f_0$ rotation invariant.

The Tietze extension theorem gives a continuous extension $\tilde f$ of $f_0$ from the compact set $\Omega$ to $\calV$.
Averaging this extension over the normalized Haar measure,
\[
  f(x)=\int_{SO(3)}\tilde f\bigl(D_{\calV}(R)x\bigr)\,dR ,
\]
gives a continuous rotation-invariant function on $\calV$.
Because $\Omega$ is rotation invariant and $\tilde f=f_0$ there, the average agrees with $f_0$ on $\Omega$.
Thus $g=f\circ\Phi$ on $K$.
\end{proof}

We first apply this lemma when one message-passing layer sees the entire physical environment.

\begin{proposition}[Universal approximation by one layer]
\label{thm:uat-one}
Fix integers $\kappa\ge1$ and $l_{\max}\ge\kappa$.
The family of one-layer BREMPs with cutoff $R_c$ and maximal feature degree $l_{\max}$, with the width, radial profiles, and BN readout chosen freely, is universal for $\calF^0_{R_c}$ on $\calE^{(R_c)}_\kappa$.
For $\kappa=M$, this includes all of $\calE^{(R_c)}$.
\end{proposition}

\begin{proof}
Let $\varepsilon\in\calF^0_{R_c}$, let $K\subseteq\calE^{(R_c)}_\kappa$ be compact, and let $\delta>0$.
Choose the complete representation $\Phi^{[1]}_\theta$ from \cref{prop:one-layer-complete}.
It is continuous by \cref{lem:rep-continuous}.
Applying \cref{lem:factorization} to $\varepsilon$ gives a continuous rotation-invariant function $f:\calV\to\RR$ with $\varepsilon=f\circ\Phi^{[1]}_\theta$ on $K$.
By \cref{thm:bn-uat} with output space $V_0$, a BN approximates $f$ within $\delta$ on the compact set $\Phi^{[1]}_\theta(K)$.
The energy $\varepsilon_\theta=\mathrm{BN}\circ\Phi^{[1]}_\theta$ then satisfies $\sup_K|\varepsilon_\theta-\varepsilon|<\delta$.
The case $\kappa=M$ follows from $\calE^{(R_c)}_M=\calE^{(R_c)}$ in \cref{def:kappa-generic}.
\end{proof}

For a shorter layer cutoff, the same argument applies once the recovered $T$-hop environment contains the physical environment.

\begin{theorem}[Universal approximation by $T$ layers]
\label{thm:uat-multi}
Fix integers $\kappa,T\ge1$ and a cutoff $0<r_c\le R_c$.
Let $A\subseteq\calE^{[r_c,T]}_\kappa$ be $G$-invariant and satisfy the coverage condition \cref{eq:physical-range-coverage}.
The family of $T$-layer BREMPs with cutoff $r_c$, filter degrees at most $\kappa$, and maximal feature degree at most $T(\kappa-1)+1$, with the widths, radial profiles, and BN readout chosen freely, is universal for $\calF^0_{R_c}$ on $A$.
\end{theorem}

\begin{proof}
Let $\varepsilon\in\calF^0_{R_c}$, let $K\subseteq A$ be compact, and let $\delta>0$.
Coverage allows us to define $g(\calD^{[r_c,T]}_j)=\varepsilon(\calD^{(R_c)}_j)$ by retaining only the neighbors within distance $R_c$.
This restriction commutes with rotations and type-preserving permutations, so $g$ is $G$-invariant.
Within each stratum, $g$ is continuous away from crossings of the physical cutoff by continuity of $\varepsilon$.
Condition~3 of \cref{def:target-space} gives continuity at those crossings, including simultaneous crossings by several atoms.

Choose $\Phi^{[T]}_\theta$ from \cref{thm:multi-complete} with the stated degree bounds.
It is complete on $A$ and continuous by \cref{lem:rep-continuous}.
Apply the factorization and scalar readout approximation used in the proof of \cref{thm:uat-one}, now to $g$ and $\Phi^{[T]}_\theta$.
This gives a BN with $\sup_K|\mathrm{BN}\circ\Phi^{[T]}_\theta-g|<\delta$, which is the required estimate \cref{eq:uat-def}.
\end{proof}

Connectivity gives a suitable depth for each finite configuration, as discussed in \cref{sec:setting-pes}.
The hypothesis of \cref{thm:uat-multi} requires one common depth for the entire class.

The universality conclusions also hold when each radial profile is restricted to a family dense in $C([0,r_c])$ in the uniform norm.
The constructions of \cref{sec:complete} use polynomial profiles, which can be approximated uniformly while retaining the fixed envelope $\chi$.
For a fixed network, the envelope, spherical harmonics, and neighbor counts are uniformly bounded, so each convolution changes uniformly little on bounded feature sets when its profiles change uniformly.
Starting from the bounded type embedding, induction over the finitely many layers gives uniformly small feature errors.
Uniform continuity of the BN on a closed ball containing the original and perturbed features then gives uniformly small energy errors on the target compact set.

Universal approximation also requires the representation family to distinguish environments that some admissible target separates.
Since the targets are reflection invariant, the relevant orbits are those of $O(3)\times\Gamma$.
This necessary separation property is weaker than completeness in \cref{def:complete}.

\begin{proposition}[Nonseparation excludes universality]
\label{prop:incomplete-not-uat}
Let $\{\Phi_\theta\}_\theta$ be a family of continuous $G$-equivariant representations $\calE^{(R_c)}\to\calV$.
Suppose two environments $\calD_1,\calD_2\in\calE^{(R_c)}$ are not related by an orthogonal transformation and a type-preserving permutation, but satisfy $\Phi_\theta(\calD_1)=\Phi_\theta(\calD_2)$ for every $\theta$.
For any family of continuous readouts $f_\theta:\calV\to\RR$, the family $\{f_\theta\circ\Phi_\theta\}_\theta$ is not universal for $\calF^0_{R_c}$ on any $G$-invariant set containing both environments.
\end{proposition}

\begin{proof}
Every model takes the same value at $\calD_1$ and $\calD_2$.
We construct a target that separates them.
Take the complete one-layer representation $\Psi:\calE^{(R_c)}\to\calW$ from \cref{prop:one-layer-complete} with $\kappa=l_{\max}=M$.
It consists of the center-type code and the moments \cref{eq:B}.
By the parity of spherical harmonics, it is also $O(3)$-equivariant when inversion acts as $(-1)^l$ in degree $l$, as in \cref{rem:reflections}.
All matrices $D_{\calW}(Q)$ for $Q\in O(3)$ are orthogonal.
Define
\begin{equation*}
  \varepsilon(\calD)=\min_{Q\in O(3)}
  \bigl\|\Psi(\calD)-D_{\calW}(Q)\Psi(\calD_2)\bigr\| .
\end{equation*}
This is the distance to the compact $O(3)$-orbit of $\Psi(\calD_2)$, so it is continuous on each stratum by \cref{lem:rep-continuous}.
Equivariance of $\Psi$ and orthogonality of $D_{\calW}(Q)$ give $O(3)\times\Gamma$ invariance of $\varepsilon$.
Each moment contribution vanishes when its neighbor reaches $R_c$, while the center-type code stays fixed.
Thus $\varepsilon$ satisfies the cutoff continuity in \cref{def:target-space} and belongs to $\calF^0_{R_c}$.

We have $\varepsilon(\calD_2)=0$.
If $\varepsilon(\calD_1)=0$, compactness gives $Q\in O(3)$ with $\Psi(\calD_1)=D_{\calW}(Q)\Psi(\calD_2)=\Psi(Q\cdot\calD_2)$.
Completeness with $R=I$ then forces $\calD_1=(Q,\pi)\cdot\calD_2$ for some type-preserving permutation $\pi$, contrary to the hypothesis.
Hence $\varepsilon(\calD_1)>0$, and every model has uniform error at least $\varepsilon(\calD_1)/2$ on the compact set $\{\calD_1,\calD_2\}$.
\end{proof}

%%%%%%%%%%%%%%%%%%%%%%%%%%%%%%%%%%%%%%%%%%%%%%%%%%%%%%%%%%%%%%%%%%
\section{Universal approximation by DPA4}
\label{sec:uat-dpa4}
%%%%%%%%%%%%%%%%%%%%%%%%%%%%%%%%%%%%%%%%%%%%%%%%%%%%%%%%%%%%%%%%%%

We transfer the universality results of \cref{sec:uat-bremp} to DPA4 by approximating the BREMP layers constructed in those proofs and realizing the chosen BN readout.
We first establish the expressive power of WBBs needed for the readout construction.

\subsection{Wigner bilinear blocks and frame truncation}
\label{sec:wbb-expressivity}

The Wigner product identity \cref{eq:wigner-product} assigns a frame coefficient to each Clebsch--Gordan coupling.
To reproduce an arbitrary BB, the retained frames must supply a nonzero coefficient for every coupling path.
The following lemma shows that frame truncation with $k_{\max}\ge1$ suffices.

\begin{lemma}
\label{lem:frame-truncation}
Let $(l_1,l_2,L)$ be a coupling path in the triangle range.
There are frames $k_1,k_2$ with $|k_1|\le\min(1,l_1)$, $|k_2|\le\min(1,l_2)$, and $C^{(L,0)}_{(l_1,k_1)(l_2,k_2)}\ne0$.
If $l_1+l_2+L$ is even, the frames $k_1=k_2=0$ serve.
If $l_1+l_2+L$ is odd, then $C^{(L,0)}_{(l_1,0)(l_2,0)}=0$.
\end{lemma}

\begin{proof}
Write $\tilde C$ for the Clebsch--Gordan coefficients in the complex spherical basis with the Condon--Shortley phase convention~\cite{varshalovich1988}.
We first establish nonvanishing in these coordinates and then pass to the real coordinates used in the statement.

Suppose $l_1+l_2+L=2\gamma$ is even.
The classical formula for the coefficient with zero magnetic indices gives
\begin{equation}\label{eq:frame-even}
  \left|\tilde C^{(L,0)}_{(l_1,0)(l_2,0)}\right|
  =\sqrt{2L+1}\,
   \frac{\gamma!}{(\gamma-l_1)!(\gamma-l_2)!(\gamma-L)!}
   \sqrt{\frac{(2\gamma-2l_1)!(2\gamma-2l_2)!(2\gamma-2L)!}{(2\gamma+1)!}} .
\end{equation}
The triangle inequalities make all factorial arguments nonnegative integers, so the coefficient is nonzero.

Suppose now that $l_1+l_2+L$ is odd.
Then $l_1,l_2\ge1$, since a zero input degree would force an even sum.
Set $s=\min(l_1,l_2)$ and $c_m=\tilde C^{(L,0)}_{(l_1,m)(l_2,-m)}$ for $-s\le m\le s$.
The complex selection rule $M=m_1+m_2$ implies that these are the only coordinates of the unit coupled vector with degree $L$ and magnetic index $0$ that can be nonzero.
Reversal symmetry gives $c_{-m}=(-1)^{l_1+l_2-L}c_m=-c_m$, hence $c_0=0$.
Normalization gives $\sum_{m=-s}^{s}|c_m|^2=1$.
We show that $c_1\ne0$ by a three-term recurrence.
Let $e_m=e^{(l_1)}_m\otimes e^{(l_2)}_{-m}$, where $e^{(l)}_m$ denotes the complex spherical basis vector of degree $l$ and index $m$.
On the $M=0$ subspace spanned by the $e_m$, the squared total angular momentum has a tridiagonal matrix with entries $a_m$ between $e_m$ and $e_{m+1}$~\cite{varshalovich1988}.
Its coupled eigenvector $\sum_m c_me_m$ has eigenvalue $L(L+1)$, giving the recurrence
\begin{gather}
  a_m=\sqrt{(l_1-m)(l_1+m+1)(l_2-m)(l_2+m+1)},\qquad -s\le m<s,\nonumber\\
  a_mc_{m+1}
  =\bigl(L(L+1)-l_1(l_1+1)-l_2(l_2+1)+2m^2\bigr)c_m-a_{m-1}c_{m-1}.
  \label{eq:frame-recurrence}
\end{gather}
Here \cref{eq:frame-recurrence} holds for $1\le m<s$.
If $c_1=0$, then $c_0=0$ and $a_m>0$ for $0\le m<s$ imply $c_m=0$ for every $1\le m\le s$ by induction.
The reversal symmetry then makes every coordinate vanish, contradicting the unit norm.
Thus $\tilde C^{(L,0)}_{(l_1,1)(l_2,-1)}\ne0$.

The basis change \cref{eq:real-harmonics} fixes coordinate $0$ and mixes only $k$ with $-k$, preserving the input subspaces with $|k_i|\le\min(1,l_i)$ and output coordinate $0$ after complexification.
By the multiplicity-one decomposition in \cref{sec:blocks}, it identifies the complexified real coupling with a nonzero scalar multiple of the standard complex coupling.
The coefficients found above therefore make output coordinate $0$ a nonzero bilinear form on the retained subspaces.
Its real matrix has a nonzero entry $C^{(L,0)}_{(l_1,k_1)(l_2,k_2)}$ with the required frame bounds.
The coefficient with all three indices zero is rescaled by the same nonzero scalar, so it remains nonzero for even sums and zero for odd sums.
\end{proof}

This nonvanishing property allows each coupling to be selected and rescaled independently.

\begin{proposition}[Equivalence of BBs and WBBs]
\label{prop:wbb-full}
On a fixed hidden space $\calH$, every WBB is a BB.
Conversely, every BB can be realized by a WBB on the same space with any frame truncation $k_{\max}\ge1$ and finitely many field pairs.
\end{proposition}

\begin{proof}
Insert the synthesized fields \cref{eq:wbb-synth} into the analysis \cref{eq:wbb-analysis} and expand each product of Wigner functions by \cref{eq:wigner-product}.
Orthogonality \cref{eq:wigner-orth} retains only the analyzed degree $l$ and indices $(m,k)$, giving
\begin{equation}\label{eq:wbb-expand}
  z^{(l)}_{m,k,a}
  =\sum_{\substack{l_1,l_2\le\lambda\\ |l_1-l_2|\le l\le l_1+l_2}}\
   \sum_{k_1\in\calS_{l_1},\,k_2\in\calS_{l_2}}
   C^{(l,k)}_{(l_1,k_1)(l_2,k_2)}\,
   \bigl(u^{(l_1)}_{\cdot,k_1,a}\otimes v^{(l_2)}_{\cdot,k_2,a}\bigr)^{(l)}_m .
\end{equation}
Substituting \cref{eq:wbb-lift,eq:wbb-expand} into \cref{eq:wbb-out} gives the coefficient tensor
\begin{equation}\label{eq:wbb-bb-coeff}
  U^{(l_1l_2l)}_{d,c_1c_2}
  =\sum_{a=1}^{F}\ \sum_{\substack{k\in\calS_l\\k_1\in\calS_{l_1},\,k_2\in\calS_{l_2}}}
    Q^{(l)}_{d,ka}\,C^{(l,k)}_{(l_1,k_1)(l_2,k_2)}\,
    A^{(l_1)}_{k_1a,c_1}B^{(l_2)}_{k_2a,c_2}
\end{equation}
for each coupling path $(l_1,l_2,l)$.
A BB with $L_1=L_2=L_4=\mathrm{Id}$, this $U$, and $L_3$ equal to the affine part of \cref{eq:wbb-out} reproduces the WBB on $\calH$.

For the converse, write each affine map of the given BB as $L_ix=W_ix+b_i$, $i=1,\ldots,4$.
Let $W_i^{(l)}$ denote the degree-$l$ channel matrix of $W_i$.
Since the biases lie in degree $0$, each product containing $b_1$ or $b_2$ is equivariant and affine in $x$.
Absorb their images under $W_4$, together with $L_3x+b_4$, into the affine part of \cref{eq:wbb-out}.
It remains to realize the quadratic terms involving $W_1x$ and $W_2x$.

Assign one field pair $a=((l_1,l_2,l),c_1,c_2)$ to each coupling path and input channel pair in \cref{eq:bb-product}.
Choose frames $(k_1,k_2)$ from \cref{lem:frame-truncation} and write $\gamma_a=C^{(l,0)}_{(l_1,k_1)(l_2,k_2)}\ne0$.
Set the lift coefficients in \cref{eq:wbb-lift} to
\begin{equation}\label{eq:bb-wbb-lift}
  A^{(l_1)}_{k_1a,c}=W^{(l_1)}_{1,c_1c},
  \qquad B^{(l_2)}_{k_2a,c}=W^{(l_2)}_{2,c_2c},
  \qquad 1\le c\le C_{\mathrm H},
\end{equation}
with all other lift coefficients of this field pair zero.
Each field is supported in one degree and one frame.
The analyzed coefficient from \cref{eq:wbb-expand} and the chosen output weights are
\[
  \begin{aligned}
  z^{(l)}_{\cdot,0,a}
  &=\gamma_a\bigl((W_1x)^{(l_1)}_{\cdot,c_1}\otimes(W_2x)^{(l_2)}_{\cdot,c_2}\bigr)^{(l)},\\
  Q^{(l)}_{d,0a}&=\gamma_a^{-1}\sum_cW^{(l)}_{4,dc}U^{(l_1l_2l)}_{c,c_1c_2}.
  \end{aligned}
\]
Set all other output coefficients of this field pair to zero.
The factor $\gamma_a$ then cancels, leaving the contribution of the chosen path and channel pair to the BB output.
Summing over field pairs reproduces the desired quadratic part, including the output mixing by $W_4$.
The number $F$ of field pairs is therefore at most the number of coupling-path and input-channel-pair choices in \cref{eq:bb-product}.
\end{proof}

\begin{remark}[Sphere projection]
\label{rem:sphere-projection}
The choice $k_{\max}=0$ gives the sphere projection used in spherical activations~\cite{cohen2018spherical,passaro2023reducing,liao2023equiformerv2}.
Its frame factor $C^{(l,0)}_{(l_1,0)(l_2,0)}$ vanishes when $l_1+l_2+l$ is odd, by \cref{lem:frame-truncation}.
Consequently, such blocks and their compositions commute with the inversion acting as $(-1)^l$ on degree $l$.
This additional symmetry excludes some $SO(3)$-equivariant maps, including the cross product $1\otimes1\to1$, so the full conclusions of \cref{prop:bn-exact,thm:bn-uat} do not hold with $k_{\max}=0$.
For $k_{\max}\ge1$, \cref{prop:wbb-full} transfers both results to networks of the form \cref{eq:bn} with WBBs in place of BBs.
The threshold $k_{\max}\ge1$ is therefore sharp for realizing arbitrary BBs and for universality of these networks on general feature spaces.
\end{remark}

\subsection{Approximation of message-passing layers}
\label{sec:dpa4-layer-approx}

We simulate the BREMP layers constructed in \cref{prop:one-layer-complete,thm:multi-complete}, which have the form
\begin{equation}\label{eq:layer-block}
  h^{[t+1]}_j=h^{[t]}_j+\iota_t\,\mathrm{Conv}^{[t]}(h^{[t]})_j ,
  \qquad t=0,\ldots,T-1 .
\end{equation}
The feature channels are partitioned into disjoint blocks $\calC_0,\ldots,\calC_T$, with the one-hot type embedding in $\calC_0$.
The map $\iota_t$ inserts the convolution output into the initially zero block $\calC_{t+1}$ without mixing or bias.
The final feature thus retains the embedding and all messages in separate blocks.
The proofs of \cref{thm:uat-one,thm:uat-multi} establish universality using these representations and a BN readout.
We transfer this result by approximating the layers and realizing the readout in DPA4.

We first encode the BREMP features in a larger DPA4 feature space.
The encoding uses a small amplitude to control normalization and nonlinear activation, paired channels to cancel the scalar mean on the message branch, and a constant channel for the readout.

Write $\calV$ and $\calV'$ for the BREMP and DPA4 feature spaces, with widths $C$ and $C'$, respectively.
Choose the DPA4 maximal degree $l_{\max}$ to accommodate the BREMP features and the BN hidden degree $\lambda$, padding the former with zeros.
Set $\calC_{\mathrm{msg}}=\calC_1\cup\cdots\cup\calC_T$.
In $\calV'$, retain the original channels and add a duplicate $\bar c$ for every $c\in\calC_{\mathrm{msg}}$, a constant channel $c_\ast$, and two spare blocks $\calC_{\mathrm{rad}}$ and $\calC_{\mathrm H}$.
Reserve one channel in $\calC_{\mathrm{rad}}$ for each radial profile $p$ in \cref{eq:radial-form} across all layers.
Reserve $C_{\mathrm H}$ channels in $\calC_{\mathrm H}$ for the BN hidden features in \cref{sec:dpa4-readout-exact}.
Thus we fix the atomic width at
\begin{equation}\label{eq:dpa4-width}
  C'=C+|\calC_{\mathrm{msg}}|+1+C_{\mathrm{rad}}+C_{\mathrm H},
  \qquad C_{\mathrm{rad}}=|\calC_{\mathrm{rad}}|,
\end{equation}
before choosing any layer parameters or error constants.
Both spare blocks remain zero in the atomic features during message passing.

For $\tau>0$, define $E_\tau:\calV\to\calV'$ by
\begin{equation}\label{eq:dpa4-encoding}
  \begin{aligned}
    (E_\tau h)^{(l)}_{m,c}&=\tau h^{(l)}_{m,c} &&(c\le C),\\
    (E_\tau h)^{(l)}_{m,\bar c}&=-\tau h^{(l)}_{m,c} &&(c\in\calC_{\mathrm{msg}}),\\
    (E_\tau h)^{(l)}_{m,c_\ast}&=\tau\delta_{l0},
  \end{aligned}
\end{equation}
with all spare components zero.
The map is affine and equivariant, with constants only in degree $0$.
Let $D_\tau:\calV'\to\calV$ select the original channels and divide by $\tau$.
This linear map has operator norm $1/\tau$ and satisfies $D_\tau E_\tau=\mathrm{Id}$.
We apply $E_\tau$ separately to each atom when its argument is a feature family.
A branch output supported on message channels and their duplicates has zero scalar mean when each pair has opposite values.
We use this property to control normalization.

The BREMP envelope is chosen as $\chi=s_5^2$, where $s_5$ is the DPA4 attention envelope.
The radial basis uses a second envelope $s_7$.
Both are continuous, take values in $[0,1]$, are positive below $r_c$, and vanish at and above $r_c$.
Throughout the DPA4 simulation, the Wigner analysis uses exact quadrature and the optional short-range mechanism is disabled.
The DPA4 operators used below are specified in Appendix~\ref{app:dpa4}.

\paragraph{Radial approximation and initialization.}
For the layer estimates, each environment is evaluated as a configuration of its center and listed atoms.
On a compact set $K$ of environments, choose a common lower bound $0<r_{\min}<r_c$ on pair distances, using compactness in the collision-free strata.
With the radial frequencies fixed at $n\pi/r_c$, the DPA4 basis functions are
\begin{equation}\label{eq:dpa4-radial-basis}
  \phi_n(r)=s_7(r)\frac{\sin(n\pi r/r_c)}{r},\qquad n=1,\ldots,n_r.
\end{equation}
The radial MLP uses the activation $\mathrm{SiLU}(u)=u/(1+e^{-u})$ and has no biases.
The next lemma approximates all required radial functions using one shared radial network.

\begin{lemma}
\label{lem:dpa4-radial}
Let $0<r_{\min}<r_c$ and let $p_1,\ldots,p_{C_{\mathrm{rad}}}$ be continuous profiles on $[0,r_c]$.
For every $\delta>0$, a bias-free MLP with SiLU activations, any fixed positive number of hidden layers, and sufficient width and radial basis size has outputs $\hat p_\nu(r)$ satisfying
\begin{equation}\label{eq:dpa4-radial-approx}
  \max_{1\le\nu\le C_{\mathrm{rad}}}\ \sup_{r_{\min}\le r\le r_c}
  |\chi(r)\hat p_\nu(r)-\chi(r)p_\nu(r)|<\delta.
\end{equation}
Each output is a finite linear combination of the basis functions \cref{eq:dpa4-radial-basis}.
\end{lemma}

\begin{proof}
Fix one profile $p$ and write $F=\chi p$ and $\varsigma(r)=\chi(r)s_7(r)/r$ on $[r_{\min},r_c]$.
The function $\varsigma$ is positive below $r_c$, and $F(r_c)=\varsigma(r_c)=0$.
Choose a continuous cutoff $\xi$ that equals $1$ away from $r_c$, vanishes near $r_c$, and satisfies $\|F-\xi F\|_\infty<\delta/2$.
Then $q=\xi F/\varsigma$, extended by zero at $r_c$, is continuous.
Extend it continuously to $[0,r_c]$ with $q(0)=0$, and then to an odd continuous $2r_c$-periodic function.
The Fej\'er means of its Fourier series are sine polynomials and converge uniformly.
Hence a finite sum $S(r)=\sum_n c_n\sin(n\pi r/r_c)$ satisfies
\[
  \|\varsigma S-F\|_\infty
  \le\|\varsigma\|_\infty\|S-q\|_\infty+\|\xi F-F\|_\infty<\delta.
\]
Taking $\hat p=\sum_n c_n\phi_n$ gives $\chi\hat p=\varsigma S$.

To realize $\hat p$, use a pair of units with preactivations $\hat p$ and $-\hat p$.
The identity $\mathrm{SiLU}(u)-\mathrm{SiLU}(-u)=u$ lets their output difference supply $\hat p$ to the next layer or the final output.
For all profiles, take a common $n_r$ and one pair per profile in every hidden layer.
The preactivations and final outputs use only linear combinations, so no biases are needed.
\end{proof}

Store these outputs in the degree-$0$ radial channels indexed by $\calC_{\mathrm{rad}}$.
Their corresponding atomic channels remain zero, and each layer selects the radial outputs it needs.
Set the species-pair radial table and all positive-degree radial outputs to zero, disabling the positive-degree geometric embedding.
Choosing $W_\alpha=W_\beta=0$ in \cref{eq:app-dpa4-embed} sets the scalar scale and shift to $1$ and $0$.
For the one-hot type code $e_z\in\calV$ supported in $\calC_0$, set the type-table entry to $E_\tau(e_z)$.
The initial features then equal the encoded BREMP embedding exactly.

\paragraph{Coupling in a bond frame.}
The angular part of a BREMP message couples a neighbor feature to a spherical harmonic of the bond direction.
In a frame whose polar axis is the bond, this coupling is an $SO(2)$-equivariant linear map~\cite{passaro2023reducing}.
The following form of this reduction is used in the construction.

\begin{lemma}
\label{lem:dpa4-bond-frame}
Fix an admissible path $(l_i,l_f,l_o)$.
There is an $SO(2)$-equivariant linear map $A^{(l_i,l_f,l_o)}:V_{l_i}\to V_{l_o}$ such that, for every $\hat{\bm r}\in S^2$ and $R\in SO(3)$ with $R\hat{\bm r}=\bm e_z$,
\begin{equation}\label{eq:dpa4-bond-frame}
  \bigl(Y^{(l_f)}(\hat{\bm r})\otimes v\bigr)^{(l_o)}
  =D^{(l_o)}(R)^{\top}A^{(l_i,l_f,l_o)}D^{(l_i)}(R)v,\qquad v\in V_{l_i}.
\end{equation}
In real coordinates, $A^{(l_i,l_f,l_o)}$ maps order $0$ to order $0$ and acts between matching pairs of orders $\pm m$ by matrices $\alpha I+\beta J$, where
$J=\bigl(\begin{smallmatrix}0&-1\\1&0\end{smallmatrix}\bigr)$.
\end{lemma}

\begin{proof}
Set $A^{(l_i,l_f,l_o)}v=(Y^{(l_f)}(\bm e_z)\otimes v)^{(l_o)}$.
Equivariance of the harmonic and the coupling gives
\[
  D^{(l_o)}(R)\bigl(Y^{(l_f)}(\hat{\bm r})\otimes v\bigr)^{(l_o)}
  =\bigl(Y^{(l_f)}(\bm e_z)\otimes D^{(l_i)}(R)v\bigr)^{(l_o)},
\]
which proves \cref{eq:dpa4-bond-frame}.
A rotation $Q$ about $\bm e_z$ fixes $Y^{(l_f)}(\bm e_z)$.
Hence $A^{(l_i,l_f,l_o)}D^{(l_i)}(Q)=D^{(l_o)}(Q)A^{(l_i,l_f,l_o)}$.
Such rotations act trivially on order $0$ and by a planar rotation of angle $m\theta$ on the pair $\pm m$.
Their distinct frequencies force an intertwining map to connect only matching orders.
On each matching nonzero pair, its matrix commutes with all planar rotations.
Commutation with $J$ gives precisely the form $\alpha I+\beta J$.
\end{proof}

The bond-frame linear maps in DPA4 allow this block structure between every pair of degrees and channels.
With full order $m_{\max}=l_{\max}$, they can therefore supply each $A^{(l_i,l_f,l_o)}$ without discarding any input coordinate.
We now combine this angular construction with the radial approximation and control the nonlinear operations in the layer.

\begin{proposition}
\label{prop:dpa4-layer}
Let $\calL$ be a BREMP layer of the form \cref{eq:layer-block}, let $K$ be a compact set of environments, and let $\calA\subset\calV$ be bounded.
For every $\eta>0$, there are a DPA4 interaction layer $\widehat{\calL}$ with $m_{\max}=l_{\max}$, finite width, radial basis size, and kernel rank, and a scale $\tau_0>0$ with the following property.
For any environment in $K$, let $h=(h_j)_j$ assign an arbitrary input feature $h_j\in\calA$ to every atom, including the center.
Then, at every atom $j$,
\[
  \bigl\|\widehat{\calL}(E_\tau h)_j-E_\tau(\calL(h)_j)\bigr\|\le\eta\tau
  \qquad (0<\tau\le\tau_0).
\]
Moreover, $\widehat{\calL}(E_\tau h)_j$ belongs to $E_\tau(\calV)$.
\end{proposition}

\begin{proof}
Write $\calL(h)_j=h_j+\iota_{t-1}\mathrm{Conv}(h)_j$ for a fixed $t\in\{1,\ldots,T\}$.
Set $x=E_\tau h$ for the input feature family.
Use one focus stream, one attention head, and full order $m_{\max}=l_{\max}$.
Zero both WBB output maps, including the one inside $\calC$.
The layer then reduces to $x_j+N_{\mathrm{post}}(\calC(x)_j)$.
Unused parameter entries and all message biases are zero.

\emph{Radial functions and bond messages.}
Choose the shared radial network from \cref{lem:dpa4-radial} with tolerance $\delta$, to be fixed below.
Let $\widehat{\mathrm{Conv}}$ denote the BREMP convolution with $p_\nu$ replaced by $\hat p_\nu$.
Boundedness of the harmonics, features, and neighbor counts gives
\begin{equation}\label{eq:dpa4-conv-error}
  \|\widehat{\mathrm{Conv}}(h)_j-\mathrm{Conv}(h)_j\|\le c_1\delta,
\end{equation}
uniformly over the environments, atoms, and input feature families in the statement.
The constant $c_1$ is independent of the approximating radial network.

Assign a hidden channel $\nu$ and a duplicate $\bar\nu$ to every path and channel-pair choice $((l_i,l_f,l_o),c_i,c_o)$ of this layer.
Fix a hidden width $H\ge C'$ large enough for these pairs.
The pre-mixer copies input channel $c_i$ to $\nu$ and its negative to $\bar\nu$.
With kernel rank $H$ and the identity channel basis, each hidden channel has an independent linear functional of the radial feature.
Choose it to multiply degree $l_i$ by $\hat p_\nu(r_{jk})$ and to set the other degrees to zero.
Choose $R_{jk}\in SO(3)$ with $R_{jk}\hat{\bm r}_{jk}=\bm e_z$.
After the bond rotation and radial kernel, the bond feature is $x_{jk}=\tau a_{jk}$, with
\[
  (a_{jk})^{(l_i)}_{\cdot,\nu}
  =\hat p_\nu(r_{jk})D^{(l_i)}(R_{jk})h^{(l_i)}_{k,\cdot,c_i},
  \qquad (a_{jk})_{\bar\nu}=-(a_{jk})_\nu.
\]
The channel pairing holds in every degree.
The family $a_{jk}$ is bounded and has zero scalar mean.

In the normalization \cref{eq:app-dpa4-norm}, choose gains $\sqrt{\epsilon_{\mathrm N}}$ and zero shifts.
With the variance $\sigma^2$ and fixed $\epsilon_{\mathrm N}>0$ defined there, features $a$ of zero scalar mean satisfy
\begin{equation}\label{eq:dpa4-norm-small}
  \begin{aligned}
    N(\tau a)&=\frac{\tau a}{\sqrt{1+\tau^2\sigma^2(a)/\epsilon_{\mathrm N}}},\\
    \|N(\tau a)-\tau a\|&\le\frac{\tau^3\sigma^2(a)}{2\epsilon_{\mathrm N}}\|a\|.
  \end{aligned}
\end{equation}
The bound follows from $0\le1-(1+u)^{-1/2}\le u/2$ for $u\ge0$.
Set the gate matrices of the activation in the first local residual step to zero and denote the resulting map by $\calG$.
It applies SiLU in degree $0$ and multiplication by $1/2$ in every positive degree.
Thus $\calG(\tau a)-\tau a/2$ is supported only in degree $0$.
The scalar estimate $|\mathrm{SiLU}(u)-u/2|\le u^2/4$ gives
\[
  \|\calG(\tau a)-\tau a/2\|\le\tau^2\|a\|^2/4.
\]
This bound holds for every $a$, without a mean constraint.

Let $A$ act on each channel $\nu,\bar\nu$ by the map $A^{(l_i,l_f,l_o)}$ of \cref{lem:dpa4-bond-frame} from degree $l_i$ to degree $l_o$, and vanish on the other input degrees.
In the first local residual step, choose the linear map $2(A-I)$ and residual scale $1$.
Set the residual scales of the remaining local steps to zero.
The two estimates above give
\[
  \tau a_{jk}+\calG\bigl(2(A-I)N(\tau a_{jk})\bigr)
  =\tau A a_{jk}+O(\tau^2).
\]
Rotating back and applying \cref{eq:dpa4-bond-frame}, the message in channel $\nu$ is
\begin{equation}\label{eq:dpa4-bond-approx}
  m_{jk,\cdot,\nu}
  =\tau\hat p_\nu(r_{jk})
    \bigl(Y^{(l_f)}(\hat{\bm r}_{jk})\otimes h^{(l_i)}_{k,\cdot,c_i}\bigr)^{(l_o)}
    +O(\tau^2).
\end{equation}
Here the leading term occupies degree $l_o$.
The remainder is uniform because the profiles and input features are bounded and the rotation matrices are orthogonal.
The duplicate messages are opposite only up to $O(\tau^2)$, since SiLU is not odd.

\emph{Attention and the neighbor sum.}
Set all logits in \cref{eq:app-dpa4-attention} to zero.
The denominator parameter $\Xi$ can be arbitrarily large, and the weights satisfy
\begin{gather}
  w_{jk}=\frac{\chi(r_{jk})}{\Xi+S_j},
  \qquad S_j=\sum_{k'\in\calN(j)}\chi(r_{jk'})\le M,
  \label{eq:dpa4-attention-zero}\\
  0\le\Xi w_{jk}\le\chi(r_{jk})\le1,
  \qquad
  |\Xi w_{jk}-\chi(r_{jk})|
  =\frac{\chi(r_{jk})S_j}{\Xi+S_j}\le\frac{M}{\Xi}.
  \label{eq:dpa4-attention-error}
\end{gather}
Set the aggregate gate to $1/2$ by zeroing its weights.
For each $\nu$ with output channel $c_o\in\calC_t$, choose post-mixer weights $2\Xi$ into $c_o$ and $-2\Xi$ into $\bar c_o$.
The factor $2$ cancels the aggregate gate.
Reading only the channels $\nu$ makes the output exactly paired even when the incoming message pairs are not.

Let $\iota'_t$ insert a message into $\calC_t$ and its negative into the duplicates, with zeros elsewhere.
Substituting \cref{eq:dpa4-bond-approx} into the neighbor sum and using \cref{eq:dpa4-attention-error} gives
\begin{equation}\label{eq:dpa4-branch-error}
  \tau^{-1}\calC(E_\tau h)_j
  =\iota'_t\widehat{\mathrm{Conv}}(h)_j+e_\Xi+e_\tau,
  \qquad \|e_\Xi\|\le c_2/\Xi,\quad \|e_\tau\|\le c_2\tau.
\end{equation}
Indeed, the weight error contributes at most a fixed multiple of $M/\Xi$ after division by $\tau$.
The bond remainder contributes $O(\tau)$ because $\Xi w_{jk}\le1$ and each sum has at most $M$ neighbors and finitely many channels.
With the radial profiles fixed, $c_2$ is independent of $\Xi$ and $0<\tau\le1$.

\emph{Normalization and the residual update.}
Write $\calC(E_\tau h)_j=\tau b_j$.
For fixed profiles and $\Xi$, \cref{eq:dpa4-branch-error} makes the family $b_j$ bounded uniformly for $0<\tau\le1$.
It has zero scalar mean by the exact pairing of the post-mixer.
Applying \cref{eq:dpa4-norm-small} at width $C'$ gives $\|N_{\mathrm{post}}(\tau b_j)-\tau b_j\|\le c_3\tau^3$.
The encoding satisfies
\[
  E_\tau\bigl(h_j+\iota_{t-1}\mathrm{Conv}(h)_j\bigr)
  =E_\tau(h_j)+\tau\iota'_t\mathrm{Conv}(h)_j.
\]
Since $\|\iota'_t v\|=\sqrt2\|v\|$, subtraction and \cref{eq:dpa4-conv-error,eq:dpa4-branch-error} yield
\begin{equation}\label{eq:dpa4-layer-error}
  \|\widehat{\calL}(E_\tau h)_j-E_\tau(\calL(h)_j)\|
  \le\tau\bigl(\sqrt2 c_1\delta+c_2/\Xi+c_2\tau+c_3\tau^2\bigr).
\end{equation}
Choose the radial approximation first so that $\sqrt2 c_1\delta\le\eta/4$.
This fixes the profiles and $c_2$.
Next choose $\Xi$ so that $c_2/\Xi\le\eta/4$.
Finally, choose $0<\tau_0\le1$ so that $c_2\tau_0+c_3\tau_0^2\le\eta/2$.
All choices are uniform over the environments, atoms, and bounded input feature families, which proves the estimate.

For preservation of the encoded space, the branch output vanishes outside $\calC_t$ and its duplicates and is exactly paired there.
By \cref{eq:dpa4-norm-small}, the post-normalization multiplies this output by one scalar.
It preserves both its support and its pairing.
Adding it to $E_\tau(h_j)$ therefore gives $E_\tau(\hat h_j)$ for some $\hat h_j\in\calV$.
\end{proof}

For prescribed layer errors, choose a common radial tolerance meeting all bounds in \cref{eq:dpa4-conv-error}.
\Cref{lem:dpa4-radial} supplies one network for all profiles, while the atomic width \cref{eq:dpa4-width} stays fixed.
After constructing all layers, take a common $\tau$ below their thresholds.
Their outputs decode as $D_\tau x$ by preservation of $E_\tau(\calV)$.
Iteration also requires these decoded features to remain in the prescribed bounded input sets.
\Cref{sec:dpa4-uat-transfer} verifies this condition and bounds the accumulated error.

\subsection{Exact realization of the bilinear readout}
\label{sec:dpa4-readout-exact}

We now choose the DPA4 readout to recover the BN energy from the encoded features of \cref{sec:dpa4-layer-approx}.
The DPA4 readout uses residual blocks $x\mapsto x+\calB_{\mathrm W}(x,x)$ without normalization.
The implemented block $\calB_{\mathrm W}$ includes scalar gates and a nonlinear degree-$0$ term.
The affine term in \cref{eq:wbb-out} is absent from $\calB_{\mathrm W}$.
We first show that its parameters can be chosen to realize a homogeneous quadratic BB exactly.

\begin{lemma}
\label{lem:dpa4-quadratic}
Let $\calB$ be a BB on $\calV'$ with $L_3=0$ and linear $L_1,L_2,L_4$.
For every $k_{\max}\ge1$, a DPA4 block with sufficient hidden width and exact quadrature can be chosen so that
$\calB_{\mathrm W}(x,x)=\calB(x)$ for all $x\in\calV'$.
\end{lemma}

\begin{proof}
Use the field pairs constructed in the proof of \cref{prop:wbb-full}, with each field supported in one degree and one frame.
The affine part of the resulting WBB vanishes because $\calB$ is homogeneous quadratic.
In the DPA4 block \cref{eq:app-dpa4-wbb-scalar}, set the channel mixing after analysis to the identity and the gate matrix to zero.
For each field pair $a$, write $\alpha_a=u^{(0)}_{0,0,a}$ and $\beta_a=v^{(0)}_{0,0,a}$ for the left and right degree-$0$ coefficients.
With $\alpha_a$ entering the nonlinear gate, the coefficients before the final linear output map are
\begin{equation}\label{eq:dpa4-quadratic-coeff}
  \tilde z^{(l)}_{m,k,a}
  =\frac12 z^{(l)}_{m,k,a}
   +\delta_{l0}\delta_{m0}\delta_{k0}\,\mathrm{SiLU}(\alpha_a)\beta_a,
\end{equation}
where $z$ is the Wigner analysis of the field product in \cref{eq:wbb-analysis}.

If either field has positive degree, then $\alpha_a=0$ or $\beta_a=0$, so doubling the output weights recovers the analyzed product exactly.
For two degree-$0$ fields, the product is constant on $SO(3)$, with only one possibly nonzero analyzed coefficient $z^{(0)}_{0,0,a}=\alpha_a\beta_a$.
Copy this field pair to $a'$ and negate its left degree-$0$ lift weights, giving scalar coefficients $(-\alpha_a,\beta_a)$.
The identity $\mathrm{SiLU}(u)-\mathrm{SiLU}(-u)=u$ gives
\begin{equation}\label{eq:dpa4-scalar-product}
  \frac{\tilde z^{(0)}_{0,0,a}-\tilde z^{(0)}_{0,0,a'}}{2}
  =\frac{\alpha_a\beta_a+
    [\mathrm{SiLU}(\alpha_a)-\mathrm{SiLU}(-\alpha_a)]\beta_a}{2}
  =\alpha_a\beta_a.
\end{equation}
The final linear map combines the recovered products with the output coefficients from \cref{prop:wbb-full} to realize $\calB$ exactly.
At most one extra field pair is needed for each product of two scalar fields.
\end{proof}

The constant channel of \cref{eq:dpa4-encoding} supplies the affine terms of a general BB through homogenization.
For an equivariant polynomial map $P:X\to Y$ of degree at most two between feature spaces, write $P=P_2+P_1+P_0$, where $P_d$ is homogeneous of degree $d$.
With a rotation-invariant scalar $t$, set
\begin{equation}\label{eq:dpa4-homogenization}
  \widetilde P(u,t)=P_2(u)+tP_1(u)+t^2P_0,
  \qquad \widetilde P(\tau u,\tau)=\tau^2P(u).
\end{equation}
As in the proof of \cref{lem:cg-trees}, each $P_d$ is equivariant, so $\widetilde P$ is homogeneous quadratic and equivariant.
The terms $tP_1$ and $t^2P_0$ use $0\otimes l\to l$ and $0\otimes0\to0$, respectively, since $P_0$ is rotation invariant.
We apply this construction to the increments required by the residual readout blocks.

\begin{proposition}
\label{prop:dpa4-readout}
Let $\mathrm{BN}$ have depth $K_{\mathrm{BN}}$, hidden maximal degree at most $l_{\max}$, and a designated degree-$0$ output channel.
For every $\tau>0$ and $k_{\max}\ge1$, a DPA4 readout with $K_{\mathrm{BN}}$ residual blocks, sufficient width, and exact quadrature can be chosen so that its energy head returns $[\mathrm{BN}(h)]^{(0)}_{0,1}+b_z$ from $E_\tau(h)$ for every $h\in\calV$ and center type $z$.
\end{proposition}

\begin{proof}
Write the BN as in \cref{eq:bn}, with hidden features $y_0=\iota(h)$ and $y_s=\calB_s(y_{s-1})\in\calH$ for $s=1,\ldots,K_{\mathrm{BN}}$.
Let $\iota_{\mathrm H}:\calH\to\calV'$ insert these features into the spare block $\calC_{\mathrm H}$ reserved in \cref{eq:dpa4-width}, with zeros elsewhere.
Starting from $x_0=E_\tau(h)$, we construct the readout so that after $s$ blocks
\begin{equation}\label{eq:dpa4-readout-state}
  x_s=E_\tau(h)+\tau\iota_{\mathrm H}y_s,
  \qquad s=1,\ldots,K_{\mathrm{BN}}.
\end{equation}
The hidden block is initially zero, and the construction will leave all channels outside $\calC_{\mathrm H}$ unchanged.

The first block must compute $G_1(h)=\calB_1(\iota(h))$ from the original feature.
Each later block adds the increment $G_s(y)=\calB_s(y)-y$, for $2\le s\le K_{\mathrm{BN}}$.
Each $G_s$ is equivariant and has polynomial degree at most two.
For $x\in\calV'$, denote its original and hidden channel projections by $x_{\mathrm{orig}}\in\calV$ and $x_{\mathrm H}\in\calH$, and its constant-channel scalar by $x_\ast=x^{(0)}_{0,c_\ast}$.
Using the homogenizations from \cref{eq:dpa4-homogenization}, define
\begin{equation}\label{eq:dpa4-readout-increments}
  \begin{aligned}
    Q_1(x)&=\tau^{-1}\iota_{\mathrm H}\widetilde G_1(x_{\mathrm{orig}},x_\ast),\\
    Q_s(x)&=\tau^{-1}\iota_{\mathrm H}\widetilde G_s(x_{\mathrm H},x_\ast),
      &&2\le s\le K_{\mathrm{BN}}.
  \end{aligned}
\end{equation}
Each $Q_s$ is homogeneous quadratic and equivariant.
By the degree-two case of \cref{lem:cg-trees}, it is a BB with $L_3=0$ and linear $L_1,L_2,L_4$.
\Cref{lem:dpa4-quadratic} therefore realizes $x\mapsto x+Q_s(x)$ exactly, with the increment supported only in $\calC_{\mathrm H}$.

The scaling identity \cref{eq:dpa4-homogenization} gives
\[
  \begin{aligned}
    Q_1(E_\tau(h))&=\tau\iota_{\mathrm H}y_1,\\
    Q_s\bigl(E_\tau(h)+\tau\iota_{\mathrm H}y_{s-1}\bigr)
      &=\tau\iota_{\mathrm H}(y_s-y_{s-1}),\qquad 2\le s\le K_{\mathrm{BN}}.
  \end{aligned}
\]
Starting from $x_0=E_\tau(h)$, the updates $x_s=x_{s-1}+Q_s(x_{s-1})$ therefore prove \cref{eq:dpa4-readout-state} by induction.

Since $\pi$ is equivariant and affine, the designated scalar output is $[\mathrm{BN}(h)]^{(0)}_{0,1}=\bm w_\pi^{\top}y^{(0)}_{K_{\mathrm{BN}}}+b_\pi$.
The final degree-$0$ slice contains $\tau y^{(0)}_{K_{\mathrm{BN}}}$ in the hidden block and $\tau$ in the constant channel.
Use one hidden unit of the gated energy head \cref{eq:dpa4-energy,eq:app-dpa4-head}.
On the hidden and constant channels, give its value and gate branches the respective weight rows $\tau^{-1}(\bm w_\pi^\top,b_\pi)$ and $\tau^{-1}(0,1)$.
Set its output weight to $1/\mathrm{SiLU}(1)$ and all other weights to zero.
The gate factor cancels, leaving the BN output plus the fixed type offset $b_z$ for every $h\in\calV$.
\end{proof}

The dependence on $\tau>0$ is absorbed into the block output maps and head weights, so the exact construction applies at the scale used in the layer simulation.

\subsection{Universal approximation by DPA4}
\label{sec:dpa4-uat-transfer}

We now combine the layer approximation of \cref{prop:dpa4-layer} with the exact readout realization of \cref{prop:dpa4-readout}.
The remaining step is to control the accumulated feature error while keeping every layer input in the bounded set required by the approximation.

\begin{corollary}[Universal approximation by DPA4]
\label{cor:dpa4}
Let $\kappa,T,r_c$, and $A$ satisfy the hypotheses of \cref{thm:uat-multi}, and fix an integer $k_{\max}\ge1$.
Consider DPA4 networks of \cref{sec:dpa4} with $T$ interaction layers, cutoff $r_c$, $m_{\max}=l_{\max}$, and an embedding that depends only on atomic types.
Assume exact quadrature for the Wigner analysis and that the optional short-range mechanism is disabled.
With the maximal degree, atomic and hidden widths, radial basis size, kernel rank, and number of readout blocks chosen freely, this family is universal for $\calF^0_{R_c}$ on $A$.
\end{corollary}

\begin{proof}
Fix $\varepsilon\in\calF^0_{R_c}$ and $\delta>0$, together with a compact set $K\subseteq A$.
Subtracting the fixed offset $b_z$ preserves the target class, since $b_z$ depends only on the center type.
The construction in the proof of \cref{thm:uat-multi} gives a BREMP whose energy approximates $\varepsilon-b_z$ within $\delta/2$ on $K$.
Its layers $\calL_1,\ldots,\calL_T$ have the form \cref{eq:layer-block}.
Let $q:\calV\to\RR$ be its scalar BN readout.
Choose $l_{\max}$ to accommodate both the BREMP message features and the hidden space of this BN.
Fix the DPA4 feature space $\calV'$ with the width and reserved channel blocks of \cref{eq:dpa4-width}.

For each $\calD\in K$, evaluate both networks on the finite configuration formed by its center and listed atoms.
The DPA4 attention normalization sums only over neighbors within $r_c$.
The restriction preserves the final center output of either network, since a $T$-layer network with a type-only embedding depends only on its $T$-hop environment.
Write $h_t(\calD)=(h_{t,k}(\calD))_k$ for the BREMP features after $t$ layers, where $k$ ranges over all atoms of this configuration, including its center $j$.
For a feature family $h=(h_k)_k$, use the norm $\|h\|_\infty=\max_k\|h_k\|$.

For fixed geometry, each $\calL_t$ is linear in the input features.
Bounded radial functions, spherical harmonics, and neighbor counts give a Lipschitz constant $L_t$ uniform over $\calD\in K$ in this norm.
Starting from the bounded type embedding, induction therefore bounds all $h_{t,k}(\calD)$ uniformly.
For $t=0,\ldots,T$, choose a closed ball $\calA_t\subset\calV$ containing the unit neighborhoods of these features.
Let $L_q$ be a Lipschitz constant for the polynomial $q$ on $\calA_T$.
These constants are independent of the DPA4 approximation parameters.

Set $\Lambda_0=0$ and $\Lambda_t=1+L_t\Lambda_{t-1}$ for $t=1,\ldots,T$.
Choose $\eta>0$ so that $\eta\Lambda_t<1$ for every $t$ and $L_q\eta\Lambda_T<\delta/2$.
Fix any positive number of hidden layers for the radial MLP.
Use \cref{lem:dpa4-radial} to choose one shared radial network with the tolerance required by all layers, as described at the end of \cref{sec:dpa4-layer-approx}.
Applying \cref{prop:dpa4-layer} to $\calL_t$ on $\calA_{t-1}$ with accuracy $\eta$ gives a DPA4 layer $\widehat{\calL}_t$ and a threshold $\tau_{0,t}>0$.
Fix one scale $0<\tau\le\min_{1\le t\le T}\tau_{0,t}$.
The type-table construction in \cref{sec:dpa4-layer-approx} initializes the DPA4 features at $E_\tau(h_0)$ exactly and disables the geometric embedding.

Let $x_t(\calD)$ be the DPA4 feature family after $t$ interaction layers and set $\hat h_t=D_\tau x_t$.
Set $e_t=\sup_{\calD\in K}\|\hat h_t(\calD)-h_t(\calD)\|_\infty$.
The initialization gives $x_0=E_\tau(h_0)$ and $e_0=0$.
For the induction step, let $1\le t\le T$ and assume $x_{t-1}=E_\tau(\hat h_{t-1})$ and $e_{t-1}\le\eta\Lambda_{t-1}<1$.
Since $\calA_{t-1}$ contains the unit neighborhoods of the BREMP features, every $\hat h_{t-1,k}$ belongs to $\calA_{t-1}$.
Then \cref{prop:dpa4-layer} gives $x_t=E_\tau(\hat h_t)$.
Applying the decoder, whose operator norm is $1/\tau$, to the estimate in \cref{prop:dpa4-layer} yields
\begin{equation}\label{eq:dpa4-error-step}
  \begin{aligned}
    \|\hat h_t-h_t\|_\infty
    &\le \|D_\tau\widehat{\calL}_t(E_\tau\hat h_{t-1})
               -\calL_t(\hat h_{t-1})\|_\infty\\
    &\quad+\|\calL_t(\hat h_{t-1})-\calL_t(h_{t-1})\|_\infty\\
    &\le\eta+L_t\|\hat h_{t-1}-h_{t-1}\|_\infty.
  \end{aligned}
\end{equation}
Taking the supremum over $K$ gives
\[
  e_t\le\eta+L_te_{t-1}\le\eta\Lambda_t<1,
  \qquad t=1,\ldots,T.
\]
Thus $\hat h_{t,k}\in\calA_t$ for every atom $k$, verifying the input condition for the next step and completing the induction.

By \cref{prop:dpa4-readout}, the DPA4 energy is $q(\hat h_{T,j})+b_{z_j}$ at the center $j$.
The difference between $q(\hat h_{T,j})$ and the BREMP energy $q(h_{T,j})$ is at most $L_qe_T<\delta/2$.
Adding the initial approximation error gives \cref{eq:uat-def}.
\end{proof}

The one-layer case follows by taking $r_c=R_c$ and $T=1$.
In this construction, the DPA4 maximal degree is chosen to accommodate both the message-feature bound $T(\kappa-1)+1$ and the hidden degree of the chosen BN.

%%%%%%%%%%%%%%%%%%%%%%%%%%%%%%%%%%%%%%%%%%%%%%%%%%%%%%%%%%%%%%%%%%
\section{Discussion and perspective}
\label{sec:discussion}
%%%%%%%%%%%%%%%%%%%%%%%%%%%%%%%%%%%%%%%%%%%%%%%%%%%%%%%%%%%%%%%%%%
We have established universal approximation of the atomic energies in $\calF^0_{R_c}$ by the BREMP, including when its layer cutoff is shorter than the physical range.
\Cref{thm:uat-multi} applies to environment classes satisfying a uniform shell multiplicity bound and the coverage condition~\cref{eq:physical-range-coverage} at one common depth $T$.
We construct complete equivariant representations of finite width for atomic environments.
We also introduce the BN, a network of Clebsch--Gordan bilinear blocks that can approximate any continuous equivariant map between feature spaces uniformly on compact sets.

The completeness construction provides explicit angular degree bounds.
For shell multiplicity at most $\kappa$, filters of degree at most $\kappa$ in every layer and message-feature degrees at most $T(\kappa-1)+1$ suffice, by~\cref{thm:multi-complete}.
When $\kappa=1$, scalar and vector features suffice for the representation at every depth.
For $\kappa=M$, the shell condition imposes no further restriction, so symmetric environments and crystalline configurations are covered whenever the coverage condition holds.
Reducing the layer cutoff can lower $\kappa$, while the depth must still ensure coverage of the physical environment.

\Cref{cor:dpa4} follows by approximating the constructed BREMP layers with DPA4 interaction layers and using the WBN to realize the BN readout.
The fixed frame truncation $k_{\max}=1$ suffices for this realization.
The number $T$ of interaction layers and the frame truncation can therefore be fixed, while the maximal degree, widths, radial basis size, kernel rank, and number of readout blocks may depend on the target, the compact set, and the requested accuracy.
For networks formed from the abstract WBBs of \cref{def:wbb}, the threshold $k_{\max}\ge1$ is sharp for realizing arbitrary BBs and for universality on general feature spaces, by \cref{rem:sphere-projection}.

The approximation results are uniform on compact subsets of the admissible environment classes in the topology of~\cref{sec:completeness-uat}.
They establish the existence of finite networks but provide no approximation rates in terms of network size.
Uniform approximation of energy alone does not control errors in forces or higher derivatives of the energy, even when the target energy is smooth.
Quantitative approximation of these quantities therefore requires further analysis.
The constructions also leave unresolved the minimal filter degree as a function of shell multiplicity, and whether the growth of the message-feature degree with depth is necessary when $\kappa\ge2$.

\appendix
\section{DPA4 operators used in the simulation}
\label{app:dpa4}

This appendix specifies the DPA4 operators used in \cref{sec:dpa4-layer-approx,sec:dpa4-readout-exact}.
We use the notation $x_j\in\calV'$ and width $C'$ of \cref{eq:dpa4-width}.
The message operator has one focus stream, one attention head, and full bond-frame order $m_{\max}=l_{\max}$, as chosen in the proof of \cref{prop:dpa4-layer}.
The optional short-range mechanism is disabled.
The proof of \cref{prop:dpa4-layer} specifies the parameters used to approximate a BREMP layer.

\subsection{Radial features and initialization}
\label{app:dpa4-input}

The envelopes $s_5,s_7$ are continuous, positive below $r_c$, bounded by $1$, and zero at and above $r_c$.
The radial feature of a bond is
\begin{equation}\label{eq:app-dpa4-rho}
  \begin{aligned}
    \rho_{jk,l,c}
      &=\bigl[\mathrm{MLP}^{\mathrm{rad}}(\phi(r_{jk}))\bigr]_{l,c}
        +\bigl[\mathbf T_{\mathrm{edge}}(z_j,z_k)\bigr]_c,\\
    \phi_n(r)&=s_7(r)\frac{\sin(\omega_n r)}{r},\qquad n=1,\ldots,n_r .
  \end{aligned}
\end{equation}
Here $\mathbf T_{\mathrm{edge}}$ is a learned type-pair table, and the frequencies $\omega_n$ are learnable.
The bias-free radial MLP has SiLU activations and $(l_{\max}+1)C'$ outputs.
These outputs are shared by all interaction layers.
Fixing $\omega_n=n\pi/r_c$ gives the basis \cref{eq:dpa4-radial-basis} used in \cref{lem:dpa4-radial}.

The scalar embedding modulates a type table $\mathbf T\in\RR^{Z\times C'}$ by a scale and a shift:
\begin{equation}\label{eq:app-dpa4-embed}
  \begin{aligned}
    x^{[0],(0)}_{j,0,c}&=\gamma_{j,c}\mathbf T_{z_j,c}+\beta_{j,c},\\
    \gamma_j&=1+e^{\lambda_\alpha}\tanh\bigl(N_0(W_\alpha\calI_j)\bigr),\qquad
    \beta_j=e^{\lambda_\beta}\tanh\bigl(N_0(W_\beta\calI_j)\bigr).
  \end{aligned}
\end{equation}
The geometric input $\calI_j$ collects inner products between weighted neighbor sums of $(1,\Delta\bm r_{jk}/\bar r_{jk})$, where $\bar r_{jk}=(r_{jk}^2+\epsilon_r^2)^{1/2}$ with fixed $\epsilon_r>0$.
The weights depend on distances and atom types, so these inner products are rotation invariant.
The matrices $W_\alpha,W_\beta$ and strengths $\lambda_\alpha,\lambda_\beta$ are learnable, and $N_0$ is the scalar normalization \cref{eq:app-dpa4-scalar-norm}.
Positive-degree features are initialized by
\begin{equation}\label{eq:app-dpa4-embed-l}
  x^{[0],(l)}_{j,\cdot,c}
    =n_j\sum_{k\in\calN(j)}Y^{(l)}(\hat{\bm r}_{jk})
      \bigl[\mathrm{MLP}^{\mathrm{rad}}(\phi(r_{jk}))\bigr]_{l,c},\qquad
  n_j=\Bigl(\epsilon_n+\sum_{k\in\calN(j)}s_5(r_{jk})^2\Bigr)^{-1/2},
\end{equation}
for $l\ge1$ and fixed $\epsilon_n>0$.
Setting $W_\alpha=W_\beta=0$ and the positive-degree radial outputs to zero leaves only the type table, which supplies the initialization in \cref{sec:dpa4-layer-approx}.

\subsection{Normalization and local activation}
\label{app:dpa4-normalization}

For a feature $a$ with $C_a$ channels and degrees up to $l_{\max}$, let
$\bar a=C_a^{-1}\sum_c a^{(0)}_{0,c}$.
The equivariant normalization is
\begin{equation}\label{eq:app-dpa4-norm}
  \begin{aligned}
    N(a)^{(l)}_{m,c}
      &=\gamma_{l,c}\frac{a^{(l)}_{m,c}-\delta_{l0}\bar a}
        {\sqrt{\sigma^2(a)+\epsilon_{\mathrm N}}}+\delta_{l0}\beta_c,\\
    \sigma^2(a)
      &=\sum_{l,m,c}\frac{|a^{(l)}_{m,c}-\delta_{l0}\bar a|^2}
        {(2l+1)(l_{\max}+1)C_a} .
  \end{aligned}
\end{equation}
The gains $\gamma_{l,c}$ and shifts $\beta_c$ are learned, and $\epsilon_{\mathrm N}>0$ is fixed.
This normalization is used on the atomic residual branches and, at full bond-frame order, inside the local message stack.
Each occurrence has its own parameters.
The choice $\gamma_{l,c}=\sqrt{\epsilon_{\mathrm N}}$ and $\beta_c=0$ gives \cref{eq:dpa4-norm-small} on features of zero scalar mean.

The scalar branches of the embedding and attention use RMS normalization without mean subtraction:
\begin{equation}\label{eq:app-dpa4-scalar-norm}
  N_0(v)_c=\frac{\gamma_c v_c}
    {\sqrt{C_a^{-1}\sum_{d=1}^{C_a}v_d^2+\epsilon_0}},
  \qquad v\in\RR^{C_a},\quad \epsilon_0>0 .
\end{equation}
The gains are learned and there is no bias, so $N_0(0)=0$.
Write $\sigma(u)=(1+e^{-u})^{-1}$ and $\mathrm{SiLU}(u)=u\sigma(u)$.
The local equivariant activation is
\begin{equation}\label{eq:app-dpa4-activation}
  \calG(a)^{(0)}_{0,c}=\mathrm{SiLU}(a^{(0)}_{0,c}),\qquad
  \calG(a)^{(l)}_{m,c}
    =a^{(l)}_{m,c}\,\sigma\Bigl(\sum_{c'}G^{(l)}_{c'c}a^{(0)}_{0,c'}\Bigr),\quad l\ge1,
\end{equation}
with learned gate matrices $G^{(l)}$.
Zero gate matrices give the activation used in the proof of \cref{prop:dpa4-layer}.

\subsection{Bond messages and attention}
\label{app:dpa4-message}

An interaction layer uses a hidden width $H\ge C'$.
A linear map $L^{\mathrm{rad}}:\RR^{C'}\to\RR^H$, applied at each degree, gives $\tilde\rho_{jk}=L^{\mathrm{rad}}\rho_{jk}$.
Let $D_H(R)=\bigoplus_{l\le l_{\max}}D^{(l)}(R)\otimes\mathrm{Id}_H$.
An equivariant linear pre-mixer $L^{\mathrm{pre}}$ widens the atomic feature, and a rotation $R_{jk}\hat{\bm r}_{jk}=\bm e_z$ places it in the bond frame:
\begin{equation}\label{eq:app-dpa4-transport}
  x_{jk}=D_H(R_{jk})L^{\mathrm{pre}}x_k .
\end{equation}
The radial kernel then mixes degrees at fixed order and channel,
\begin{equation}\label{eq:app-dpa4-kernel}
  \begin{aligned}
    (x_{jk})^{(l)}_{m,c}
      &\leftarrow\sum_{l'=|m|}^{l_{\max}}
        \calK_{l,l',|m|,c}(\tilde\rho_{jk})(x_{jk})^{(l')}_{m,c},\\
    \calK_{l,l',|m|,c}(\tilde\rho)
      &=\sum_{r=1}^{R}\bigl[K_{l,l',|m|}(\tilde\rho)\bigr]_r B_{r,c} .
  \end{aligned}
\end{equation}
Here $K_{l,l',|m|}:\RR^{(l_{\max}+1)\times H}\to\RR^R$ is linear, $B\in\RR^{R\times H}$ is learned, and $R$ is the kernel rank.
In particular, $R=H$ and $B=I$ allow an independent radial functional in each hidden channel.

The bond feature passes through $S\ge1$ local residual steps, with $\odot$ denoting channelwise multiplication,
\begin{equation}\label{eq:app-dpa4-local}
  x_{jk}\leftarrow x_{jk}
    +\Lambda_s\odot\calG_s\bigl(L^{SO2}_sN_s(x_{jk})\bigr),
  \qquad s=1,\ldots,S .
\end{equation}
The maps $N_s$ and $\calG_s$ have the forms \cref{eq:app-dpa4-norm,eq:app-dpa4-activation}.
The learned scale $\Lambda_s$ acts channelwise and is shared across degrees and orders.
The map $L^{SO2}_s$ mixes degrees and channels freely at order $0$.
At each nonzero order it connects matching pairs $\pm m$ by blocks $\alpha I+\beta J$, with $J$ as in \cref{lem:dpa4-bond-frame}.
Scalar biases are set to zero in the simulation.
After the local steps, rotation back gives $m_{jk}=D_H(R_{jk})^{\top}x_{jk}$.
With one focus stream its competition weight is $1$.

Let $x'_j\in\RR^H$ be the degree-$0$ slice of $L^{\mathrm{pre}}x_j$.
The attention logit and weight are
\begin{equation}\label{eq:app-dpa4-attention}
  \begin{aligned}
    \ell_{jk}
      &=\frac{\langle QN_0(x'_j),KN_0(x'_k)\rangle}{\sqrt d}
        +\langle W^{\mathrm{rb}},\tilde\rho_{jk,0}\rangle,\\
    w_{jk}
      &=\frac{s_5(r_{jk})^2 e^{\ell_{jk}}}
        {\Xi+\sum_{k'\in\calN(j)}s_5(r_{jk'})^2e^{\ell_{jk'}}},
      \qquad \Xi=\epsilon_{\mathrm{att}}+\log(1+e^\zeta).
  \end{aligned}
\end{equation}
The query and key maps $Q,K:\RR^H\to\RR^d$, the radial weights $W^{\mathrm{rb}}$, and $\zeta\in\RR$ are learned.
The constant $\epsilon_{\mathrm{att}}>0$ is fixed.
Zeroing $Q,K,W^{\mathrm{rb}}$ gives \cref{eq:dpa4-attention-zero}, and increasing $\zeta$ makes $\Xi$ arbitrarily large.
With the aggregate $A_j=\sum_{k\in\calN(j)}w_{jk}m_{jk}$ and gate
$G_j=\sigma(\langle W^{\mathrm{og}},N_0(x'_j)\rangle)$, the message operator is
\begin{equation}\label{eq:app-dpa4-conv}
  \calC(x)_j=L^{\mathrm{post}}\Bigl[G_jA_j
    +\calB_{\mathrm W}(G_jA_j,x_j)\Bigr] .
\end{equation}
The linear post-mixer $L^{\mathrm{post}}$ maps the hidden width $H$ back to $C'$ separately in each degree.
The two atomic residual updates are those of \cref{eq:dpa4-layer}.

\subsection{Wigner blocks and the energy head}
\label{app:dpa4-readout}

The block $\calB_{\mathrm W}(x,y)$ uses independent linear lifts of its two inputs, whose widths may differ.
With $F$ field pairs indexed by $a$, its lifts, synthesis, and analysis have the forms \cref{eq:wbb-lift,eq:wbb-synth,eq:wbb-analysis}, with coefficients $u$ from $x$ and $v$ from $y$.
The degrees extend to $l_{\max}$ and the frame sets are truncated at $k_{\max}$.
The analysis uses the exact quadrature of Appendix~\ref{app:wbb-quadrature}, with $\lambda=l_{\max}$.

Set $\alpha_a=u^{(0)}_{0,0,a}$, $\beta_a=v^{(0)}_{0,0,a}$, and $u^0=(\alpha,\beta)\in\RR^{2F}$.
The left scalar coefficients enter the nonlinear gate, and the right scalar coefficients supply its value factor.
Channel mixing and scalar gating transform the analyzed coefficients into
\begin{equation}\label{eq:app-dpa4-wbb-scalar}
  \tilde z^{(l)}_{m,k,a}
    =\sigma(W_{\mathrm g}u^0)_a\sum_{b=1}^{F}(W_{\mathrm m})_{ab}z^{(l)}_{m,k,b}
      +\delta_{l0}\delta_{m0}\delta_{k0}\,\mathrm{SiLU}(\alpha_a)\beta_a .
\end{equation}
Here $W_{\mathrm m}\in\RR^{F\times F}$ and $W_{\mathrm g}\in\RR^{F\times2F}$ are learned.
An equivariant linear output map contracts frames and channels:
\begin{equation}\label{eq:app-dpa4-wbb-out}
  \calB_{\mathrm W}(x,y)=L^{\mathrm{ch}}_{\mathrm{out}}[\tilde z] .
\end{equation}
Its output width is $H$ inside \cref{eq:app-dpa4-conv} and $C'$ in the atomic and readout blocks.
The block has no residual term of its own.
The readout applies the residual updates $x\mapsto x+\calB_{\mathrm W}(x,x)$ without normalization, as in \cref{sec:dpa4}.

For the final scalar slice $s_j$, the MLP in the energy head \cref{eq:dpa4-energy} is
\begin{equation}\label{eq:app-dpa4-head}
  \mathrm{MLP}(s)=W_2\bigl[(W_{\mathrm v}s)\odot\mathrm{SiLU}(W_{\mathrm g}s)\bigr] .
\end{equation}
These head matrices are independent of the block parameters and carry no biases.
Together with the constant channel, this head supplies the exact scalar output constructed in \cref{prop:dpa4-readout}.

\section{Exact quadrature for Wigner analysis}
\label{app:wbb-quadrature}

The analysis integral \cref{eq:wbb-analysis} can be evaluated by a finite sum over rotations.
We construct such a rule by combining quadrature on the sphere with equally spaced rotations about the pole.
Let $\sigma$ be the surface measure on $S^2$ normalized to total mass $1$, and let $R_z(\phi)$ denote rotation through angle $\phi$ about $\bm e_z$.

\begin{lemma}
\label{lem:wbb-quadrature}
Fix integers $\lambda,k_{\max}\ge0$.
Suppose that nodes $\bm n_q\in S^2$ and weights $w_q$, $q=1,\ldots,Q$, integrate all polynomials of degree at most $3\lambda$ exactly against $\sigma$.
Choose an integer $N\ge3k_{\max}+1$ and rotations $s_q\in SO(3)$ satisfying $s_q\bm e_z=\bm n_q$.
Then
\begin{equation}\label{eq:app-wbb-quadrature}
  \int_{SO(3)}F(g)\,dg
  =\sum_{q=1}^{Q}\frac{w_q}{N}\sum_{p=0}^{N-1}
    F\bigl(s_qR_z(2\pi p/N)\bigr)
\end{equation}
for every linear combination $F$ of products
$D^{(l_1)}_{m_1k_1}D^{(l_2)}_{m_2k_2}D^{(l_3)}_{m_3k_3}$
with $0\le l_i\le\lambda$, $|m_i|\le l_i$, and $|k_i|\le\min(k_{\max},l_i)$.
\end{lemma}

\begin{proof}
For $\bm n\in S^2$, choose any $s(\bm n)\in SO(3)$ with $s(\bm n)\bm e_z=\bm n$ and define
\[
  \bar F(\bm n)=\frac{1}{2\pi}\int_0^{2\pi}
    F\bigl(s(\bm n)R_z(\phi)\bigr)\,d\phi .
\]
Changing $s(\bm n)$ amounts to a shift of $\phi$, so $\bar F$ is independent of this choice.
Normalized Haar measure splits into the normalized surface measure and the uniform angle measure~\cite{varshalovich1988}, giving
\begin{equation}\label{eq:app-haar-factor}
  \int_{SO(3)}F(g)\,dg=\int_{S^2}\bar F(\bm n)\,d\sigma(\bm n).
\end{equation}

In real spherical coordinates, $D^{(l)}(R_z(\phi))$ fixes coordinate $0$ and rotates each coordinate pair $\pm k$ through angle $k\phi$.
The identity $D^{(l)}(sR_z(\phi))=D^{(l)}(s)D^{(l)}(R_z(\phi))$ therefore makes
$\phi\mapsto D^{(l)}_{mk}(sR_z(\phi))$ a trigonometric polynomial of degree at most $|k|$.
Thus $F(sR_z(\phi))$ has trigonometric degree at most $3k_{\max}<N$.
For $0<|j|<N$, the identity $N^{-1}\sum_{p=0}^{N-1}e^{2\pi ijp/N}=0$ shows that the discrete angular average is exact.
Consequently, the inner average on the right of \cref{eq:app-wbb-quadrature} equals $\bar F(\bm n_q)$.

Applying the product identity \cref{eq:wigner-product} twice expands $F$ into Wigner functions $D^{(J)}_{MK}$ with $J\le3\lambda$.
Averaging their right rotation $R_z(\phi)$ removes all terms with $K\ne0$, so $\bar F$ is a linear combination of $D^{(J)}_{M0}(s(\bm n))$.
At the pole, $Y^{(J)}(\bm e_z)=y_Je_0$ with $y_J\ne0$, where $e_0\in V_J$ is the coordinate vector of index $0$.
Harmonic equivariance then gives
\[
  D^{(J)}_{M0}(s(\bm n))=y_J^{-1}Y^{(J)}_M(\bm n).
\]
Thus $\bar F$ is a sum of spherical harmonics of degrees at most $3\lambda$.
These are restrictions of polynomials of the same degrees to $S^2$.
The sphere rule integrates $\bar F$ exactly, and \cref{eq:app-haar-factor} proves \cref{eq:app-wbb-quadrature}.
\end{proof}

Each integrand in \cref{eq:wbb-analysis} is a sum of products covered by the lemma.
One factor comes from the analysis and two from the synthesized fields.
For the DPA4 blocks of Appendix~\ref{app:dpa4-readout}, take $\lambda=l_{\max}$.
This gives the exact quadrature required in \cref{cor:dpa4}.

A finite sphere rule of the required precision exists for every $\lambda$.
Choose a Gauss--Legendre rule on $[-1,1]$ with $n_u$ nodes $u_a$ and weights $\omega_a$, where $2n_u-1\ge3\lambda$.
Choose an integer $n_{\mathrm{az}}>3\lambda$ and angles $\psi_b=2\pi b/n_{\mathrm{az}}$, $b=0,\ldots,n_{\mathrm{az}}-1$.
The sphere nodes and weights are
\begin{equation}\label{eq:app-sphere-rule}
  \bm n_{ab}
    =\bigl(\sqrt{1-u_a^2}\cos\psi_b,\,
           \sqrt{1-u_a^2}\sin\psi_b,\,u_a\bigr),
  \qquad w_{ab}=\frac{\omega_a}{2n_{\mathrm{az}}}.
\end{equation}
For a Cartesian polynomial of degree at most $3\lambda$, substitution of these spherical coordinates gives a trigonometric polynomial in $\psi$ of degree at most $3\lambda$.
The equally spaced azimuths integrate it exactly.
The azimuthal average of a monomial $x^\alpha y^\beta z^\gamma$ is zero or a constant multiple of $(1-u^2)^{(\alpha+\beta)/2}u^\gamma$, with $\alpha+\beta$ even in the latter case.
It is therefore a polynomial in $u$ of degree at most $\alpha+\beta+\gamma\le3\lambda$, which the Gauss--Legendre rule integrates exactly.
The factor $1/2$ in \cref{eq:app-sphere-rule} accounts for the normalized surface measure.

The Lebedev rules used by DPA4 also serve whenever their precision reaches $3l_{\max}$.
For arbitrary $l_{\max}$, the product construction \cref{eq:app-sphere-rule} supplies a suitable rule.
The family in \cref{cor:dpa4} allows any sphere rule with this precision, so the maximal degree is not limited by the tabulated Lebedev orders.

\section*{Acknowledgments}
Pingbing Ming is supported by the National Key R\&D Program of China (Grant no.~2024YFA1012502) and by National Natural Science Foundation of China (Grant No.~12371438).
Han Wang is supported by the National Key R\&D Program of China (Grant No.~2022YFA1004300) and by the National Natural Science Foundation of China (Grants No.~12525113 and No.~12561160120).

The authors used generative AI tools to assist with proof development and checking, literature searches and reference verification, and manuscript drafting and editing.
The authors reviewed and revised the AI-assisted material and checked the mathematical proofs.
The authors assume responsibility for all content.

\bibliographystyle{siamplain}
\bibliography{refs}

\clearpage
\let\section\arxivmainsection
\makeatletter
\let\H@refstepcounter\arxivmainrefstepcounter
\let\Hy@chapapp\arxivmainchapapp
\makeatother
\setcounter{section}{0}
\setcounter{subsection}{0}
\setcounter{equation}{0}
\setcounter{theorem}{0}
\renewcommand{\thesection}{S\arabic{section}}
\renewcommand{\theequation}{\thesection.\arabic{equation}}
\renewcommand{\thetheorem}{\thesection.\arabic{theorem}}
\setlength{\emergencystretch}{3em}
\pdfbookmark[0]{Supplementary Materials}{supplement}
\begin{center}
  {\bfseries\large Supplementary Materials}
\end{center}

\ifdefined\arxivbuild\else
\maketitle
\fi

\noindent This supplement uses the notation and abbreviations of the main text.
Sections, equations, and results of this supplement are numbered with the prefix S.
References without the prefix point to the main text.
We give detailed coordinate and coefficient calculations for \cref{sec:blocks,prop:wbb-full}.
The frame-truncation proof and the DPA4 simulation proofs are given in \cref{sec:uat-dpa4}.
The DPA4 operators and exact quadrature are specified in Appendices~\ref{app:dpa4} and~\ref{app:wbb-quadrature}.

%%%%%%%%%%%%%%%%%%%%%%%%%%%%%%%%%%%%%%%%%%%%%%%%%%%%%%%%%%%%%%%%%%
\section{Clebsch--Gordan conventions and the Wigner product identity}
\label{sm:cg}
%%%%%%%%%%%%%%%%%%%%%%%%%%%%%%%%%%%%%%%%%%%%%%%%%%%%%%%%%%%%%%%%%%

The main text works in real coordinates on every $V_l$, in which the Wigner matrices are real and orthogonal and the Clebsch--Gordan coefficients are real.
The standard tables and identities are stated in complex coordinates~\cite{varshalovich1988}.
This section expands the coordinate conventions of \cref{sec:blocks} and derives the real Wigner identities \cref{eq:wigner-product,eq:wigner-orth}.
Objects in complex coordinates carry a tilde.

\subsection{Real and complex spherical bases}
\label{sm:cg-bases}

Let $\tilde Y^{(l)}_m$, $|m|\le l$, be the complex spherical harmonics of degree $l$ in the Condon--Shortley phase convention~\cite{varshalovich1988}.
They are orthonormal in $L^2(S^2)$ with respect to the surface measure $d\Omega$, of total mass $4\pi$.
They satisfy $\tilde Y^{(l)}_{-m}=(-1)^m\,\overline{\tilde Y^{(l)}_{m}}$.
In spherical coordinates, $\tilde Y^{(l)}_m$ is a function of the polar angle times $e^{im\phi}$.
Two consequences are used below.
A rotation $R_z(\alpha)$ about $\bm e_z$ shifts the azimuth, so $\tilde Y^{(l)}_m(R_z(\alpha)\hat{\bm r})=e^{im\alpha}\,\tilde Y^{(l)}_m(\hat{\bm r})$.
At the pole only the harmonic with $m=0$ is nonzero, $\tilde Y^{(l)}_m(\bm e_z)=\delta_{m0}\,y_l$ with a constant $y_l\ne0$.

The real spherical harmonics of \cref{eq:real-harmonics} can be written as~\cite{ivanic1996rotation,blanco1997rotation}
\begin{equation}\label{eq:sm-real-harmonics}
  \begin{gathered}
    Y^{(l)}_0=\tilde Y^{(l)}_0,\\
    Y^{(l)}_{m}=\frac{(-1)^m\tilde Y^{(l)}_{m}+\tilde Y^{(l)}_{-m}}{\sqrt2},\qquad
    Y^{(l)}_{-m}=\frac{(-1)^m\tilde Y^{(l)}_{m}-\tilde Y^{(l)}_{-m}}{i\sqrt2},\qquad m=1,\ldots,l .
  \end{gathered}
\end{equation}
By the conjugation relation, $Y^{(l)}_{m}=\sqrt2\,(-1)^m\operatorname{Re}\tilde Y^{(l)}_m$ and $Y^{(l)}_{-m}=\sqrt2\,(-1)^m\operatorname{Im}\tilde Y^{(l)}_m$ for $m\ge1$, so the real harmonics are real functions.
We write \cref{eq:sm-real-harmonics} as $Y^{(l)}=U^{(l)}\tilde Y^{(l)}$ with the matrix $U^{(l)}$ that has the entry $1$ at $(0,0)$ and, for $m\ge1$, the block
\begin{equation}\label{eq:sm-real-unitary}
  \begin{pmatrix} U^{(l)}_{m,m} & U^{(l)}_{m,-m}\\ U^{(l)}_{-m,m} & U^{(l)}_{-m,-m}\end{pmatrix}
  =\frac{1}{\sqrt2}\begin{pmatrix} (-1)^m & 1\\ -i(-1)^m & i\end{pmatrix}
\end{equation}
on the coordinates $m$ and $-m$, and zero elsewhere.
Each block is unitary, so $U^{(l)}$ is unitary and the real harmonics are orthonormal in $L^2(S^2)$.
The coordinate change mixes only each pair $m,-m$ and leaves coordinate $0$ unchanged.
The real components of a vector in $V_l$ transform under rotations as the column $Y^{(l)}$ of harmonic values, while its complex components transform as $\tilde Y^{(l)}$.
Thus a vector with complex coordinates $\tilde u$ has real coordinates $u=U^{(l)}\tilde u$.
These are components in the sense of the transformation rule, not the expansion coefficients of a function in the basis of the harmonics, which pass from one basis to the other by $\overline{U^{(l)}}$.
% (2026-09-17: reworded after review item B; the earlier sentence "the coordinates in the
%  basis of the real harmonics" suggested expansion coefficients, for which the matrix is
%  the conjugate of U.)

The span of the $\tilde Y^{(l)}_m$ over $|m|\le l$ is invariant under rotations.
Hence for every $R\in SO(3)$ there is a matrix $\tilde D^{(l)}(R)$ with $\tilde Y^{(l)}(R\hat{\bm r})=\tilde D^{(l)}(R)\,\tilde Y^{(l)}(\hat{\bm r})$, and $R\mapsto\tilde D^{(l)}(R)$ is a homomorphism~\cite{varshalovich1988}.
The matrix is unitary, since the harmonics are orthonormal and the surface measure is rotation invariant.
The real Wigner matrix is
\begin{equation}\label{eq:sm-real-wigner}
  D^{(l)}(R)=U^{(l)}\,\tilde D^{(l)}(R)\,U^{(l)\dagger},
  \qquad\text{so that}\qquad
  Y^{(l)}(R\hat{\bm r})=D^{(l)}(R)\,Y^{(l)}(\hat{\bm r}) .
\end{equation}
Its entries $D^{(l)}_{mk}(R)=\int_{S^2}Y^{(l)}_m(R\hat{\bm r})\,Y^{(l)}_k(\hat{\bm r})\,d\Omega(\hat{\bm r})$ are real, since the real harmonics are real and orthonormal, and a real unitary matrix is orthogonal.
This is the matrix $D^{(l)}$ of the main text, the real orthogonal rotation matrix in the basis of real spherical harmonics of~\cite{ivanic1996rotation}, obtained from the complex one as in~\cite{blanco1997rotation}.

The following two formulas give the rotation and pole values used in Appendix~\ref{app:wbb-quadrature}.
For a rotation about the pole, the azimuth shift gives $Y^{(l)}_m(R_z(\alpha)\hat{\bm r})=\sqrt2(-1)^m\operatorname{Re}\bigl(e^{im\alpha}\tilde Y^{(l)}_m(\hat{\bm r})\bigr)=\cos(m\alpha)\,Y^{(l)}_m(\hat{\bm r})-\sin(m\alpha)\,Y^{(l)}_{-m}(\hat{\bm r})$ for $m\ge1$.
The imaginary part gives $Y^{(l)}_{-m}(R_z(\alpha)\hat{\bm r})=\sin(m\alpha)\,Y^{(l)}_m(\hat{\bm r})+\cos(m\alpha)\,Y^{(l)}_{-m}(\hat{\bm r})$.
By \cref{eq:sm-real-wigner}, the matrix $D^{(l)}(R_z(\alpha))$ therefore has the entry $1$ at $(0,0)$ and, for $m\ge1$, the block
\begin{equation}\label{eq:sm-rz-block}
  \begin{pmatrix} \cos m\alpha & -\sin m\alpha\\ \sin m\alpha & \cos m\alpha\end{pmatrix}
\end{equation}
on the coordinates $m$ and $-m$, and zero elsewhere.
For the pole value, \cref{eq:sm-real-wigner} at $\hat{\bm r}=\bm e_z$ gives $Y^{(l)}(s\bm e_z)=D^{(l)}(s)\,Y^{(l)}(\bm e_z)$.
Since $U^{(l)}$ leaves coordinate $0$ unchanged, $Y^{(l)}(\bm e_z)=U^{(l)}\tilde Y^{(l)}(\bm e_z)=y_l\,e_0$, where $e_0$ is the coordinate vector of index $0$.
Hence for every $s\in SO(3)$ and $|m|\le l$
\begin{equation}\label{eq:sm-pole}
  Y^{(l)}_m(s\,\bm e_z)=y_l\,D^{(l)}_{m0}(s) .
\end{equation}
% Superseded form: the two consequences were stated as Lemma [Rotations about the pole] (label sm:lem-pole)
% with a separate proof. Both follow in one line from the definitions, so they are stated in the text.

\subsection{Clebsch--Gordan coefficients and their symmetries}
\label{sm:cg-symmetries}

Let $l_1,l_2\ge0$ and let $L$ lie in the triangle range $|l_1-l_2|\le L\le l_1+l_2$.
The complex Clebsch--Gordan coefficients $\tilde C^{(L,M)}_{(l_1,m_1)(l_2,m_2)}$ in the Condon--Shortley convention are real numbers~\cite{varshalovich1988}.
They vanish unless $M=m_1+m_2$, and they satisfy the orthogonality relations
\begin{equation}\label{eq:sm-cg-orth}
\begin{gathered}
  \sum_{m_1,m_2}\tilde C^{(L,M)}_{(l_1,m_1)(l_2,m_2)}\,\tilde C^{(L',M')}_{(l_1,m_1)(l_2,m_2)}=\delta_{LL'}\delta_{MM'},\\
  \sum_{L,M}\tilde C^{(L,M)}_{(l_1,m_1)(l_2,m_2)}\,\tilde C^{(L,M)}_{(l_1,m_1')(l_2,m_2')}=\delta_{m_1m_1'}\delta_{m_2m_2'} ,
\end{gathered}
\end{equation}
where the second sum runs over the triangle range.
Under the exchange of the two factors
\begin{equation}\label{eq:sm-cg-exchange}
  \tilde C^{(L,M)}_{(l_2,m_2)(l_1,m_1)}=(-1)^{l_1+l_2-L}\,\tilde C^{(L,M)}_{(l_1,m_1)(l_2,m_2)} ,
\end{equation}
and under the reversal of all magnetic indices,
\begin{equation}\label{eq:sm-cg-reversal}
  \tilde C^{(L,-M)}_{(l_1,-m_1)(l_2,-m_2)}=(-1)^{l_1+l_2-L}\,\tilde C^{(L,M)}_{(l_1,m_1)(l_2,m_2)} .
\end{equation}
The corresponding Wigner product identity is
\begin{equation}\label{eq:sm-cg-series}
  \tilde D^{(l_1)}_{m_1k_1}(g)\,\tilde D^{(l_2)}_{m_2k_2}(g)
  =\sum_{L=|l_1-l_2|}^{l_1+l_2}\ \sum_{|M|,|K|\le L}
    \tilde C^{(L,M)}_{(l_1,m_1)(l_2,m_2)}\,\tilde C^{(L,K)}_{(l_1,k_1)(l_2,k_2)}\,\tilde D^{(L)}_{MK}(g) .
\end{equation}
The identity also holds when every Wigner matrix is replaced by its complex conjugate or transpose.
Conjugation leaves the real coefficients unchanged, while transposition interchanges row and column indices.

We arrange the coefficients as a matrix.
For fixed $(l_1,l_2)$ let $\tilde C$ be the matrix with row index $(m_1,m_2)$ and column index $(L,M)$, and let $\tilde C_L$ be the block of the columns with the given $L$, a matrix of size $(2l_1+1)(2l_2+1)\times(2L+1)$.
The relations \cref{eq:sm-cg-orth} say that $\tilde C$ is a real orthogonal matrix, and the series \cref{eq:sm-cg-series} reads
\begin{equation}\label{eq:sm-cg-series-matrix}
\begin{gathered}
  \tilde D^{(l_1)}(g)\otimes\tilde D^{(l_2)}(g)=\tilde C\,\Bigl(\bigoplus_{L}\tilde D^{(L)}(g)\Bigr)\,\tilde C^{T},\\
  \text{equivalently}\qquad
  \bigl(\tilde D^{(l_1)}(g)\otimes\tilde D^{(l_2)}(g)\bigr)\,\tilde C_L=\tilde C_L\,\tilde D^{(L)}(g) .
\end{gathered}
\end{equation}
The second form says that $\tilde C_L$ is an equivariant embedding of $V_L$ into $V_{l_1}\otimes V_{l_2}$ with orthonormal columns.
The tensor product contains $V_L$ exactly once for $L$ in the triangle range and not at all outside it~\cite{varshalovich1988}.
By Schur's lemma, the space of equivariant linear maps $V_L\to V_{l_1}\otimes V_{l_2}$ over $\CC$ is therefore one-dimensional in the triangle range and zero outside it.

Transporting $\tilde C$ to real coordinates by the matrices $U^{(l)}$ gives a Clebsch--Gordan series for the real Wigner matrices, but the transported coefficients are complex in general, since $U^{(l)}$ has imaginary entries.
The following lemma shows that a phase for each path makes the coefficients themselves real, which is the form used in the main text.

\begin{lemma}[Real Clebsch--Gordan coefficients]
\label{sm:lem-real-cg}
Let $(l_1,l_2,L)$ be a path in the triangle range, and let $D^{(l)}$ be the real Wigner matrices of \cref{eq:sm-real-wigner}.
There is a real matrix $C_L$ of size $(2l_1+1)(2l_2+1)\times(2L+1)$ with orthonormal columns such that
\begin{equation}\label{eq:sm-cg-real-embed}
  \bigl(D^{(l_1)}(g)\otimes D^{(l_2)}(g)\bigr)\,C_L=C_L\,D^{(L)}(g)\qquad\text{for all }g\in SO(3),
\end{equation}
and $C_L$ is unique up to sign.
The matrix $C$ with the blocks $C_L$, $|l_1-l_2|\le L\le l_1+l_2$, as columns is orthogonal and satisfies
\begin{equation}\label{eq:sm-cg-real-series}
  D^{(l_1)}(g)\otimes D^{(l_2)}(g)=C\,\Bigl(\bigoplus_{L}D^{(L)}(g)\Bigr)\,C^{T} .
\end{equation}
The entries $C^{(L,M)}_{(l_1,m_1)(l_2,m_2)}=(C_L)_{(m_1,m_2),M}$ are the Clebsch--Gordan coefficients of the main text.
They define the equivariant bilinear map $(u,v)\mapsto(u\otimes v)^{(L)}$ with coordinates $(u\otimes v)^{(L)}_M=\sum_{m_1,m_2}C^{(L,M)}_{(l_1,m_1)(l_2,m_2)}u_{m_1}v_{m_2}$.
Every equivariant bilinear map $V_{l_1}\times V_{l_2}\to V_L$ is a real multiple of this map.
For $L$ outside the triangle range, no nonzero equivariant bilinear map exists.
\end{lemma}

\begin{proof}
Transport the complex embedding to real coordinates,
\begin{equation}\label{eq:sm-cg-transport}
  G_L=\bigl(U^{(l_1)}\otimes U^{(l_2)}\bigr)\,\tilde C_L\,U^{(L)\dagger} .
\end{equation}
By \cref{eq:sm-real-wigner,eq:sm-cg-series-matrix},
$(D^{(l_1)}\otimes D^{(l_2)})G_L=(U^{(l_1)}\otimes U^{(l_2)})(\tilde D^{(l_1)}\otimes\tilde D^{(l_2)})\tilde C_LU^{(L)\dagger}=(U^{(l_1)}\otimes U^{(l_2)})\tilde C_L\tilde D^{(L)}U^{(L)\dagger}=G_LD^{(L)}$,
so $G_L$ is an equivariant embedding for the real matrices, with orthonormal columns since the three factors have orthonormal columns.
The real matrices are unchanged by complex conjugation, so the entrywise conjugate $\overline{G_L}$ is equivariant as well.
By the one-dimensionality of the space of equivariant maps there is a scalar $\zeta$ with $\overline{G_L}=\zeta G_L$.
Conjugating once more gives $G_L=\bar\zeta\,\overline{G_L}=|\zeta|^2G_L$, so $|\zeta|=1$.
Write $\zeta=e^{2i\alpha}$ and set $C_L=e^{i\alpha}G_L$.
Then $\overline{C_L}=e^{-i\alpha}\,\overline{G_L}=e^{i\alpha}G_L=C_L$, so $C_L$ is real, and it inherits \cref{eq:sm-cg-real-embed} and the orthonormal columns from $G_L$.
If $C_L'$ is another real matrix with the same properties, it is a complex multiple of $G_L$ by one-dimensionality, hence a real multiple of the real matrix $C_L$, and the unit norm of the columns leaves the sign.

The matrix $G=(U^{(l_1)}\otimes U^{(l_2)})\,\tilde C\,\bigoplus_LU^{(L)\dagger}$ is unitary as a product of unitary matrices.
The matrix $C$ differs from $G$ by a phase on each column block, so it is also unitary.
Since $C$ is real, it is orthogonal.
Combining \cref{eq:sm-cg-real-embed} for all $L$ and multiplying on the right by $C^T$ gives \cref{eq:sm-cg-real-series}, since $CC^T=\mathrm{Id}$.

Transposing \cref{eq:sm-cg-real-embed} and using the orthogonality of the real Wigner matrices gives $C_L^{T}\,(D^{(l_1)}\otimes D^{(l_2)})=D^{(L)}\,C_L^{T}$.
Thus the map $u\otimes v\mapsto C_L^T(u\otimes v)$ is equivariant, and its coordinate $M$ is the sum stated in the lemma.
An equivariant bilinear map $V_{l_1}\times V_{l_2}\to V_L$ is an equivariant linear map $B:V_{l_1}\otimes V_{l_2}\to V_L$.
Its transpose $B^T$ is an equivariant map $V_L\to V_{l_1}\otimes V_{l_2}$ by the same orthogonality, hence a complex multiple of $G_L$, hence a real multiple of $C_L$ when $B$ is real.
Outside the triangle range the space of such maps is zero.
\end{proof}

The lemma fixes the coefficients $C$ of the main text up to the sign of each path $(l_1,l_2,L)$, which does not enter the identities below, where $C$ appears twice.
The selection rule $M=m_1+m_2$ does not survive the passage to real coordinates, as the cross product $1\otimes1\to1$ shows.
The proof also gives the statement used in \cref{lem:frame-truncation}, that the real coupling of a path, extended complex-linearly, is a nonzero multiple of the complex coupling transported to the same coordinates.

\subsection{Proof of the Wigner product identity}
\label{sm:cg-product}

The product identity \cref{eq:wigner-product} of the main text is \cref{eq:sm-cg-real-series} written entrywise.
The entry $((m_1,m_2),(k_1,k_2))$ of the left side is $D^{(l_1)}_{m_1k_1}(g)\,D^{(l_2)}_{m_2k_2}(g)$, and the same entry of the right side is
\begin{equation*}
\begin{split}
  &\sum_{L}\sum_{|M|,|K|\le L}(C_L)_{(m_1,m_2),M}\,D^{(L)}_{MK}(g)\,(C_L)_{(k_1,k_2),K}\\
  &\qquad=\sum_{L}\sum_{|M|,|K|\le L}C^{(L,M)}_{(l_1,m_1)(l_2,m_2)}\,C^{(L,K)}_{(l_1,k_1)(l_2,k_2)}\,D^{(L)}_{MK}(g),
\end{split}
\end{equation*}
which is \cref{eq:wigner-product}.

The orthogonality \cref{eq:wigner-orth} follows from its complex form~\cite{varshalovich1988},
\begin{equation}\label{eq:sm-wigner-orth-complex}
  (2l+1)\int_{SO(3)}\tilde D^{(l)}_{mk}(g)\,\overline{\tilde D^{(l')}_{m'k'}(g)}\,dg=\delta_{ll'}\delta_{mm'}\delta_{kk'} .
\end{equation}
Insert \cref{eq:sm-real-wigner} for $l=l'$,
\begin{equation*}
\begin{split}
  (2l+1)\int_{SO(3)}D^{(l)}_{mk}\,\overline{D^{(l)}_{m'k'}}\,dg
  &=\sum_{\mu,\kappa,\mu',\kappa'}U^{(l)}_{m\mu}\overline{U^{(l)}_{k\kappa}}\,\overline{U^{(l)}_{m'\mu'}}U^{(l)}_{k'\kappa'}\;\delta_{\mu\mu'}\delta_{\kappa\kappa'}\\
  &=\bigl(U^{(l)}U^{(l)\dagger}\bigr)_{mm'}\,\overline{\bigl(U^{(l)}U^{(l)\dagger}\bigr)_{kk'}}
  =\delta_{mm'}\delta_{kk'} ,
\end{split}
\end{equation*}
and for $l\ne l'$ the integral vanishes termwise.
Since the real Wigner matrices are real, the conjugation can be dropped, which is \cref{eq:wigner-orth}.

%%%%%%%%%%%%%%%%%%%%%%%%%%%%%%%%%%%%%%%%%%%%%%%%%%%%%%%%%%%%%%%%%%
\section{Coefficient calculations for Wigner bilinear blocks}
\label{sm:wbb-bb}
%%%%%%%%%%%%%%%%%%%%%%%%%%%%%%%%%%%%%%%%%%%%%%%%%%%%%%%%%%%%%%%%%%

We detail the product expansion and channel assignments in the proof of \cref{prop:wbb-full}.
The frame coefficients and quadrature are supplied by \cref{lem:frame-truncation,lem:wbb-quadrature} in the main text.

\subsection{Expansion of the analyzed product}
\label{sm:wbb-expand}

We derive \cref{eq:wbb-expand} by expanding the synthesized product and applying the analysis integral term by term.
Throughout, we use the real coordinates of \cref{sm:cg}.
For $u\in V_{l_1}$ and $v\in V_{l_2}$, the Clebsch--Gordan product is the bilinear map of \cref{sm:lem-real-cg}, with coordinates $(u\otimes v)^{(L)}_M=\sum_{m_1,m_2}C^{(L,M)}_{(l_1,m_1)(l_2,m_2)}u_{m_1}v_{m_2}$.
The Wigner functions satisfy the orthogonality relation \cref{eq:wigner-orth}.

Fix one field pair of the WBB of \cref{def:wbb} and drop its index $a$.
The synthesis \cref{eq:wbb-synth} writes the two fields as
\begin{equation*}
  \begin{aligned}
    \hat u(g)&=\sum_{l_1\le\lambda}\ \sum_{|m_1|\le l_1}\ \sum_{k_1\in\calS_{l_1}}u^{(l_1)}_{m_1,k_1}\,D^{(l_1)}_{m_1k_1}(g),\\
    \hat v(g)&=\sum_{l_2\le\lambda}\ \sum_{|m_2|\le l_2}\ \sum_{k_2\in\calS_{l_2}}v^{(l_2)}_{m_2,k_2}\,D^{(l_2)}_{m_2k_2}(g),
  \end{aligned}
\end{equation*}
and their pointwise product is the double sum
\begin{equation}\label{eq:sm-wbb-product-sum}
  \hat z(g)=\hat u(g)\hat v(g)=\sum_{l_1,l_2\le\lambda}\ \sum_{m_1,m_2}\ \sum_{k_1\in\calS_{l_1},\,k_2\in\calS_{l_2}}
  u^{(l_1)}_{m_1,k_1}\,v^{(l_2)}_{m_2,k_2}\,D^{(l_1)}_{m_1k_1}(g)\,D^{(l_2)}_{m_2k_2}(g).
\end{equation}

\begin{lemma}[Expansion of the product]
\label{sm:lem-wbb-expand}
The product \cref{eq:sm-wbb-product-sum} is a finite combination of Wigner functions of degrees $L\le2\lambda$,
\begin{equation}\label{eq:sm-wbb-product-expand}
  \hat z(g)=\sum_{L\le2\lambda}\ \sum_{|M|,|K|\le L}\pi^{(L)}_{MK}\,D^{(L)}_{MK}(g),
\end{equation}
whose coefficients are
\begin{equation}\label{eq:sm-wbb-product-coeff}
  \pi^{(L)}_{MK}
  =\sum_{\substack{l_1,l_2\le\lambda\\ |l_1-l_2|\le L\le l_1+l_2}}\
   \sum_{k_1\in\calS_{l_1},\,k_2\in\calS_{l_2}}
   C^{(L,K)}_{(l_1,k_1)(l_2,k_2)}\,
   \bigl(u^{(l_1)}_{\cdot,k_1}\otimes v^{(l_2)}_{\cdot,k_2}\bigr)^{(L)}_M .
\end{equation}
The analysis \cref{eq:wbb-analysis} returns these coefficients for the degrees of the hidden space,
\begin{equation}\label{eq:sm-wbb-analysis-coeff}
  z^{(l)}_{m,k}=\pi^{(l)}_{mk},\qquad l\le\lambda,\ |m|\le l,\ k\in\calS_l ,
\end{equation}
which is \cref{eq:wbb-expand}.
The analysis discards the coefficients of degrees $\lambda<L\le2\lambda$ and, for $L\le\lambda$, of frames $K\notin\calS_L$.
\end{lemma}

\begin{proof}
Every product of two Wigner functions in \cref{eq:sm-wbb-product-sum} expands by the product identity \cref{eq:wigner-product} into Wigner functions of the degrees $L$ in the triangle range of $(l_1,l_2)$,
\begin{equation*}
  D^{(l_1)}_{m_1k_1}(g)\,D^{(l_2)}_{m_2k_2}(g)
  =\sum_{L=|l_1-l_2|}^{l_1+l_2}\ \sum_{|M|,|K|\le L}
    C^{(L,M)}_{(l_1,m_1)(l_2,m_2)}\,C^{(L,K)}_{(l_1,k_1)(l_2,k_2)}\,D^{(L)}_{MK}(g),
\end{equation*}
and $L\le l_1+l_2\le2\lambda$.
Inserting the identity into \cref{eq:sm-wbb-product-sum} and collecting the coefficient of $D^{(L)}_{MK}$ gives \cref{eq:sm-wbb-product-expand} with
\begin{equation*}
  \pi^{(L)}_{MK}
  =\sum_{\substack{l_1,l_2\le\lambda\\ |l_1-l_2|\le L\le l_1+l_2}}\
   \sum_{k_1\in\calS_{l_1},\,k_2\in\calS_{l_2}}
   C^{(L,K)}_{(l_1,k_1)(l_2,k_2)}
   \sum_{m_1,m_2}C^{(L,M)}_{(l_1,m_1)(l_2,m_2)}\,u^{(l_1)}_{m_1,k_1}\,v^{(l_2)}_{m_2,k_2},
\end{equation*}
where the sum over $(l_1,l_2)$ is restricted to the pairs whose triangle range contains $L$, since the identity produces no other degrees.
The inner sum over $(m_1,m_2)$ is the coordinate $M$ of the Clebsch--Gordan product of the vectors $u^{(l_1)}_{\cdot,k_1}\in V_{l_1}$ and $v^{(l_2)}_{\cdot,k_2}\in V_{l_2}$, which gives \cref{eq:sm-wbb-product-coeff}.

For \cref{eq:sm-wbb-analysis-coeff}, insert \cref{eq:sm-wbb-product-expand} into the analysis \cref{eq:wbb-analysis} and exchange the finite sum with the integral,
\begin{equation*}
  z^{(l)}_{m,k}=\sum_{L\le2\lambda}\ \sum_{|M|,|K|\le L}\pi^{(L)}_{MK}\,(2l+1)\int_{SO(3)}D^{(l)}_{mk}(g)\,D^{(L)}_{MK}(g)\,dg
  =\pi^{(l)}_{mk},
\end{equation*}
by the orthogonality \cref{eq:wigner-orth}, which keeps only the term $L=l$, $M=m$, $K=k$.
The analysis is evaluated for $l\le\lambda$ and $k\in\calS_l$ only, so the remaining coefficients of \cref{eq:sm-wbb-product-expand} do not enter the output.
\end{proof}

Two remarks on the expansion.
First, for the discrete block the integral in \cref{eq:wbb-analysis} is replaced by the rule of Appendix~\ref{app:wbb-quadrature}.
The integrand $D^{(l)}_{mk}\,\hat z$ is a combination of products of three Wigner functions of degrees at most $\lambda$ and frames in the frame sets.
Thus \cref{lem:wbb-quadrature} evaluates every integral in the proof exactly, and \cref{eq:sm-wbb-analysis-coeff} holds unchanged.
Second, each summand in \cref{eq:sm-wbb-product-coeff} is an equivariant Clebsch--Gordan product of lifted vectors multiplied by a scalar frame coefficient.
Expanding the linear lifts in \cref{eq:wbb-lift} expresses these products as linear combinations of the input-channel couplings in \cref{eq:bb-product}.
This supplies the quadratic part of the BB in the first claim of \cref{prop:wbb-full}.
A frame pair whose factor $C^{(l,k)}_{(l_1,k_1)(l_2,k_2)}$ vanishes contributes nothing to the output frame $k$.
For every coupling path, \cref{lem:frame-truncation} supplies a pair with $|k_1|,|k_2|\le1$ whose factor is nonzero at the output frame $k=0$.

\subsection{Channel bookkeeping for the converse}
\label{sm:wbb-bookkeeping}

This subsection writes out the parameters of the WBB that the converse of \cref{prop:wbb-full} assigns to a given bilinear block, and verifies with \cref{eq:sm-wbb-product-coeff} that the two blocks agree.
Let $\calB$ be a bilinear block of \cref{def:bb} on the hidden space $\calH$ of maximal degree $\lambda$ and width $C_{\mathrm H}$, and write its four affine maps in the form \cref{eq:linear},
\begin{equation*}
  (L_ix)^{(l)}_{m,d}=\sum_{c}W^{(l)}_{i,dc}\,x^{(l)}_{m,c}+\delta_{l0}\,b_{i,d},\qquad i=1,\ldots,4,
\end{equation*}
with one matrix $W^{(l)}_i$ per degree and a bias $b_i$ in the degree-$0$ block.
Let $k_{\max}\ge1$ be the frame truncation of the WBB to be built, so that its frame sets contain the frames $0$ and, for $l\ge1$, the frames $\pm1$.

\paragraph{The bias terms.}
The factors are $u=W_1x+b_1$ and $v=W_2x+b_2$, and the product \cref{eq:bb-product} is bilinear in $(u,v)$, so
\begin{equation}\label{eq:sm-bb-split}
  t=t_{11}+t_{10}+t_{01}+t_{00},
\end{equation}
where $t_{11}$ is \cref{eq:bb-product} with $(W_1x,W_2x)$ in place of $(u,v)$, $t_{10}$ with $(W_1x,b_2)$, $t_{01}$ with $(b_1,W_2x)$, and $t_{00}$ with $(b_1,b_2)$.
The vector $b_2$ lies in degree $0$, so $t_{10}$ contains only the paths $(l,0,l)$.
The Clebsch--Gordan product of $w\in V_l$ with a scalar $\beta\in V_0$ is $(w\otimes\beta)^{(l)}=\varsigma_l\,\beta\,w$, with $\varsigma_l\in\{1,-1\}$.
Indeed, the identity on $V_l$ is an equivariant map with orthonormal columns, so uniqueness in \cref{sm:lem-real-cg} leaves only this sign.
Hence
\begin{equation*}
  (t_{10})^{(l)}_{m,c}=\varsigma_l\sum_{c_1,c_2}U^{(l0l)}_{c,c_1c_2}\,b_{2,c_2}\,(W_1x)^{(l)}_{m,c_1},
\end{equation*}
which is an equivariant linear map of $x$, with the matrix $\varsigma_l\,U^{(l0l)}_{\cdot,\cdot c_2}b_{2,c_2}$ summed over $c_2$ and composed with $W^{(l)}_1$ in each degree.
The same argument makes $t_{01}$ equivariant and linear, using the paths $(0,l,l)$ and the factor $W_2$.
The term $t_{00}$ is the constant $\varsigma_0\sum_{c_1,c_2}U^{(000)}_{c,c_1c_2}b_{1,c_1}b_{2,c_2}$ in the degree-$0$ block.
The output \cref{eq:bb-out} is $\calB(x)=L_3x+L_4t$, and inserting \cref{eq:sm-bb-split} gives
\begin{equation}\label{eq:sm-bb-affine-part}
  \calB(x)=\underbrace{W_3x+W_4(t_{10}+t_{01})}_{\text{equivariant linear}}
  +\underbrace{b_3+b_4+W_4t_{00}}_{\text{degree-$0$ constant}}
  +W_4t_{11}.
\end{equation}
Choose $P^{(l)}$ to be the degree-$l$ matrix of the first group and $b$ to be the second group.
These choices supply the affine term in \cref{eq:wbb-out}.
It remains to produce $W_4t_{11}$ from the analyzed coefficients, and $t_{11}$ involves the linear parts $W_1x$ and $W_2x$ only.

\paragraph{The field pairs.}
Index the terms of $t_{11}$ by a coupling path and an input-channel pair,
\begin{equation}\label{eq:sm-bb-triples}
  a=\bigl((l_1,l_2,l),c_1,c_2\bigr),\qquad l_1,l_2\le\lambda,\quad |l_1-l_2|\le l\le\min(\lambda,l_1+l_2),\quad c_1,c_2\le C_{\mathrm H},
\end{equation}
where the output degree is restricted to $l\le\lambda$ because the hidden space has no higher degree.
Each triple contributes the vector $U^{(l_1l_2l)}_{\cdot,c_1c_2}\,((W_1x)^{(l_1)}_{\cdot,c_1}\otimes(W_2x)^{(l_2)}_{\cdot,c_2})^{(l)}$ to degree $l$ of $t_{11}$.
The WBB has one field pair per triple, so $F$ is the number of triples \cref{eq:sm-bb-triples}, as in the construction proving \cref{prop:wbb-full}.
For the triple $a$ let $(k_1,k_2)$ be the frame pair of \cref{lem:frame-truncation} for its path, with $|k_1|\le\min(1,l_1)$, $|k_2|\le\min(1,l_2)$, and
\begin{equation}\label{eq:sm-bb-frame-factor}
  \gamma_a=C^{(l,0)}_{(l_1,k_1)(l_2,k_2)}\ne0 .
\end{equation}

\paragraph{The lift.}
The lift \cref{eq:wbb-lift} of the pair $a$ carries the channel $c_1$ of $W_1x$ in the degree $l_1$ and the frame $k_1$ on the left, and the channel $c_2$ of $W_2x$ in the degree $l_2$ and the frame $k_2$ on the right,
\begin{equation}\label{eq:sm-bb-lift}
  A^{(l')}_{ka,c}=\delta_{l'l_1}\,\delta_{kk_1}\,W^{(l_1)}_{1,c_1c},\qquad
  B^{(l')}_{ka,c}=\delta_{l'l_2}\,\delta_{kk_2}\,W^{(l_2)}_{2,c_2c},
\end{equation}
so that the lifted coefficients are
\begin{equation*}
  u^{(l')}_{m,k,a}=\delta_{l'l_1}\,\delta_{kk_1}\,(W_1x)^{(l_1)}_{m,c_1},\qquad
  v^{(l')}_{m,k,a}=\delta_{l'l_2}\,\delta_{kk_2}\,(W_2x)^{(l_2)}_{m,c_2}.
\end{equation*}
Each field of the pair contains one degree and one frame.

\paragraph{The analyzed coefficients.}
Apply \cref{eq:sm-wbb-product-coeff} to the pair $a$.
In the sum over $(l'_1,l'_2)$ and $(k'_1,k'_2)$ only the term $l'_1=l_1$, $l'_2=l_2$, $k'_1=k_1$, $k'_2=k_2$ survives, since the lifted coefficients vanish otherwise.
The full product has the degrees of the triangle range of $(l_1,l_2)$, up to $2\lambda$, but the analysis \cref{eq:wbb-analysis} computes $z^{(l')}$ only for $l'\le\lambda$ and $k\in\calS_{l'}$, so
\begin{equation}\label{eq:sm-bb-analyzed}
  \begin{aligned}
    z^{(l')}_{m,k,a}&=C^{(l',k)}_{(l_1,k_1)(l_2,k_2)}\,
    \bigl((W_1x)^{(l_1)}_{\cdot,c_1}\otimes(W_2x)^{(l_2)}_{\cdot,c_2}\bigr)^{(l')}_m\\
    &\quad\text{for }|l_1-l_2|\le l'\le\min(\lambda,l_1+l_2),\ k\in\calS_{l'},
  \end{aligned}
\end{equation}
and $z^{(l')}_{\cdot,\cdot,a}=0$ for the other degrees $l'\le\lambda$.
The analysis includes all retained entries in these degrees and frames.
The output map selects only degree $l$ of the assigned path and frame $k=0$, whose frame coefficient is $\gamma_a$ in \cref{eq:sm-bb-frame-factor}.

\paragraph{The output map.}
Set
\begin{equation}\label{eq:sm-bb-Q}
  Q^{(l')}_{d,ka}=\delta_{l'l}\,\delta_{k0}\,\frac{1}{\gamma_a}\sum_{c}W^{(l)}_{4,dc}\,U^{(l_1l_2l)}_{c,c_1c_2}
  \qquad\text{for }a=\bigl((l_1,l_2,l),c_1,c_2\bigr),
\end{equation}
so that $Q$ reads every pair in its own output degree and in the frame $0$ only.
Then, by \cref{eq:sm-bb-analyzed} with $l'=l$ and $k=0$,
\begin{equation*}
\begin{split}
  \sum_{k\in\calS_{l}}\sum_{a}Q^{(l)}_{d,ka}\,z^{(l)}_{m,k,a}
  &=\sum_{c}W^{(l)}_{4,dc}\sum_{\substack{l_1,l_2\le\lambda\\ |l_1-l_2|\le l\le l_1+l_2}}\ \sum_{c_1,c_2}
    U^{(l_1l_2l)}_{c,c_1c_2}\,\bigl((W_1x)^{(l_1)}_{\cdot,c_1}\otimes(W_2x)^{(l_2)}_{\cdot,c_2}\bigr)^{(l)}_m\\
  &=\bigl(W_4t_{11}\bigr)^{(l)}_{m,d},
\end{split}
\end{equation*}
where the sum over $a$ has been split into the sum over the paths with output degree $l$ and the sum over the channel pairs, and the factor $\gamma_a$ has cancelled.
With $P$ and $b$ as chosen after \cref{eq:sm-bb-affine-part}, the output \cref{eq:wbb-out} of the WBB is
\begin{equation*}
  y=Px+W_4t_{11}+b=\calB(x)
\end{equation*}
for every $x\in\calH$, which is the converse of \cref{prop:wbb-full}.

The count of field pairs is not the smallest possible.
A pair whose two fields contain several degrees and channels produces several products at once by \cref{eq:sm-wbb-product-coeff}, and the frames $k\ne0$ of the output offer further slots to read them from.
The one-pair-per-triple assignment keeps every product in a slot of its own without further conditions, and the count $F$ enters the results of the main text only as a finite number.

\ifdefined\arxivbuild\else
\bibliographystyle{siamplain}
\bibliography{refs}

\begin{thebibliography}{10}

\bibitem{bartok2010gaussian}
{\sc A.~P. Bart{\'o}k, M.~C. Payne, R.~Kondor, and G.~Cs{\'a}nyi}, {\em
  Gaussian approximation potentials: The accuracy of quantum mechanics, without
  the electrons}, Phys. Rev. Lett., 104 (2010), p.~136403.

\bibitem{batatia2022mace}
{\sc I.~Batatia, D.~P. Kov{\'a}cs, G.~N.~C. Simm, C.~Ortner, and
  G.~Cs{\'a}nyi}, {\em {MACE}: Higher order equivariant message passing neural
  networks for fast and accurate force fields}, in Advances in Neural
  Information Processing Systems, 2022.

\bibitem{nequip2022}
{\sc S.~Batzner, A.~Musaelian, L.~Sun, M.~Geiger, J.~P. Mailoa, M.~Kornbluth,
  N.~Molinari, T.~E. Smidt, and B.~Kozinsky}, {\em E(3)-equivariant graph
  neural networks for data-efficient and accurate interatomic potentials},
  Nature Communications, 13 (2022), p.~2453,
  \url{https://doi.org/10.1038/s41467-022-29939-5}.

\bibitem{behler2007generalized}
{\sc J.~Behler and M.~Parrinello}, {\em Generalized neural-network
  representation of high-dimensional potential-energy surfaces}, Phys. Rev.
  Lett., 98 (2007), p.~146401.

\bibitem{blanco1997rotation}
{\sc M.~A. Blanco, M.~Fl{\'o}rez, and M.~Bermejo}, {\em Evaluation of the
  rotation matrices in the basis of real spherical harmonics}, Journal of
  Molecular Structure: THEOCHEM, 419 (1997), pp.~19--27.

\bibitem{cen2024hegnn}
{\sc J.~Cen, A.~Li, N.~Lin, Y.~Ren, Z.~Wang, and W.~Huang}, {\em Are
  high-degree representations really unnecessary in equivariant graph neural
  networks?}, in Advances in Neural Information Processing Systems, vol.~37,
  2024.

\bibitem{cen2025universally}
{\sc J.~Cen, A.~Li, N.~Lin, T.~Xu, Y.~Rong, D.~Zhao, Z.~Wang, and W.~Huang},
  {\em Universally invariant learning in equivariant {GNN}s}, in Proc. NeurIPS,
  2025.
\newblock arXiv:2510.13169.

\bibitem{chen2016locality}
{\sc H.~Chen and C.~Ortner}, {\em {QM/MM} methods for crystalline defects.
  {Part} 1: Locality of the tight binding model}, Multiscale Model. Simul., 14
  (2016), pp.~232--264.

\bibitem{cohen2018spherical}
{\sc T.~S. Cohen, M.~Geiger, J.~K{\"o}hler, and M.~Welling}, {\em Spherical
  {CNN}s}, in International Conference on Learning Representations (ICLR),
  2018.

\bibitem{deng2023chgnet}
{\sc B.~Deng, P.~Zhong, K.~Jun, J.~Riebesell, K.~Han, C.~J. Bartel, and
  G.~Ceder}, {\em {CHGNet} as a pretrained universal neural network potential
  for charge-informed atomistic modelling}, Nat. Mach. Intell., 5 (2023),
  pp.~1031--1041.

\bibitem{dusson2022ace}
{\sc G.~Dusson, M.~Bachmayr, G.~Cs{\'a}nyi, R.~Drautz, S.~Etter, C.~van~der
  Oord, and C.~Ortner}, {\em Atomic cluster expansion: Completeness, efficiency
  and stability}, Journal of Computational Physics, 454 (2022), p.~110946.

\bibitem{dym2024low}
{\sc N.~Dym and S.~J. Gortler}, {\em Low-dimensional invariant embeddings for
  universal geometric learning}, Found. Comput. Math., 25 (2025), pp.~375--415.

\bibitem{dym2021universality}
{\sc N.~Dym and H.~Maron}, {\em On the universality of rotation equivariant
  point cloud networks}, in Proc. ICLR, 2021.

\bibitem{finkelshtein2022simple}
{\sc B.~Finkelshtein, C.~Baskin, H.~Maron, and N.~Dym}, {\em A simple and
  universal rotation equivariant point-cloud network}, in Proc. Topological,
  Algebraic, and Geometric Learning Workshops, PMLR 196, 2022.

\bibitem{ivanic1996rotation}
{\sc J.~Ivanic and K.~Ruedenberg}, {\em Rotation matrices for real spherical
  harmonics. {D}irect determination by recursion}, The Journal of Physical
  Chemistry, 100 (1996), pp.~6342--6347.

\bibitem{joshi2023expressive}
{\sc C.~K. Joshi, C.~Bodnar, S.~V. Mathis, T.~Cohen, and P.~Li{\`o}}, {\em On
  the expressive power of geometric graph neural networks}, in Proc. ICML,
  2023, pp.~15330--15355.

\bibitem{dpa4}
{\sc T.~Li, W.~Li, A.~Peng, J.~Xue, L.~Zhang, D.~Zhang, and H.~Wang}, {\em
  Pushing the accuracy--cost frontier of machine-learning interatomic
  potentials}.
\newblock Manuscript in preparation, 2026.

\bibitem{liao2023equiformerv2}
{\sc Y.-L. Liao, B.~Wood, A.~Das, and T.~Smidt}, {\em {EquiformerV2}: Improved
  equivariant transformer for scaling to higher-degree representations}, in
  International Conference on Learning Representations (ICLR), 2024.

\bibitem{hgnn}
{\sc P.~Ming and H.~Wang}, {\em Why multi-layer message passing works:
  Completeness theory for graph neural network interatomic potentials}.
\newblock arXiv:2609.00528, 2026.

\bibitem{nigam2024completeness}
{\sc J.~Nigam, S.~N. Pozdnyakov, K.~K. Huguenin-Dumittan, and M.~Ceriotti},
  {\em Completeness of atomic structure representations}, APL Mach. Learn., 2
  (2024), p.~016110.

\bibitem{ortner2020locality}
{\sc C.~Ortner, J.~Thomas, and H.~Chen}, {\em Locality of interatomic forces in
  tight binding models for insulators}, ESAIM Math. Model. Numer. Anal., 54
  (2020), pp.~2295--2318.

\bibitem{passaro2023reducing}
{\sc S.~Passaro and C.~L. Zitnick}, {\em Reducing {SO}(3) convolutions to
  {SO}(2) for efficient equivariant {GNN}s}, in International Conference on
  Machine Learning (ICML), 2023.

\bibitem{pozdnyakov2022incompleteness}
{\sc S.~N. Pozdnyakov and M.~Ceriotti}, {\em Incompleteness of graph
  convolutional neural networks for point clouds in three dimensions}, arXiv
  preprint arXiv:2201.07136,  (2022).

\bibitem{sverdlov2025expressive}
{\sc Y.~Sverdlov and N.~Dym}, {\em On the expressive power of sparse geometric
  {MPNN}s}, in Proc. ICLR, 2025.
\newblock arXiv:2407.02025.

\bibitem{thomas2022body}
{\sc J.~Thomas, H.~Chen, and C.~Ortner}, {\em Body-ordered approximations of
  atomic properties}, Arch. Ration. Mech. Anal., 246 (2022), pp.~1--60.

\bibitem{thomas2018tfn}
{\sc N.~Thomas, T.~Smidt, S.~Kearnes, L.~Yang, L.~Li, K.~Kohlhoff, and
  P.~Riley}, {\em Tensor field networks: Rotation- and translation-equivariant
  neural networks for 3d point clouds}, arXiv preprint arXiv:1802.08219,
  (2018).

\bibitem{varshalovich1988}
{\sc D.~A. Varshalovich, A.~N. Moskalev, and V.~K. Khersonskii}, {\em Quantum
  Theory of Angular Momentum: Irreducible Tensors, Spherical Harmonics, Vector
  Coupling Coefficients, $3nj$ Symbols}, World Scientific, Singapore, 1988.

\bibitem{villar2021scalars}
{\sc S.~Villar, D.~W. Hogg, K.~Storey-Fishman, W.~Yao, and B.~Blum-Smith}, {\em
  Scalars are universal: Equivariant machine learning, structured like
  classical physics}, Advances in Neural Information Processing Systems, 34
  (2021).

\bibitem{yarotsky2022universal}
{\sc D.~Yarotsky}, {\em Universal approximations of invariant maps by neural
  networks}, Constr. Approx., 55 (2022), pp.~407--474.

\bibitem{zhang2025dpa3}
{\sc D.~Zhang, J.~Zeng, W.~Jia, H.~Wang, and L.~Zhang}, {\em {DPA-3}: A
  scalable model for the era of large atomistic models}, arXiv:2506.01686,
  (2025).

\end{thebibliography}
\fi

\end{document}